%% file: main.tex
\RequirePackage{fix-cm}
\documentclass[twocolumn]{svjour3}          
\smartqed  
\usepackage{graphicx}
\usepackage[utf8]{inputenc} 
\usepackage[T1]{fontenc}    
\usepackage{url}            
\usepackage{booktabs}       
\usepackage{amsfonts}       
\usepackage{nicefrac}       
\usepackage{microtype}      
\usepackage{epsfig}
\usepackage{float}
\usepackage{multicol}
\usepackage{multirow}
\usepackage{amssymb}
\usepackage{amsmath}
\usepackage{wrapfig}
\usepackage{xspace}
\usepackage{color}
\usepackage{colortbl}
\usepackage[dvipsnames, svgnames, x11names, table]{}
\usepackage{xcolor}
\usepackage[misc]{ifsym}
\usepackage{makecell}
\usepackage{soul}
\usepackage{bm}
\usepackage{wrapfig}
\usepackage{subcaption, overpic, textpos}
\usepackage{ulem}
\usepackage{adjustbox}
\usepackage{pifont}
\usepackage{caption}
\usepackage{array}
\usepackage{enumitem}
\newcolumntype{C}[1]{>{\centering\arraybackslash}p{#1}} 

\usepackage{cite}

\definecolor{linkcolor}{RGB}{255,0,0}
\definecolor{urlcolor}{RGB}{255,105,180}
\definecolor{citecolor}{RGB}{66,168,235}
\usepackage[pagebackref,breaklinks=true,colorlinks,citecolor=blue,urlcolor=blue,linkcolor=blue,bookmarks=false]{hyperref}

\newcommand{\Fig}{Fig.\@\xspace}

\newcommand{\Tab}{Table\@\xspace}

\newcommand{\Sec}{Sec.\@\xspace}

\newlength\savewidth
\newcommand{\tablestyle}[2]{\setlength{\tabcolsep}{#1}\renewcommand{\arraystretch}{#2}\centering\footnotesize}
\renewcommand{\paragraph}[1]{\vspace{1.25mm}\noindent\textbf{#1}}

\def\onedot{.\xspace}
\def\eg{\textit{e.g}\onedot} 
\def\Eg{\textit{E.g}\onedot}
\def\ie{\textit{i.e}\onedot}

\def\cf{\textit{c.f}\onedot}

\def\vs{\textit{vs}\onedot}

\def\aka{\textit{a.k.a}\onedot}

\def \pzo {\phantom{0}} 
\newcommand{\cmark}{\ding{52}\xspace}%
\newcommand{\xmark}{\ding{56}\xspace}%
\newcommand{\xmarkg}{\textcolor{lightgray}{\ding{56}}\xspace}%

\definecolor{lightgray}{rgb}{0.8, 0.8, 0.8}
\definecolor{lgray}{rgb}{0.66, 0.66, 0.66}
\definecolor{whit_tab}{RGB}{255, 255, 255}
\definecolor{gray_tab}{RGB}{235, 235, 235}
\definecolor{oran_tab}{RGB}{254, 247, 241}
\definecolor{blue_tab}{RGB}{200, 227, 245}
\definecolor{lblu_tab}{RGB}{231, 239, 248}
\definecolor{teaser_blue}{RGB}{106, 153, 208}
\definecolor{teaser_orange}{RGB}{222, 131, 68}

\newcommand{\deepred}{\textcolor[RGB]{176, 36, 24}}

\newcommand{\blue}{\textcolor[RGB]{0, 0, 255}}
\newcommand{\redzjn}{\textcolor[RGB]{255, 0, 0}}

\usepackage{tablefootnote}
\usepackage{diagbox}

\begin{document}
\begin{sloppypar}
\title{Learning Feature Inversion for Multi-class Anomaly Detection under General-purpose COCO-AD Benchmark
}

\author{Jiangning Zhang$^{1,2}$ \and
        Chengjie Wang$^{2}$ \and
        Xiangtai Li$^{3}$ \and
        Guanzhong Tian$^{1}$ \and
        Zhucun Xue$^{1}$ \and
        Yong Liu$^{1}$ \and
        Guansong Pang$^{4}$ \and
        Dacheng Tao$^{3}$
}

\authorrunning{Jiangning Zhang, et al.}
\institute{
  Jiangning Zhang (186368@zju.edu.cn) \\\
  \Letter~Yong Liu (yongliu@iipc.zju.edu.cn) \\
$^{1}$ Zhejiang University, Hangzhou, China. \\
$^{2}$ Youtu Lab, Tencent, Shanghai, China. \\
$^{3}$ Nanyang Technological University, Singapore. \\
$^{4}$ Singapore Management University, Singapore. 
}

\date{Received: date / Accepted: date}
\maketitle

\input{secs/0_abstract}

\input{secs/1_introduction}

\input{secs/2_related_work}
\input{secs/3_method}

\input{secs/4_experiments}
\input{secs/5_conclusion}

\bibliographystyle{spmpsci}
\bibliography{main}

\input{secs/6_appendix}

\end{sloppypar}
\end{document}

%% file: secs/0_abstract.tex
\begin{abstract}
Anomaly detection (AD) is often focused on detecting anomaly areas for industrial quality inspection and medical lesion examination. 
However, due to the specific scenario targets, the data scale for AD is relatively small, and evaluation metrics are still deficient compared to classic vision tasks, such as object detection and semantic segmentation. 
To fill these gaps, this work first constructs a large-scale and general-purpose COCO-AD dataset by extending COCO to the AD field. 
This enables fair evaluation and sustainable development for different methods on this challenging benchmark.
Moreover, current metrics such as AU-ROC have nearly reached saturation on simple datasets, which prevents a comprehensive evaluation of different methods. 
Inspired by the metrics in the segmentation field, we further propose several more practical threshold-dependent AD-specific metrics, \ie, m$F_1$$^{.2}_{.8}$, mAcc$^{.2}_{.8}$, mIoU$^{.2}_{.8}$, and mIoU-max.
Motivated by GAN inversion's high-quality reconstruction capability, we propose a simple but more powerful InvAD framework to achieve high-quality feature reconstruction.
Our method improves the effectiveness of reconstruction-based methods on popular MVTec AD, VisA, and our newly proposed COCO-AD datasets under a multi-class unsupervised setting, where only a single detection model is trained to detect anomalies from different classes. 
Extensive ablation experiments have demonstrated the effectiveness of each component of our InvAD. 
Full codes and models are available at \url{https://github.com/zhangzjn/ader}. 

\end{abstract}

%% file: secs/1_introduction.tex
\section{Introduction} 
\label{sec:intro}

Visual unsupervised anomaly detection only requires training on normal images to achieve classification/segmentation of abnormal images, \aka, detection/localization by works~\cite{survey_ad,cao2024survey}. 
This technique is receiving increasing attention due to its low-cost unsupervised training approach and significant practical implications, \eg, industrial defect detection and medical lesion detection. 
Current methods~\cite{ocrgan,rd,simplenet} generally follow a single-class setting in which each class requires a separate model, significantly increasing the training and memory costs. 
Recently, UniAD~\cite{uniad} proposes a challenging yet more practical multi-class setting (detailed in \Sec\ref{sec:related}), 
in which we need to handle multiple anomaly classes using just a single model. 
This paper investigates this challenging task, delving into the three key elements: dataset, metrics, and methods.

\begin{figure*}[tp]
    \centering
    \includegraphics[width=1.0\linewidth]{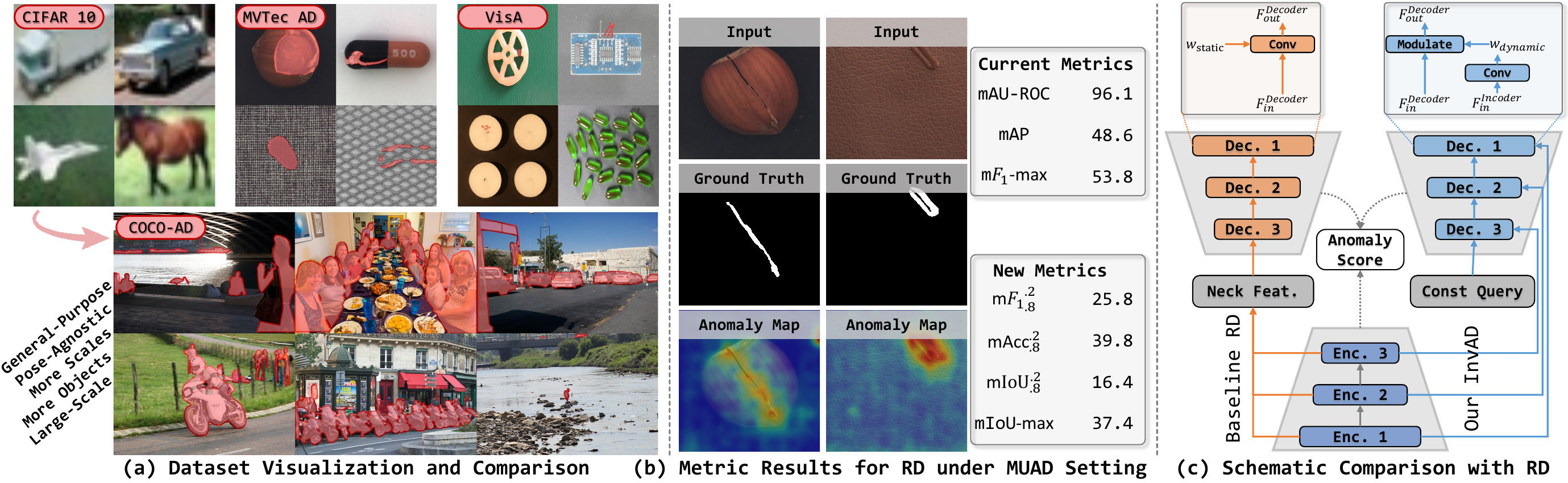}
    \vspace{-1.0em}
    \caption{
    \textbf{Left}: Comparison among representative anomaly detection datasets and our proposed general-purpose COCO-AD. 
    \textbf{Middle}: Example visualization of RD on MVTec AD dataset~\cite{mvtec}. The excessively high current metrics (especially mAU-ROC) does not align with the visualization results, while our segmentation-inspired metrics provide more objective outcomes. 
    \textbf{Right}: Schematic comparison between RD~\cite{rd} and our \textit{feature inversion} framework.} 
    \label{fig:teaser}
    \vspace{-1.5em}
\end{figure*}

By analyzing current representative AD datasets~\cite{mvtec,visa,vitad}, we observe that they are domain-specific with the simple background scene, and the number of images and categories is much smaller compared with related semantic segmentation dataset like COCO~\cite{coco}. 
Taking the popular MVTec AD~\cite{mvtec} dataset as an example (\Fig\ref{fig:teaser}-Left), it only contains 15 industrial products in 2 types with 3,629 normal images for training and 467/1,258 normal/anomaly images for testing (5,354 images in total). 
There is no AD dataset with a status equivalent to ImageNet-1K~\cite{imagenet} in the classification field or COCO~\cite{coco} in detection/segmentation fields. 
Therefore, starting from essential one-class classification ability of visual AD task and considering the cost of constructing a dataset, this paper proposes an AD-specific, general-purpose, and large-scale COCO-AD based on recycling COCO 2017~\cite{coco} to evaluate different algorithms under a more challenging and fair benchmark (\Sec\ref{sec:method_cocoad}). 
For example, if ``car" is considered as an normal class, it will not appear in the training set but need to be detected in the test phase. This can assist us in detecting non-specific new semantic regions for practical application.

In addition, we analyze the current evaluation metrics, \ie, image-/pixel-level AU-ROC~\cite{dae}, AP~\cite{draem}, and $F_1$-max~\cite{visa}, as well as AU-PRO~\cite{uninformedstudents}. 
Since these metrics are not primitively designed for the AD field, they have certain flaws in evaluation (\Sec\ref{sec:dataset_metric}). \Eg, RD~\cite{rd} reaches very high threshold-independent sorting AU-ROC score on MVTec AD (\Fig\ref{fig:teaser}-Middle) and it has nearly reached saturation under the single-class setting, meaning that AU-ROC cannot effectively distinguish the merits of different algorithms. 
Appropriate AD evaluation criteria are essential for judging the merits of different methods. 
Therefore, based on practical application perspectives, we design four practical and threshold-dependent AD metrics by observing anomaly map distribution and borrowing from semantic segmentation field (\Sec\ref{sec:method_metric}).

\begin{figure*}[tp]
    \centering
    \includegraphics[width=0.9\linewidth]{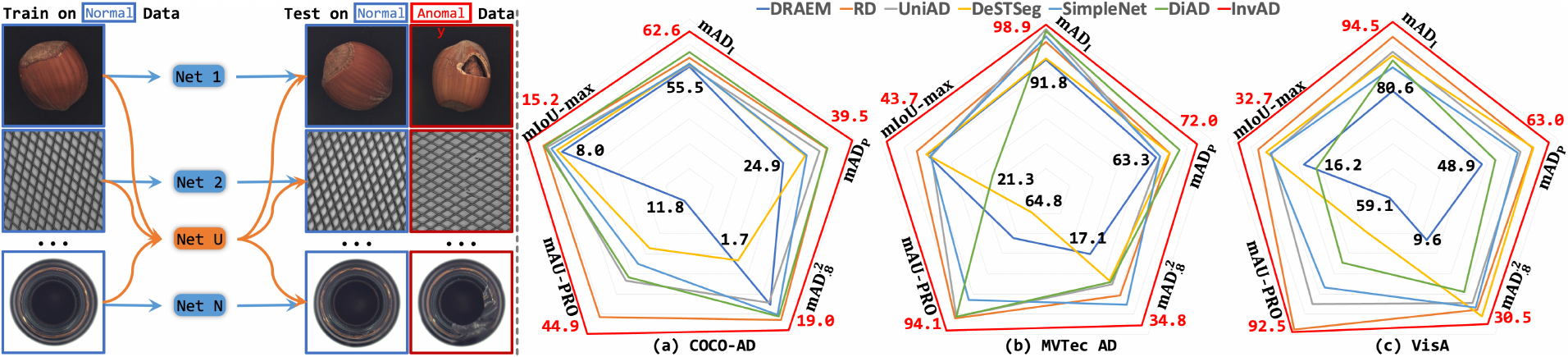}
    \caption{
    \textbf{Left}: Schematic AD task. 
    \textbf{Right}: Performance comparison among our InvAD and SoTA methods.
    }
    \label{fig:muad}
    \vspace{-2.0em}
\end{figure*}

After that, we categorize current AD methods into three types. \textbf{\textit{1)}} Augmentation-based methods~\cite{cutpaste,draem,defectgan,dtgan,simplenet} design various anomaly synthesis methods to assist model training. However, these approaches increase training complexity and may lead to dataset bias, resulting in weaker generalization ability. 
\textbf{\textit{2)}} Embedding-based methods~\cite{cflow,destseg,uninformedstudents,padim,patchcore} generally require feature comparison in high-dimensional latent space, which is computationally expensive and susceptible to noise data. 
\textbf{\textit{3)}} Reconstruction-based methods~\cite{dae,ganomaly,ocrgan,rd,uniad,gnl} follow the naive auto-encoder architecture, which can locate abnormal areas through reconstruction error without the need for carefully designed complex modules. 
For instance, RD~\cite{rd} achieves impressive results using only cosine distance~\cite{cosine} as the reconstruction loss. 
This paper explores this type of technique in-depth. 
Essentially, modeling under the multi-class setting is particularly challenging, while the effectiveness of reconstruction-based methods largely benefits from their reconstruction capabilities~\cite{survey_ad}. 
Inspired by the high-quality image reconstruction ability of GAN inversion~\cite{gan_inversion_survey}, we extend this concept to feature-level reconstruction by borrowing the StyleGAN~\cite{stylegan1} design, expecting locating the abnormal area through high-precision reconstruction error. 
As shown in \Fig\ref{fig:teaser}-Right, we introduce a dynamic modulation mechanism for high-precision feature reconstruction from constant query, which is significantly different from RD~\cite{rd} using static convolution to reconstruct target features from neck features. 
Benefit from input-aware modeling, the proposed InvAD achieves obvious superiority over different types of state-of-the-art (SoTA) methods (see \Fig\ref{fig:muad}-Right and \Sec\ref{exp:comparison} for more details).

In summary, this research elevates the multi-class AD task to a more practical level from the perspectives of the dataset, evaluation metrics, and methods. Our contributions are four-fold: \\
\textbf{\textit{1)}} For dataset, in response to the issues in current mainstream visual AD datasets, we extend the COCO dataset to general visual AD scenarios and propose a highly challenging benchmark to evaluate different methods fairly. We reproduce different types of recent methods on this benchmark (\Sec\ref{sec:method_cocoad}). \\
\textbf{\textit{2)}} For metric, considering the limitations of current AD metrics, we propose four additional pixel-level metrics that are more in line with actual application scenarios and suggest reporting five average metrics to holistically evaluate the merits of different methods (\Sec\ref{sec:method_metric}). \\
\textbf{\textit{3)}} For method, drawing on the concept of GAN inversion, we design a novel InvAD and propose an Spatial Style Modulation (SSM) module to ensure input-dependent and high-quality reconstruction. Without bells and whistles, InvAD achieves very competitive results by using only pixel-level MSE constraint (\Sec\ref{sec:invad}). \\
\textbf{\textit{4)}} Our method shows a significant improvement over SoTA methods on the proposed challenging COCO-AD dataset. We also conduct experiments on other representative datasets, achieving obvious new heights in multi-class AD task (\Sec\ref{sec:exp}).

%% file: secs/2_related_work.tex
\vspace{-1.0em}
\section{Related Work} \label{sec:related}
\vspace{-0.5em}
\noindent\textbf{Visual Anomaly Detection} follows an unsupervised setting, assuming the model only sees normal data during training and must distinguish between normal and anomaly data during testing. This paper categorizes current methods into three main taxonomies. 
\textbf{\textit{1)}} \textit{Augmentation-based} methods typically involve synthesizing anomalous regions on normal images~\cite{cutpaste,draem,dtd,defectgan,dtgan,anomalydiffusion} or adding anomalous information on normal features to construct pseudo-supervisory signals for training~\cite{uniad,simplenet}. 
Recently, multi-class UniAD~\cite{uniad} introduced a feature jittering strategy. The above methods achieve good results, and the augmentation parts can generally be embedded as plug-and-play modules into other approaches to improve performance further, but they require carefully designed model architectures and significantly increase model complexity.  
\textbf{\textit{2)}} \textit{Embedding-based} methods generally use a pre-trained model~\cite{imagenet} to extract rich feature representations and then distinguish between normal and abnormal features in a compact feature space. They can generally be divided into three categories~\cite{survey_ad}: 
\textit{i)} distribution-map based~\cite{nf1,pyramidflow,wu2023open,inctrl}, 
\textit{ii)} teacher-student based~\cite{pfm,cao2022informative,destseg,uninformedstudents,wang2021glancing}, 
and \textit{iii)} memory-bank based~\cite{salehi2021multiresolution,liu2023diversity,gcpf,patchcore,jang2023n} methods.
\textbf{\textit{3)}} \textit{Reconstruction-based} methods typically include an encoder that maps the input to a high-dimensional hidden space and a decoder that restores the original data, with anomalies located based on data differences during testing. They can be divided into two categories based on the reconstructed data type. \textit{i)} Image-level methods~\cite{dae,ganomaly,yan2021unsupervised,draem,AnoSeg,ocrgan,diad} reconstruct images in the original RGB space, which can be seen as an auto-encoder structure to some extent.  
\textit{ii)} Feature-level methods~\cite{adtr,vitad,mambaad} further reconstruct input at a more expressive feature level. 
UniAD~\cite{uniad} introduces a layer-wise query decoder and feature jittering to better model feature-level distribution, while recent works~\cite{rd,rrd,gnl} introduce a reverse distillation paradigm to reconstruct multi-resolution features, further enhancing the model's sensitivity to the different granularity of features. Considering effectiveness and ease-of-use of reconstruction-based methods, this paper introduces the feature inversion concept to improve the technique, hoping to enhance the performance by improving the reconstruction ability.

\noindent\textbf{Multi-class Anomaly Detection} 
Most current methods~\cite{draem,rd,patchcore} predominantly focus on single-class unsupervised anomaly detection (SUAD). As illustrated in \Fig\ref{fig:muad}-Left (marked in blue), each class of objects requires a separate model for training and testing, leading to increased costs in training and deployment. Comparatively, we investigate the more challenging task of multi-class unsupervised anomaly detection (MUAD) that is widely employed by recent works~\cite{uniad,diad,vitad}. 
As shown in the orange indicator of \Fig\ref{fig:muad}, this more practical setting only requires one unified model to complete testing and deployment for multiple classes, and this paper focuses on this challenging setting. 

\begin{table*}[tp]
    \centering
    \caption{\textbf{Comparison of current popular AD datasets on different attributes.} \cmark: Satisfied. \xmark: Unsatisfied.}
    \vspace{-0.5em}
    \label{tab:dataset}
    \renewcommand{\arraystretch}{1.0}
    \setlength\tabcolsep{3.0pt}
    \resizebox{1.0\linewidth}{!}{
        \begin{tabular}{p{1.5cm}<{\centering} p{2.5cm}<{\centering} p{1.0cm}<{\centering} p{1.2cm}<{\centering} p{0.1cm}<{\centering} p{1.5cm}<{\centering} p{0.1cm}<{\centering} p{1.5cm}<{\centering} p{1.5cm}<{\centering} p{1.0cm}<{\centering} p{1.7cm}<{\centering} p{1.5cm}<{\centering} p{1.2cm}<{\centering} p{1.2cm}<{\centering} p{1.7cm}<{\centering} p{1.7cm}<{\centering}}
            \toprule[0.17em]
            \multirow{3}{*}{\makecell[c]{Information-\\ level}} & \multirow{3}{*}{Dataset} & \multicolumn{2}{c}{Classes} & & \multicolumn{4}{c}{Image Number} & \multirow{3}{*}{Split} & \multirow{3}{*}{\makecell[c]{Multi-\\ object/defect\\ per image}} & \multirow{3}{*}{\makecell[c]{General\\ Purpose} } & \multirow{3}{*}{\makecell[c]{Pose-\\ Agnostic}} & \multirow{3}{*}{\makecell[c]{Scale-\\ Agnostic}} & \multirow{3}{*}{\makecell[c]{Semantic-/\\ Appearance-\\ Agnostic}} & \multirow{3}{*}{\makecell[c]{Complex \\Background}} \\ 
            \cline{3-4} \cline{6-9}
            & & \multirow{2}{*}{Train} & \multirow{2}{*}{Test} & & \multicolumn{1}{c}{Train} & & \multicolumn{2}{c}{Test} & & & & & & & \\
            \cline{6-6} \cline{8-9}
            & & & & & Normal & & Normal & Anomaly & & & & & & & \\
            \hline
            Semantic & CIFAR10 & \pzo5 & 5+5 & & 25,000 & & 5,000 & 5,000 & 4 & \xmark & \xmark & \cmark & \cmark & \cmark & \xmark \\
            \hline
            \multirow{3}{*}{Sensory} & MVTec AD & 15 & 15 & & \pzo3,629 & & \pzo467 & 1,258 & 1 & \xmark & \xmark & \xmark & \xmark & \xmark & \xmark \\
            & MVTec AD-3D & 10 & 10 & & \pzo2,950 & & \pzo249 & \pzo948 & 1 & \xmark & \xmark & \xmark & \xmark & \xmark & \xmark \\
            & VisA & 12 & 12 & & \pzo8,659 & & \pzo962 & 1,200 & 1 & \xmark & \xmark & \xmark & \xmark & \xmark & \xmark \\
            \hline
            \multirow{4}{*}{\makecell[c]{Semantic\\ + \\Sensory}} & \multirow{4}{*}{COCO-AD} & \multirow{4}{*}{\makecell[c]{\pzo1\\ \pzo+\\ 60}} & \multirow{4}{*}{\makecell[c]{\pzo1\\ \pzo+\\ 60+20}} & & 30,438 & & 1,291 & 3,661 & \multirow{4}{*}{4} & \multirow{4}{*}{\cmark} & \multirow{4}{*}{\cmark} & \multirow{4}{*}{\cmark} & \multirow{4}{*}{\cmark} & \multirow{4}{*}{\cmark} & \multirow{4}{*}{\cmark} \\
            & & & & & 65,133 & & 2,785 & 2,167 & &  &  &  &  & \\
            & & & & & 79,083 & & 3,328 & 1,624 & &  &  &  &  & \\
            & & & & & 77,580 & & 3,253 & 1,699 & &  &  &  &  & \\
            \toprule[0.12em]
        \end{tabular}
    }
    \vspace{-1.5em}
\end{table*}

\noindent\textbf{Challenging AD Datasets} are extremely important for developing the AD field. Early methods~\cite{survey_ad} experiment with CIFAR10 dataset~\cite{cifar10} under the semantic setting. Since MVTec AD dataset~\cite{mvtec} is proposed, the conventional sensory setting has started to get on track. It gradually enters the field of view of researchers and practitioners, followed by a series of datasets being proposed thrillingly. \Eg, 
KolektorSDD~\cite{KolektorSDD}, MTD~\cite{mtd}, BTAD~\cite{btad}, MVTec LOCO AD~\cite{mvtecloco}, PAD~\cite{pad}, and the recently popular VisA~\cite{visa}. Recently, some 3D-related AD datasets have further been proposed, such as MVTec 3D AD~\cite{mvtec3d} 
and Real3D~\cite{real3d}. This paper focuses on unsupervised AD tasks in 2D scenarios. However, most current datasets are designed for manufacturing or specific application scenarios. In contrast, the category number and scale of datasets are relatively small, \ie, there is no AD dataset equivalent to the ImageNet-1K~\cite{imagenet} in the classification field or the COCO dataset~\cite{coco} in the detection/segmentation fields. Therefore, we extend the well-known COCO to a general-purpose and large-scale AD dataset, termed COCO-AD, to evaluate different algorithms. We also conduct a detailed discussion with popular MVTec AD and VisA datasets under the multi-class unsupervised AD setting in \Sec\ref{sec:method_cocoad}. 

\noindent\textbf{Reasonable Evaluation Metrics} are essential for judging the merits of different methods. Current AD metrics include AU-ROC~\cite{dae}, AP~\cite{draem}, $F_1$-max~\cite{visa}, and AU-PRO~\cite{uninformedstudents}, where the first three indicators will simultaneously report both image-/pixel-level results. Current methods generally use part of these metrics for comparison, but this is unreasonable since these metrics are not perfectly fitted, let alone use less (\cf, \Sec\ref{sec:method_metric}). \Eg, AU-ROC has almost reached saturation, and threshold-independent sorting metrics are difficult to guide practical applications. We argue that the current AD evaluation manner is insufficient, so we propose several more practical threshold-dependent metrics for a comprehensive comparison.

\noindent\textbf{GAN Inversion} aims to utilize a pre-trained generator to find the latent space encoding corresponding to a given image~\cite{gan_inversion_survey}, thereby enabling the faithful reconstruction of the input image via the inverted code by the generator. Current methods can be principally divided into \textbf{\textit{optimization-based}} and \textbf{\textit{learning-based}}. The former directly optimizes the latent vector to minimize the reconstruction error between input/output images~\cite{inv_optim2,stylegan2}. 
The latter trains a learnable encoder to implement the transformation from real images to hidden features~\cite{inv_learn1,inv_learn2,inv_learn3} (\cf, \Fig\ref{fig:invad}-Left). This method trades a certain level of editing accuracy for a one-forward process, and due to its faster speed, this approach has received more attention from researchers. Inspired by the latter, 
we extend this concept to high-quality feature reconstruction to determine anomalous regions. In other words, while GAN inversion plays an essential role in bridging real and fake image domains, our approach bridges real and fake feature domains.

%% file: secs/3_method.tex
\vspace{-1.0em}
\section{COCO-AD Dataset and Related Metrics} \label{sec:dataset_metric}

\begin{table*}[htp]
    \centering
    \caption{\textbf{Comparison of various AD metrics.} `m' prefix: averaged results. \redzjn{Red} metrics are recommended.}
    \label{tab:metric}
    \renewcommand{\arraystretch}{0.9}
    \setlength\tabcolsep{6.0pt}
    \resizebox{1.0\linewidth}{!}{
        \begin{tabular}{p{0.3cm}<{\centering} p{2.3cm}<{\centering} p{2.0cm}<{\centering} p{19.0cm}<{\centering}}
            \toprule[0.17em]
            & Fineness-level & Metric & Highlight \\
            \hline
            \multirow{3}{*}{\rotatebox{90}{\redzjn{mAD$_I$}}} & \multirow{3}{*}{\makecell[c]{Image-level\\ (Classification)}} & mAU-ROC & \makecell[l]{Area Under the ROC curve (TPR \vs FPR) to evaluate binary classification models | Threshold-independent sorting metric} \\ 
            & & mAP & \makecell[l]{Average Precision (\aka, AU-PR) that is Area Under the PR curve (Precision \vs Recall) | Threshold-independent sorting metric} \\ 
            & & m$F_1$-max & \makecell[l]{$F_1$-score at optimal threshold | Threshold-dependent metric | Against potential data imbalance} \\ 
            \hline
             & \multirow{1}{*}{\makecell[c]{Region-level}} & \redzjn{mAU-PRO} & \makecell[l]{Area Under the Per-Region-Overlap (PRO \vs FPR) | Weighting regions of different size equally | Threshold-independent sorting metric} \\ 
            \hline
            \multirow{3}{*}{\rotatebox{90}{\redzjn{mAD$_P$}}} & \multirow{3}{*}{\makecell[c]{Pixel-level\\ (Segmentation)}} & mAU-ROC & \makecell[l]{Same as above | Large predicted correct areas can compensate small incorrect areas} \\ 
            & & mAP & \makecell[l]{Same as above | More appropriate for highly imbalanced classes} \\ 
            & & m$F_1$-max & \makecell[l]{Same as above | Advantageous deployment-time} \\
            \cline{3-4}
            \multirow{3}{*}{\rotatebox{90}{\redzjn{mAD$^{.2}_{.8}$}}} & \multirow{4}{*}{\makecell[c]{Proposed}} & m$F_1$$^{.2}_{.8}$ & \makecell[l]{Average $F_1$-score with thresholds ranging from 0.2 to 0.8 | Threshold-dependent metric} \\
            & & mAcc$^{.2}_{.8}$ & \makecell[l]{Average Accuracy with thresholds ranging from 0.2 to 0.8 | Threshold-dependent metric} \\
            & & mIoU$^{.2}_{.8}$ & \makecell[l]{Average Intersection over Union with thresholds ranging from 0.2 to 0.8 | Threshold-dependent metric} \\
            & & \redzjn{mIoU-max} & \makecell[l]{Intersection over Union | Intuitively reflect the accuracy and overlap | Threshold-dependent metric} \\
            \toprule[0.12em] 
        \end{tabular}
    }
    \vspace{-1.0em}
\end{table*}

\vspace{-0.5em}
\subsection{General-purpose COCO-AD Benchmark} \label{sec:method_cocoad}
\vspace{-0.5em}

Anomaly detection datasets can be divided into two categories: semantic and sensory settings~\cite{survey_ad}. 
The former is typically represented by the CIFAR10 classification dataset~\cite{cifar10}. 
As shown in the first row in~\Tab\ref{tab:dataset}, this semantic setting treats a subset of 5 categories as normal and the rest as anomalies. 
It can be seen as a special case of novelty detection, differing from the general definition of visual AD. 
The popular sensory MVTec AD dataset~\cite{mvtec} represents the latter, commonly used in manufacturing scenarios to capture and detect unknown types of sensory defects, such as broken, scratched, or holed items. 
Due to its high practical value, subsequent research has primarily adopted this setting, and a series of related datasets have been proposed, such as recent VisA~\cite{visa} and Real-IAD~\cite{realiad}. 

\noindent\textbf{Motivation.}
\textit{We find that current datasets are mostly targeted at manufacturing or specific application scenarios, and the category number and scale are relatively small.} 
In addition, there is no general-purpose AD dataset with a status similar to the ImageNet-1K~\cite{imagenet} in the classification field or the COCO~\cite{coco} dataset in the detection and segmentation fields, which could result in insufficient method evaluation for practical use and limit the healthy development in the AD field. 
Therefore, it is particularly necessary to construct a large-scale and general-purpose challenging AD dataset.

\noindent\textbf{Adapting COCO to AD field.} 
Starting from essential one-class classification ability of visual AD task, we summarize the characteristics that an anomaly detection dataset for general scenarios should include: \textbf{\textit{1)}} large-scale data volume, \textbf{\textit{2)}} category diversity, \textbf{\textit{3)}} scenario diversity, \textbf{\textit{4)}} semantic texture diversity, and \textbf{\textit{5)}} non-fixed pose shooting. 
Specifically, we ponder the feasibility of current detection and segmentation datasets for AD tasks and intriguingly find that general scenario datasets with semantic segmentation annotations naturally fit AD tasks. 
Considering the cost and effectiveness of data acquisition, we propose a general-purpose COCO-AD benchmark based on COCO 2017~\cite{coco} to evaluate different AD methods fairly, \textit{which includes a variety of semantic categories and challenging sensory differences within each category}. 
Specifically, we take 20 categories in sequence as anomaly classes and the remaining 60 categories plus one background class as normal classes, constructing four splits without category overlapping for benchmarking. 
As a result, each split supports an MUAD setting of 80 classes natively. 
As shown in the right five columns of \Tab\ref{tab:dataset}, its intuitive attribute comparison with current popular datasets indicates the challenges of COCO-AD from multiple dimensions. 
Appendix~\blue{\textbf{A}} provides a statistical analysis of the four splits of the COCO-AD dataset across four dimensions to understand its characteristics for the AD task comprehensively. 
Furthermore, intuitive visualizations are shown in \Fig\ref{fig:teaser}-Left and Appendix~\blue{\textbf{C}}. 
Notably, we provides a construction approach that can be adopted for other related datasets. However, given the challenging nature of the current results (\Sec\ref{exp:comparison}) from COCO-AD, there is no need to construct larger datasets. 

\begin{table}[tb]
    \centering
    \caption{Results of the Pearson correlation coefficient between each pair of datasets based on the fundamental seven indicators and additional four new metrics.}
    \vspace{-0.5em}
    \label{tab:pearson}
    \renewcommand{\arraystretch}{0.9}
    \setlength\tabcolsep{1.0pt}
    \resizebox{1.0\linewidth}{!}{
        \begin{tabular}{p{0.3cm}<{\centering} p{2.6cm}<{\centering} p{1.0cm}<{\centering} p{1.0cm}<{\centering} p{1.0cm}<{\centering} p{0.2cm}<{\centering} p{1.0cm}<{\centering} p{1.0cm}<{\centering} p{1.0cm}<{\centering}}
        \toprule[1.5pt]
        \multirow{4}{*}{\makecell[c]{\rotatebox{90}{Dataset}}} & & \multicolumn{3}{c}{seven metrics} & & \multicolumn{3}{c}{w/ new metrics} \\
        \cline{3-5} \cline{7-9}
        & MVTec AD~\cite{mvtec} & \cmark & \cmark & \xmarkg & & \cmark & \cmark & \xmarkg \\
        & VisA~\cite{visa} & \cmark & \xmarkg & \cmark & & \cmark & \xmarkg & \cmark \\
        & COCO-AD & \xmarkg & \cmark & \cmark & & \xmarkg & \cmark & \cmark \\
        \hline
        \multirow{5}{*}{\makecell[c]{\rotatebox{90}{Method}}} & RD~\cite{rd} & 0.994 & 0.933 & 0.924 & & 0.987 & 0.924 & 0.953 \\
        & UniAD~\cite{uniad} & 0.988 & 0.893 & 0.890 & & 0.982 & 0.925 & 0.945 \\
        & SimpleNet~\cite{simplenet} & 0.984 & 0.903 & 0.872 & & 0.987 & 0.920 & 0.926 \\
        & DiAD~\cite{diad} & 0.989 & 0.878 & 0.912 & & 0.965 & 0.893 & 0.952 \\
        & InvAD & 0.996 & 0.931 & 0.930 & & 0.990 & 0.929 & 0.954 \\
        \bottomrule[1.5pt]
        \end{tabular}
    }
    \vspace{-1.0em}
\end{table}

\noindent\textbf{Feasibility evidence.} 
\Tab\ref{tab:pearson}-Left shows Pearson correlation coefficient between each pair of datasets based on the current seven image-/pixel-level AU-ROC~\cite{dae}/AP~\cite{draem}/$F_1$-max~\cite{visa} and region-level AU-PRO~\cite{uninformedstudents}. 
It can be observed that COCO-AD has a high correlation with the other two datasets, indicating its appropriateness for AD tasks. However, its correlation is slightly lower compared to the correlation between industrial MVTec AD and VisA datasets, which suggests its distinguished and general-purpose attributes. 
Notably, the consistency in the ranking of evaluation results in \Sec\ref{exp:comparison} from different SoTA methods validates the rationality of extending COCO to the AD task. If a method performs better on MVTec AD and VisA, it also tends to perform better on COCO-AD. 

\begin{figure}[tp]
    \centering
    \includegraphics[width=0.8\linewidth]{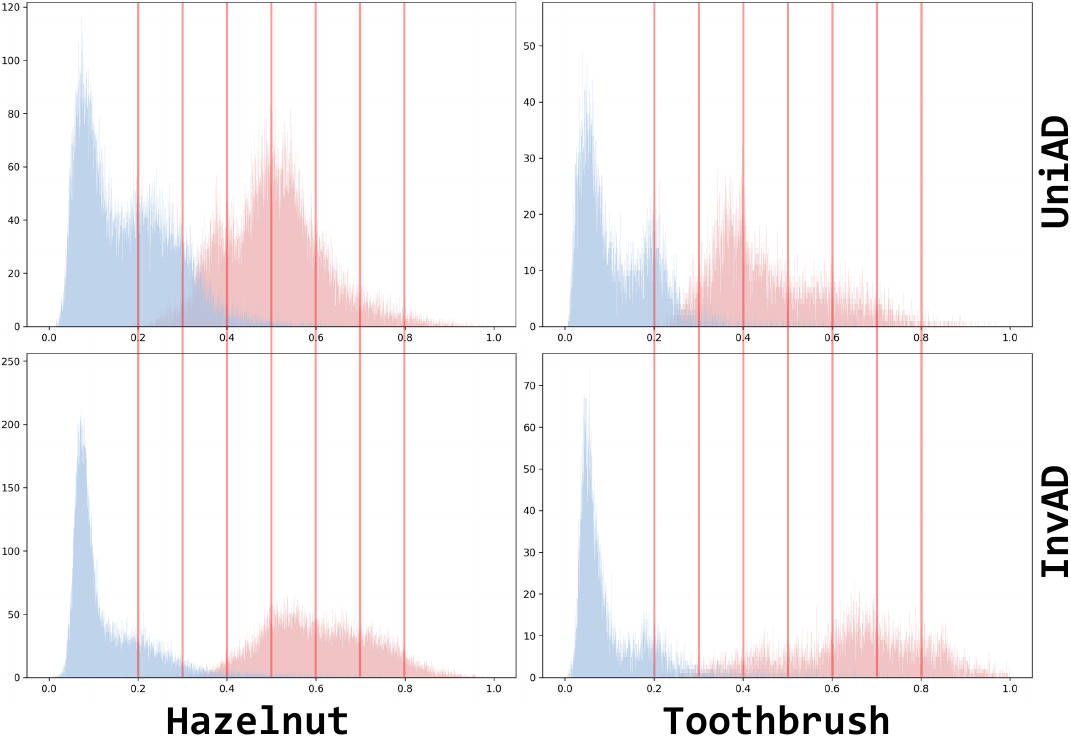}
    \caption{Anomaly score distribution on object Hazelnut and texture Toothbrush. Our method exhibits less overlap between normal and abnormal values and has a clearer demarcation line. Furthermore, the values tend to be closer to 0 (normal) and 1 (anomaly), respectively.}
    \label{fig:metric_28}
    \vspace{-0.5em}
\end{figure}

\begin{figure*}[tp]
    \centering
    \includegraphics[width=1.0\linewidth]{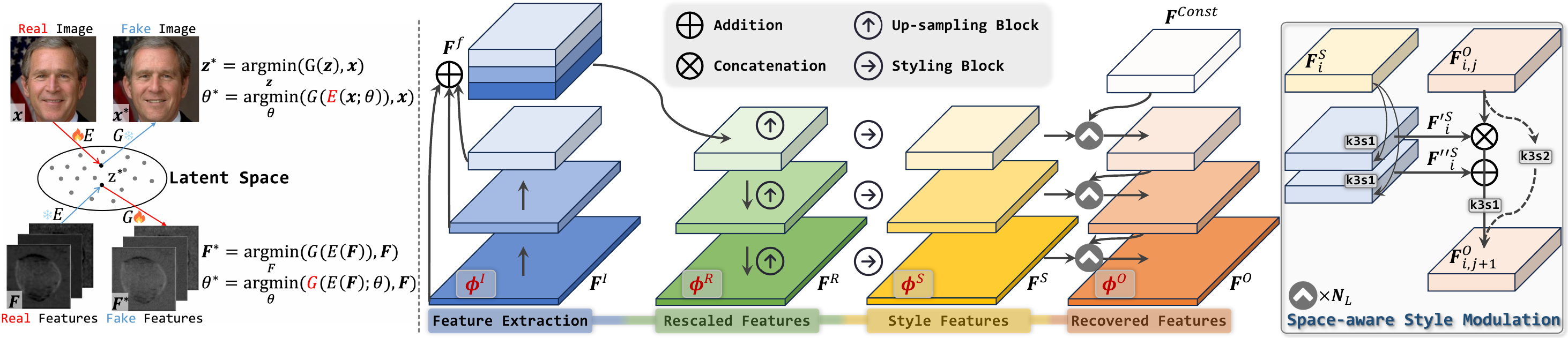}
    \caption{\textbf{Left}: \textbf{Feature inversion} concept inspired by GAN inversion. 
    \textbf{Right}: \textbf{Overview of the proposed InvAD framework} that consists of four components in tandem: 
    \textbf{\textit{1)}} Multi-scale features $\bm{F}^{I}$ extracted by image encoder $\deepred{\bm{\phi}^{I}}$ is aggregated into low-resolution $\bm{F}^{f}$ to avoid spatial consistency mapping; 
    \textbf{\textit{2)}} Re-scaling upsampler $\deepred{\bm{\phi}^{R}}$ obtains re-scaled features $\bm{F}^{R}$ by several \textit{Up-sampling Blocks} \begin{adjustbox}{valign=c}{\includegraphics[width=0.017\linewidth]{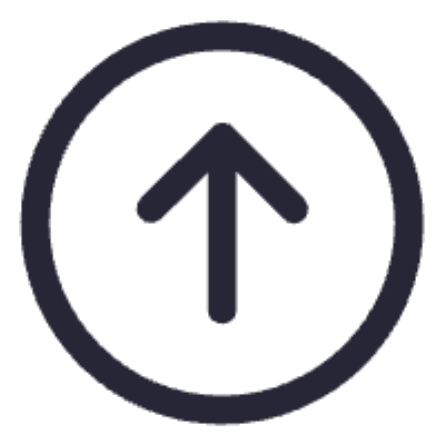}}\end{adjustbox}; 
    \textbf{\textit{3)}} An style translator $\deepred{\bm{\phi}^{S}}$, composed of \textit{Styling Block} \begin{adjustbox}{valign=c}{\includegraphics[width=0.017\linewidth]{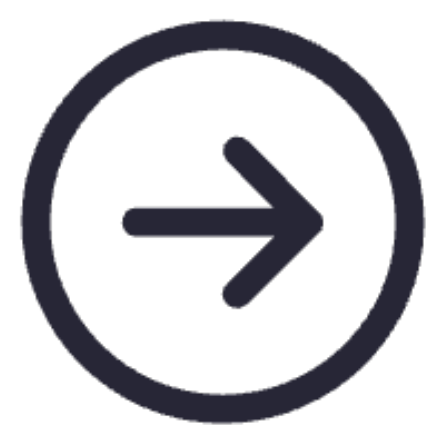}}\end{adjustbox}, to obtain style features $\bm{F}^{S}$ as the modulation control signal for the next stage; 
    \textbf{\textit{4)}} An feature decoder $\deepred{\bm{\phi}^{O}}$ to recover input features $\bm{F}^{O}$ for loss and anomaly map calculation by cascaded \textit{Space-aware Style Modulation} modules \begin{adjustbox}{valign=c}{\includegraphics[width=0.017\linewidth]{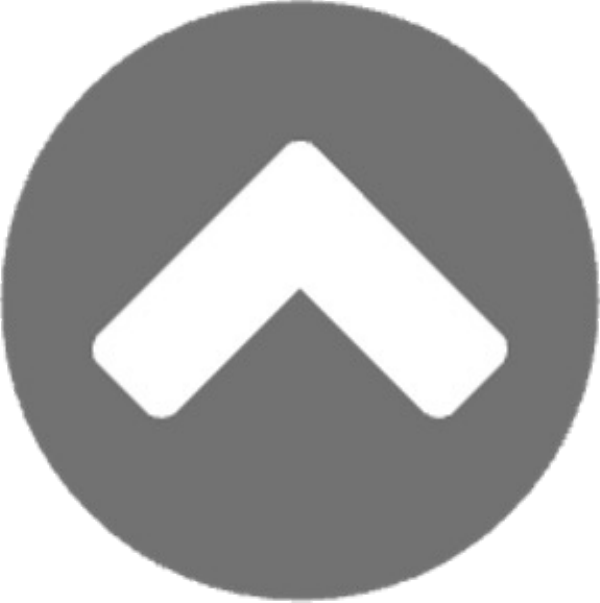}}\end{adjustbox}.
    }
    \label{fig:invad}
    \vspace{-1.5em}
\end{figure*}

\vspace{-1.5em}

\subsection{Comprehensive mM$^{.2}_{.8}$ and mIoU-max Metrics} \label{sec:method_metric}
As shown in \Tab\ref{tab:metric}, we categorize the current evaluation criteria into three types: image-level mAU-ROC~\cite{dae}/mAP~\cite{draem}/m$F_1$-max~\cite{visa}, region-level mAU-PRO~\cite{uninformedstudents}, and pixel-level mAU-ROC/mAP/,$F_1$-max. 

\noindent\textbf{Limitations of previous metrics.} 
Since above metrics are not specifically designed for the AD domain, they have certain limitations in evaluation.
As shown in the `Highlight' column of \Tab\ref{tab:metric}, threshold-independent mAU-ROC, mAP, and mAU-PRO disregard the absolute value distribution of prediction probabilities, which may cause accuracy fluctuations due to inappropriate threshold selection in practical applications, while m$F_1$-max can be affected by potential data imbalance. 
In addition, above metrics are excessively high (especially mAU-ROC) in current datasets like MVTec AD, but the quantitative results does not align with the visualization results (\cf, \Fig\ref{fig:teaser}-Middle), meaning that they cannot effectively distinguish the merits of different algorithms. 

\noindent\textbf{Advancements inspired by segmentation.}
Considering the higher practical value of pixel-level segmentation and evaluation, we further adapt segmentation metrics to the AD domain, proposing threshold-dependent m$F_1$$^{.2}_{.8}$, mAcc$^{.2}_{.8}$, and mIoU$^{.2}_{.8}$. 
0.2 is a high-confidence threshold often chosen in industry, so we chose a threshold range from 0.2 to 0.8 to align as closely as possible with actual applications. 
Specifically, these metrics are inspired by the binary classification definition of the AD task. 
\Fig\ref{fig:metric_28} analyzes the pixel-level anomaly score distribution with SoTA UniAD~\cite{uniad} on object Hazelnut and texture Toothbrush, observing that most anomaly and normal values fall around 0.5 and exhibit overlap. 
We argue that a good AD method should output values as close as possible to 0 (normal) or 1 (anomaly) with clear anomaly result judgment and less in intermediate states. Therefore, we calculate the average metric scores for thresholds ranging from 0.2 to 0.8 with a step of 0.1 as the comprehensive score, \ie, mM$^{.2}_{.8}$ (mM $\in$ m$F_1$, mAcc, and mIoU). 
Moreover, drawing from the more practical $F_1$-max metric design, we introduce the popular mIoU from segmentation into the AD domain, designing the mIoU-max metric to measure the overlap upper limit between different AD method results and ground truth. Considering the dazzling of excessive metrics, we take the average of image-level mAU-ROC/mAP/m$F_1$-max, pixel-level mAU-ROC/mAP/m$F_1$-max, and M$^{.2}_{.8}$ metrics as comprehensive metrics, respectively, marked in \redzjn{red} in \Tab\ref{tab:metric}. Time consumption (s) for testing various metrics on different datasets by AMD EPYC 7K62 CPU are as follows, and this demonstrates the high efficiency of mAD$^{.2}_{.8}$ and mIoU-max: 
\begin{center}
    \vspace{-1.5em}
    \tablestyle{3pt}{1.05}
    \renewcommand{\arraystretch}{1.0}
    \setlength\tabcolsep{3.0pt}
    \resizebox{1.0\linewidth}{!}{
        \begin{tabular}{p{1.8cm}<{\centering} p{1.0cm}<{\centering} p{1.0cm}<{\centering} p{1.0cm}<{\centering} p{1.8cm}<{\centering} p{1.8cm}<{\centering}}
        Dataset & mAD$_I$ & mAD$_P$ & \textbf{mAD$^{.2}_{.8}$} & mAU-PRO & \textbf{mIoU-max} \\
        \toprule[1.5pt]
        MVTec AD & 0.003 & 13.401 & \textbf{0.654} & 19.555 & \textbf{0.630} \\
        VisA & 0.005 & 16.788 & \textbf{0.841} & 24.195 & \textbf{0.795} \\
        COCO-AD & 0.005 & 374.877 & \textbf{20.435} & 589.95 & \textbf{18.99} \\
        \end{tabular}
    }
\end{center}
We suggest that subsequent works should continue to use comprehensive metrics for evaluation, which has important guiding significance for the practicality. Notably, these new metrics are applicable to single-class AD and they all in the range of [0, 100\%].

\noindent\textbf{Rationality explanation.} 
\Tab\ref{tab:pearson}-Right shows Pearson correlation coefficient between each pair of datasets with additional four new metrics. 
The correlation between the two industrial datasets slightly decreases, indicating that the new metrics are more challenging for evaluating industry-specific datasets. 
In addition, these metrics increase when relating more general COCO-AD, demonstrating their universality for practical application.

\section{Methodology: InvAD} \label{sec:invad}

\noindent\textbf{Motivation.} 
In essence, AD aims at overfitting to the training set with only normal data. 
Modeling under the multi-class setting is particularly challenging, while the effectiveness of reconstruction-based methods largely benefits from their reconstruction capabilities~\cite{survey_ad}. 
As shown in \Fig\ref{fig:invad}-Left, inspired by the high-quality image reconstruction ability of GAN inversion~\cite{gan_inversion_survey}, learning-based GAN inversion fixes the generator $\bm{G}$ to optimize the best parameters of the encoder $\bm{E}$:
\begin{equation}
    \vspace{-1.5em}
    \begin{aligned}
        \theta^*=\underset{\theta}{\operatorname{argmin}}(\bm{G}(\bm{E}(\boldsymbol{x} ; \theta)), \boldsymbol{x}).
    \end{aligned}
    \vspace{-1.5em}
\end{equation}
Our work extends this concept to feature-level reconstruction, fixing the encoder $\bm{E}$ to optimize the best parameters of the generator $\bm{G}$, and then locating the abnormal area through the positional difference between the original feature $\bm{F}$ and the generated feature $\bm{F}^*$:
\begin{equation}
    \vspace{-1.5em}
    \begin{aligned}
        \theta^*=\underset{\theta}{\operatorname{argmin}}(\bm{G}(\bm{E}(\boldsymbol{F}) ; \theta), \boldsymbol{F}).
    \end{aligned}
    \vspace{-1.5em}
\end{equation}
Generally, we follow the powerful StyleGAN~\cite{stylegan1} generator that is frequently employed in GAN-inversion structures and adapt it to InvAD's structural design. 

\noindent\textbf{Overall structure.} 
Inspired by the high-quality reconstruction capability of GAN inversion~\cite{gan_inversion_survey}, we design a novel feature inversion AD framework in \Fig\ref{fig:invad}-Right to perform high-quality feature reconstruction. Specifically, InvAD consists of four components in tandem: 
\noindent\textbf{\textit{1)}} Image encoder $\deepred{\bm{\phi}^{I}}$ performs multi-scale feature extraction on input image $\bm{I}$ to obtain $\bm{F}^{I}$, which goes through down-sampling, concatenation, and $\bm{N}_{B}$ BottleNecks~\cite{resnet} to get the fusion feature $\bm{F}^{f}$. Consistent with RD~\cite{rd}, WideResNet-50~\cite{wideresnet} is used by default, but combining powerful backbones~\cite{swin,eatformer,emo} can potentially improve model performance. 
\textbf{\textit{2)}} Re-scaling upsampler $\deepred{\bm{\phi}^{R}}$ leverages several up-sampling blocks to obtain $\bm{F}^{R}$, with each block containing an up-sampling convolution followed by $\bm{N}_{C}$ convolution modules. 
\textbf{\textit{3)}} Style translator $\deepred{\bm{\phi}^{S}}$ applies two convolution layers to $\bm{F}^{R}_{i}$ at each resolution to obtain the channel-shrunk style feature $\bm{F}^{S}_{i}$. 
\textbf{\textit{4)}} Feature decoder $\deepred{\bm{\phi}^{O}}$ uses $\bm{F}^{S}$ as the space-aware style condition to recover feature $\bm{F}^{O}$ from a constant feature $\bm{F}^{Const}$ through several stacked Space-aware Style Modulation (SSM) modules. Each stage simply uses $\bm{N}_{L}$ SSMs.

\noindent\textbf{SSM module.} 
To increase dynamic modeling capabilities, we introduce a modulation mechanism for high-precision feature reconstruction from constant query, which is significantly different from RD~\cite{rd} using static convolution to reconstruct target features from neck features. 
Specific structure is shown in \Fig\ref{fig:invad}-Right: $\bm{F}^{S}_{i}$ passes through two convolution layers with kernel 3 and stride 1 (k3s1) to obtains $\bm{F'}^{S}_{i}$ and $\bm{F''}^{S}_{i}$, which serves as the control signal for SSM$_{i,j}$: \\
\begin{equation}
    \begin{aligned}
        \bm{F}^{O}_{i,j+1} = \left(\bm{F'}^{S}_{i} \otimes (\bm{F}^{O}_{i,j}-\mu(\bm{F}^{O}_{i,j})) / \sigma(\bm{F}^{O}_{i,j})\right) \oplus \bm{F''}^{S}_{i}, 
    \end{aligned}
\end{equation}
where $i$ represents the stage, and $j \in [1, \bm{N}_{L}]$ represents the $j$-th SSM module under the $i$-th resolution. $\mu(\cdot)$ and $\sigma(\cdot)$ compute the mean and variance of a feature. Notably, this module also supports up-sampling operations labeled as k3s2 with kernel 3 and stride 2.

\noindent\textbf{Implementation details.} 
The channel numbers $\bm{N}_{R}$ and $\bm{N}_{S}$ of $\bm{F}^{R}$ and $\bm{F}^{S}$ at each stage are set to 64 and 16 by default, ensuring a relatively low increase in computational load. Cosine distance between $\bm{F}^{I}$ and $\bm{F}^{O}$ is viewed as anomaly map during inference~\cite{rd}. Overall, the entire InvAD only contains convolution operations and requires only MSE reconstruction constraints, making it simple, efficient, and effective. 
We provide InvAD with Wide-ResNet-50~\cite{wideresnet} as the default encoder model, and also offer a lighter InvAD-lite with ResNet-34~\cite{resnet}.

\noindent\textbf{InvAD Summary.} 
Starting from the basic RD~\cite{rd}, we gradually fuse each proposed component to build our InvAD step-by-step. 
\Tab\ref{tab:incremental} reveals the effectiveness of each design, and we obtain the powerful InvAD model when using all components. 

\begin{table}[tp]
    \caption{\textbf{Technical incremental trajectory of InvAD.} Each line modifies the immediately preceding line incrementally. Detailed ablations in \Sec\ref{exp:ablation}.}
    \vspace{-0.5em}
    \renewcommand{\arraystretch}{1.0}
    \setlength\tabcolsep{1.0pt}
    \resizebox{1.0\linewidth}{!}{
        \begin{tabular}{p{4.0cm}<{\raggedright} p{1.2cm}<{\centering} p{1.2cm}<{\centering} p{1.2cm}<{\centering} p{1.7cm}<{\centering} p{1.7cm}<{\centering}}
        \toprule[1.5pt]
        \pzo\pzo\pzo Method & mAD$_I$ & mAD$_P$ & mAD$^{.2}_{.8}$ & mAU-PRO & mIoU-max \\
        \hline
        Baseline (RD) & 95.4 &	66.2 & 27.4 & 91.1 & 37.3 \\
        \hline
        +~Use Skip AutoEncoder & 96.5 & 69.9 & 29.8 & 91.4 & 41.7 \\
        +~Use Rescaled Features $\deepred{\bm{\phi}^{R}}$ & 97.7 & 70.3 & 34.1 & 93.4 & 41.8 \\
        +~Use Style Features $\deepred{\bm{\phi}^{S}}$ & 98.3 & 71.0 & 34.4 & 93.8 & 42.5 \\
        +~Use SSM (InvAD) & 98.9 & 72.0 & 34.8 & 94.1 & 43.7 \\
        \bottomrule[1.5pt]
        \end{tabular}
    }
    \label{tab:incremental}
    \vspace{-2.0em}
\end{table}

%% file: secs/4_experiments.tex
\vspace{-1.0em}
\section{Experiments} \label{sec:exp}
\vspace{-0.5em}
\subsection{Setup for Multi-class  Unsupervised AD} \label{section:setup}
\vspace{-1.0em}
\noindent
\textbf{Datasets.}
This paper investigates the problem of anomaly classification and segmentation under a multi-class setting. In addition to evaluating the potential of different methods on the newly proposed general-purpose large-scale COCO-AD benchmark, we also conduct comparative experiments on MVTec AD~\cite{mvtec} and VisA~\cite{visa}. Considering computational costs, all ablation and analysis experiments are conducted on the popular MVTec AD by default. We also provide results on the MVTec3D~\cite{mvtec3d}, as well as the recently proposed Uni-Medical~\cite{vitad}, and Real-IAD~\cite{realiad} datasets.

\begin{table*}[htp]
    \centering
    \caption{\textbf{Different fine-grained multi-class anomaly classification and segmentation results with SoTAs on representative AD datasets.} $^*$: reproduce MUAD results. Prefix `m' stands for averaged metric results for all categories. Metrics marked in \redzjn{red} are recommended. \textbf{Bold} and \underline{underline} represent optimal and sub-optimal results, respectively. More results on MVTec 3D AD~\cite{mvtec3d}, Uni-Medical~\cite{vitad}, and Real-IAD~\cite{realiad} in Appendix~\blue{\textbf{D}}.\textbf{Abundant results for each category are in Appendix~\blue{\textbf{E~F~G~H~I~J}} for the restricted space.}}
    \vspace{-0.5em}
    \label{tab:sota}
    \renewcommand{\arraystretch}{0.8}
    \setlength\tabcolsep{6.0pt}
    \resizebox{1.0\linewidth}{!}{
        \begin{tabular}{p{0.5cm}<{\centering} p{2.3cm}<{\centering} p{2.0cm}<{\centering} p{2.0cm}<{\centering} p{2.0cm}<{\centering}p{2.0cm}<{\centering} p{2.0cm}<{\centering} p{2.0cm}<{\centering} p{2.0cm}<{\centering} p{2.0cm}<{\centering} p{2.0cm}<{\centering}}
            \toprule[0.17em]
            & Fineness-level & Metric & DRÆM$^*$~\cite{draem} & RD$^*$~\cite{rd} & UniAD$^\dagger$~\cite{uniad} & DeSTSeg$^*$~\cite{destseg} & SimpleNet$^*$~\cite{simplenet} & DiAD~\cite{diad} & InvAD-lite & InvAD \\
            \hline
            \multirow{14}{*}{\makecell[c]{\rotatebox{90}{\textbf{COCO-AD}}}} & \multirow{3}{*}{\makecell[c]{Image-level\\ (Classification)}} & mAU-ROC & 55.1 & 58.4 & 56.2 & 56.2 & 57.1 & 59.0 & \underline{64.7} & \textbf{65.9} \\ 
            & & mAP & 48.9 & 51.1 & 49.0 & 50.3 & 49.4 & 53.0 & \underline{56.7} & \textbf{57.8} \\ 
            & & m$F_1$-max & 62.5 & 62.3 & 61.7 & 61.9 & 61.7 & 63.2 & \underline{63.5} & \textbf{64.1} \\  
            \cline{2-11}
            & Region-level & \redzjn{mAU-PRO} & 11.8 & \underline{40.7} & 31.7 & 23.6 & 27.5 & 30.8 & 38.2 & \textbf{44.9} \\  
            \cline{2-11}
            & \multirow{3}{*}{\makecell[c]{Pixel-level\\ (Segmentation)}} & mAU-ROC & 51.8 & 67.4 & 65.4 & 60.9 & 59.5 & 68.1 & \underline{70.6} & \textbf{73.3} \\ 
            & & mAP & \pzo8.4 & 14.3 & 12.9 & 11.3 & 12.2 & \textbf{20.5} & 18.4 & \underline{19.7} \\ 
            & & m$F_1$-max & 14.5 & 20.6 & 19.4 & 16.3 & 17.9 & 14.2 & \underline{23.4} & \textbf{25.4} \\ 
            \cline{3-11}
            & \multirow{4}{*}{\makecell[c]{Proposed}} & m$F_1$$^{.2}_{.8}$ & \pzo7.9 & \pzo9.9 & \pzo6.6 & \pzo2.4 & \pzo8.5 & \pzo9.6 & \underline{11.2} & \textbf{12.4} \\ 
            & & mAcc$^{.2}_{.8}$ & 24.4 & 33.8 & 26.4 & \pzo1.5 & 32.3 & 31.1 & \underline{34.2} & \textbf{37.5} \\ 
            & & mIoU$^{.2}_{.8}$ & \pzo5.6 & \pzo5.5 & \pzo3.7 & \pzo1.2 & \pzo4.7 & \pzo6.1 & \pzo\underline{6.4} & \pzo\textbf{7.1} \\ 
            & & \redzjn{mIoU-max} & \pzo8.0 & 11.9 & 11.1 & \pzo9.0 & 10.2 & 11.6 & \underline{13.8} & \textbf{15.2} \\ 
            \cline{2-11}
            & \multirow{3}{*}{\makecell[c]{Averaged \\ Metrics}} & \redzjn{mAD$_I$} & 55.5 & 57.3 & 55.6 & 56.1 & 56.0 & 58.4 & \underline{61.6} & \textbf{62.6} \\ 
            & & \redzjn{mAD$_P$} & 24.9 & 34.1 & 32.6 & 29.5 & 29.9 & 34.2 & \underline{37.4} & \textbf{39.5} \\ 
            & & \redzjn{mAD$^{.2}_{.8}$} & 12.6 & 16.4 & 12.2 & \pzo1.7 & 15.2 & 15.6 & \underline{17.3} & \textbf{19.0} \\ 
            
            \hline
            \multirow{14}{*}{\makecell[c]{\rotatebox{90}{\textbf{MVTec AD}~\cite{mvtec}}}} & \multirow{3}{*}{\makecell[c]{Image-level\\ (Classification)}} & mAU-ROC & 88.8 & 94.6 & 97.5 & 89.2 & 95.3 & 97.2 & \underline{98.2} & \textbf{98.9} \\ 
            & & mAP & 94.7 & 96.5 & 99.1 & 95.5 & 98.4 & 99.0 & \underline{99.2} & \textbf{99.6} \\ 
            & & m$F_1$-max & 92.0 & 95.2 & \underline{97.3} & 91.6 & 95.8 & 96.5 & 97.2 & \textbf{98.1} \\  
            \cline{2-11}
            & Region-level & \redzjn{mAU-PRO} & 71.1 & 91.1 & 90.7 & 64.8 & 86.5 & 90.7 & \underline{92.7} & \textbf{94.1} \\  
            \cline{2-11}
            & \multirow{3}{*}{\makecell[c]{Pixel-level\\ (Segmentation)}} & mAU-ROC & 88.6 & 96.1 & 97.0 & 93.1 & 96.9 & 96.8 & \underline{97.4} & \textbf{98.2} \\  
            & & mAP & 52.6 & 48.6 & 45.1 & 54.3 & 45.9 & 52.6 & \underline{55.0} & \textbf{57.6} \\ 
            & & m$F_1$-max & 48.6 & 53.8 & 50.3 & 50.9 & 49.7 & 55.5 & \underline{58.1} & \textbf{60.1} \\ 
            \cline{3-11}
            & \multirow{4}{*}{\makecell[c]{Proposed}} & m$F_1$$^{.2}_{.8}$ & 21.8 & 25.8 & 22.4 & 29.7 & 25.3 & 19.5 & \underline{32.6} & \textbf{34.6} \\ 
            & & mAcc$^{.2}_{.8}$ & 15.3 & 39.8 & 37.5 & 22.7 & \textbf{47.7} & 40.7 & \underline{47.1} & 46.9 \\ 
            & & mIoU$^{.2}_{.8}$ & 14.2 & 16.4 & 13.9 & 18.8 & 16.0 & 12.0 & \underline{21.3} & \textbf{23.0} \\ 
            & & \redzjn{mIoU-max} & 35.1 & 37.3 & 34.2 & 35.3 & 34.4 & 21.3 & \underline{41.7} & \textbf{43.7} \\ 
            \cline{2-11}
            & \multirow{3}{*}{\makecell[c]{Averaged \\ Metrics}} & \redzjn{mAD$_I$} & 91.8 & 95.4 & 98.0 & 92.1 & 96.5 & 97.6 & \underline{98.2} & \textbf{98.9} \\ 
            & & \redzjn{mAD$_P$} & 63.3 & 66.2 & 64.1 & 66.1 & 64.2 & 68.3 & \underline{70.2} & \textbf{72.0} \\ 
            & & \redzjn{mAD$^{.2}_{.8}$} & 17.1 & 27.4 & 24.6 & 23.7 & 29.7 & 24.1 & \underline{33.7} & \textbf{34.8} \\ 

            \hline
            \multirow{14}{*}{\makecell[c]{\rotatebox{90}{\textbf{VisA}~\cite{visa}}}} & \multirow{3}{*}{\makecell[c]{Image-level\\ (Classification)}} & mAU-ROC & 79.5 & 92.4 & 88.8 & 88.9 & 87.2 & 86.8 & \underline{94.9} & \textbf{95.5} \\ 
            & & mAP & 82.8 & 92.4 & 90.8 & 89.0 & 87.0 & 88.3 & \underline{95.2} & \textbf{95.8} \\ 
            & & m$F_1$-max & 79.4 & 89.6 & 85.8 & 85.2 & 81.7 & 85.1 & \underline{90.8} & \textbf{92.1} \\ 
            \cline{2-11}
            & Region-level & \redzjn{mAU-PRO} & 59.1 & \underline{91.8} & 85.5 & 67.4 & 81.4 & 75.2 & \textbf{92.5} & \textbf{92.5} \\  
            \cline{2-11}
            & \multirow{3}{*}{\makecell[c]{Pixel-level\\ (Segmentation)}} & mAU-ROC & 91.4 & 98.1 & 98.3 & 96.1 & 96.8 & 96.0 & \underline{98.6} & \textbf{98.9} \\ 
            & & mAP & 24.8 & 38.0 & 33.7 & 39.6 & 34.7 & 26.1 & \underline{40.3} & \textbf{43.1} \\ 
            & & m$F_1$-max & 30.4 & 42.6 & 39.0 & 43.4 & 37.8 & 33.0 & \underline{44.3} & \textbf{47.0} \\ 
            \cline{3-11}
            & \multirow{4}{*}{\makecell[c]{Proposed}} & m$F_1$$^{.2}_{.8}$ & 12.6 & 21.2 & 17.9 & \underline{27.4} & 17.5 & 13.2 & 25.8 & \textbf{28.0} \\ 
            & & mAcc$^{.2}_{.8}$ & \pzo8.7 & 46.5 & \underline{47.1} & 41.0 & \textbf{50.6} & 46.2 & 44.2 & 45.6 \\ 
            & & mIoU$^{.2}_{.8}$ & \pzo7.4 & 13.1 & 10.9 & \underline{17.3} & 11.0 & \pzo8.0 & 16.1 & \textbf{17.9} \\ 
            & & \redzjn{mIoU-max} & 18.8 & 28.5 & 25.7 & 26.8 & 25.9 & 16.2 & \underline{30.0} & \textbf{32.7} \\ 
            \cline{2-11}
            & \multirow{3}{*}{\makecell[c]{Averaged \\ Metrics}} & \redzjn{mAD$_I$} & 80.6 & 91.5 & 88.5 & 87.7 & 85.3 & 86.7 & \underline{93.6} & \textbf{94.5} \\ 
            & & \redzjn{mAD$_P$} & 48.9 & 59.6 & 57.0 & 59.7 & 56.4 & 51.7 & \underline{61.1} & \textbf{63.0} \\ 
            & & \redzjn{mAD$^{.2}_{.8}$} & \pzo9.6 & 26.9 & 25.3 & 28.6 & 26.4 & 22.5 & \underline{28.7} & \textbf{30.5} \\ 
            \toprule[0.12em] 
        \end{tabular}
    }
    \vspace{-1.5em}
\end{table*}

\noindent
\textbf{Metrics.} 
This paper suggests using image-level AU-ROC~\cite{dae}/AP~\cite{draem}/$F_1$-max~\cite{visa}, region-level AU-PRO~\cite{uninformedstudents}, and pixel-level AU-ROC/AP/$F_1$-max to evaluate the results of different methods comprehensively. Furthermore, we evaluate the newly proposed metrics (\Sec\ref{sec:method_metric}) and standardized several average AD metrics for comprehensive comparison.

\noindent\textbf{Comparison Methods.} 
We reproduce mainstream models under the MUAD setting for fair comparison: augmentation-based DRÆM~\cite{draem}, embedding-based SimpleNet~\cite{simplenet}/DeSTSeg~\cite{destseg}, and reconstruction-based RD~\cite{rd}/UniAD~\cite{uniad}/DiAD~\cite{diad}. Note that all methods are re-trained under 256$\times$256 resolution. 

\noindent\textbf{Training Details.} 
InvAD is trained under 256 $\times$ 256 resolution with only MSE reconstruction loss, without any extra datasets and training tricks/augmentations for all experiments. AdamW optimizer~\cite{adamw} is used with an initial learning rate of 1e$^{-4}$, a weight decay of 1e$^{-4}$, and a batch size of 32. Our model is trained for 300 epochs by a single V100 GPU on MVTec AD and VisA datasets, and the learning rate drops by 0.1 after 80 percent epochs. For the COCO-AD dataset, all methods are trained fairly for 100 epochs.

\vspace{-1.5em}
\subsection{Results on COCO-AD, MVTec AD, and VisA} \label{exp:comparison}
\vspace{-1.0em}
\noindent\textbf{Quantitative results.}
\Tab\ref{tab:sota} presents a quantitative comparison of our InvAD with SoTA methods on the COCO-AD (complex general scenarios), MVTec AD (specific industrial scenarios), and VisA (small defects) datasets regarding comprehensive metrics.
Overall, our InvAD consistently achieves significantly higher results on all metrics, validating the effectiveness of the proposed feature inversion structure. 
Additionally, we discover several interesting findings through the results: 
\textit{i)} Although the image-level metrics of MVTec are nearly saturated, they still hold research value for the region-/pixel-level results, especially on the proposed more practical mIoU-max. 
\textit{ii)} As the first general-purpose dataset, the highly challenging COCO-AD has lower results on all metrics, serving as a standard benchmark to promote the continuous development of AD algorithms. 
\textit{iii)} Different types of existing algorithms tend to a certain dataset, but the reconstruction-based RD performs relatively stable. \\

\begin{figure}[tp]
    \centering
    \includegraphics[width=1.0\linewidth]{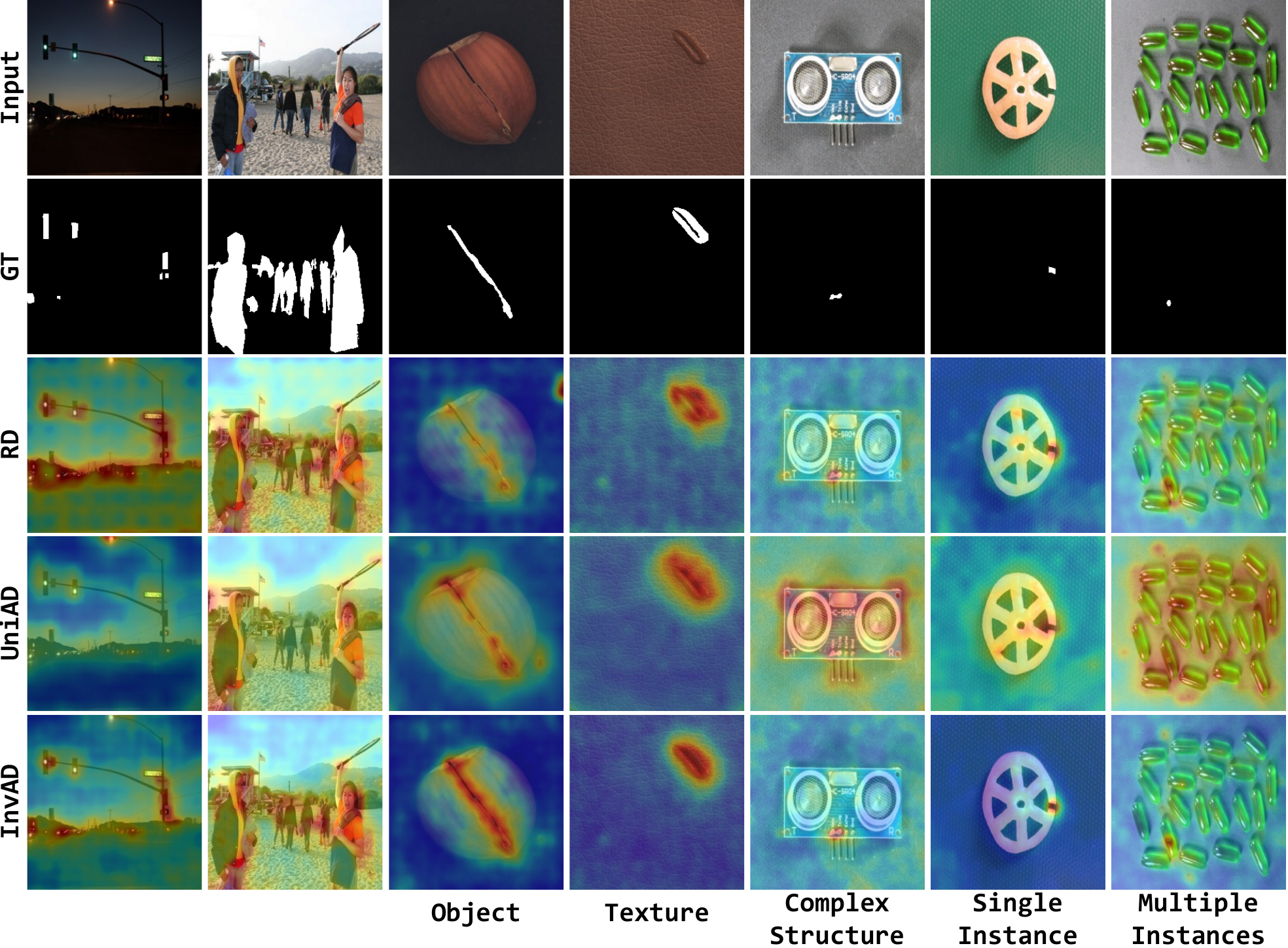}
    \caption{Visual comparison of anomaly maps for different types of objects on COCO-AD (Left), MVTec AD (Middle), and VisA (Right) datasets. Our InvAD locates more \textit{compact} segmentation results with ground truths while having a smaller response in normal areas.}
    \label{fig:qualitative_comparison}
\end{figure}

\noindent\textbf{Qualitative results.}
\Fig\ref{fig:qualitative_comparison} further provides a visual comparison with RD and UniAD on three datasets, selecting one object from each category. Comparatively, InvAD produces more compact anomaly localization while having a smaller feature response in normal areas.

\vspace{-1.0em}
\subsection{Ablation and Analysis} \label{exp:ablation}
\vspace{-0.5em}

\Sec\ref{exp:comparison} demonstrates the consistent superiority of our method across three datasets, so all subsequent ablation and analysis experiments are conducted on the MVTec AD dataset by default to save computational resources.
\noindent \textbf{\textit{1)} Channel numbers of $\bm{F}^{R}$ and $\bm{F}^{S}$.} The upper part of \Tab\ref{tab:ablation_architecture} shows the results under different configurations of the channel numbers $\bm{N}_{R}$ and $\bm{N}_{S}$ for $\bm{F}^{R}$ and $\bm{F}^{S}$. Results indicate that fewer channels are sufficient to achieve satisfactory results. In contrast, an excessive number of channels does not significantly improve the results and instead incurs additional computational costs. \\
\textbf{\textit{2)} Stack numbers of different sub-modules.} The lower part of \Tab\ref{tab:ablation_architecture} shows the impact of different sub-module stacking layers in InvAD on the results, indicating that a configuration of 1/1/2 is sufficient to achieve satisfactory results, and more layers may adversely affect the model performance. \\
\textbf{\textit{3)} Threshold ranges and step sizes of M$^{.2}_{.8}$ metrics.} As shown in \Tab\ref{tab:ablation_28}, more refined steps will produce more accurate results, but the step of 0.1 is enough while reasonably reducing the evaluation duration. Taking InvAD as an example, more stringent (0.1-0.9) or relaxed (0.4-0.6) threshold ranges would change the results. Considering that 0.2 is a high-confidence threshold often chosen in industry, we chose a threshold in range [0.2, 0.8] to align with actual applications. \\
\textbf{\textit{4)} Ablations on \textbf{Style Features} $\deepred{\bm{\phi}^{S}}$ and \textbf{SSM}.} We remove $\deepred{\bm{\phi}^{S}}$ and replace \textbf{SSM} with cascade concatenation and convolution operations in \Tab\ref{tab:ablation_component}. Our InvAD still achieves competitive results even without two modules, and each module contributes performance. \\
\textbf{\textit{5)} Time and memory consumption.} We further simplify the hyper-parameter and backbone to ResNet34 to obtain InvAD-lite. While significantly improving efficiency, it achieves \textit{\textbf{significant}} better results over state-of-the-art methods in \Tab\ref{tab:ablation_efficiency}. 
More ablation can be viewed in Appendix~\blue{\textbf{B}}.

\begin{table}[tb]
    \centering
    \caption{Ablation on architecture: channel numbers of $\bm{N}_{R}$ and $\bm{N}_{S}$, and stack numbers $\bm{N}_{B}$, $\bm{N}_{C}$, and $\bm{N}_{L}$.}
    \vspace{-0.5em}
    \label{tab:ablation_architecture}
    \renewcommand{\arraystretch}{0.9}
    \setlength\tabcolsep{1.0pt}
    \resizebox{1.0\linewidth}{!}{
        \begin{tabular}{p{0.3cm}<{\centering} p{1.0cm}<{\centering} p{1.0cm}<{\centering} p{1.0cm}<{\centering} p{1.7cm}<{\centering} p{1.7cm}<{\centering} p{1.7cm}<{\centering} p{1.7cm}<{\centering} p{1.7cm}<{\centering}}
        \toprule[1.5pt]
        & & $\bm{N}_{R}$ & $\bm{N}_{S}$ & mAD$_I$ & mAD$_P$ & mAD$^{.2}_{.8}$ & mAU-PRO & mIoU-max \\
        \hline
        \multirow{9}{*}{\makecell[c]{\rotatebox{90}{Channel Configure}}} & & 16 & 8 & 98.3 & 71.1 & 35.0 & 94.0 & 42.7 \\
        & & 16 & 16 & 98.5 & 71.6 & 35.1 & 94.1 & 43.4 \\
        & & 32 & 8 & 98.7 & 71.1 & 34.8 & 94.0 & 42.8 \\
        & & 32 & 16 & 98.7 & 71.5 & 36.0 & 94.0 & 43.0 \\
        & \cellcolor{lblu_tab}{} & \cellcolor{lblu_tab}{64} & \cellcolor{lblu_tab}{16} & \cellcolor{lblu_tab}{98.9} & \cellcolor{lblu_tab}{72.0} & \cellcolor{lblu_tab}{34.8} & \cellcolor{lblu_tab}{94.1} & \cellcolor{lblu_tab}{43.7} \\
        & & 64 & 32 & 98.9 & 71.7 & 35.6 & 93.8 & 43.1 \\
        & & 64 & 64 & 98.8 & 71.7 & 35.4 & 94.1 & 43.3 \\
        & & 128 & 32 & 98.8 & 71.6 & 34.2 & 94.2 & 43.5 \\
        & & 256 & 64 & 99.0 & 71.5 & 34.0 & 93.9 & 42.9 \\
        \toprule[0.8pt]
        & $\bm{N}_{B}$ & $\bm{N}_{C}$ & $\bm{N}_{L}$ & mAD$_I$ & mAD$_P$ & mAD$^{.2}_{.8}$ & mAU-PRO & mIoU-max \\
        \hline
        \multirow{9}{*}{\makecell[c]{\rotatebox{90}{Stack Number}}} & 0 & 0 & 1 & 98.5 & 71.1 & 33.0 & 94.0 & 42.9 \\
        & 0 & 0 & 2 & 98.7 & 71.5 & 34.4 & 93.9 & 43.0 \\
        & 0 & 1 & 1 & 98.7 & 71.6 & 34.1 & 94.2 & 43.4 \\
        & 1 & 0 & 1 & 98.4 & 71.2 & 33.4 & 94.1 & 42.9 \\
        & \cellcolor{lblu_tab}{1} & \cellcolor{lblu_tab}{1} & \cellcolor{lblu_tab}{2} & \cellcolor{lblu_tab}{98.9} & \cellcolor{lblu_tab}{72.0} & \cellcolor{lblu_tab}{34.8} & \cellcolor{lblu_tab}{94.1} & \cellcolor{lblu_tab}{43.7} \\
        & 1 & 1 & 4 & 98.8 & 71.7 & 35.9 & 94.1 & 43.3 \\
        & 1 & 3 & 2 & 98.5 & 71.9 & 35.1 & 94.1 & 43.8 \\
        & 3 & 1 & 2 & 98.7 & 71.8 & 35.5 & 94.1 & 43.3 \\
        & 3 & 3 & 4 & 98.6 & 71.0 & 35.0 & 93.6 & 42.7 \\
        \bottomrule[1.5pt]
        \end{tabular}
    }
    \vspace{-1.0em}
\end{table}

\begin{table}[tb]
    \centering
    \caption{Ablation experiments for M$^{.2}_{.8}$ metrics in terms of start/end thresholds $\theta_{start}$/$\theta_{end}$ and step.}
    \vspace{-0.5em}
    \label{tab:ablation_28}
    \renewcommand{\arraystretch}{0.9}
    \setlength\tabcolsep{3.0pt}
        \resizebox{1.0\linewidth}{!}{
            \begin{tabular}{p{0.5cm}<{\centering} p{0.8cm}<{\centering} p{0.8cm}<{\centering} p{1.0cm}<{\raggedright} p{1.7cm}<{\centering} p{1.7cm}<{\centering} p{1.7cm}<{\centering} p{1.7cm}<{\centering}}
            \toprule[1.5pt]
            No. & $\theta_{start}$ & $\theta_{end}$ & Step & m$F_1$-max$^{.2}_{.8}$ & mAcc$^{.2}_{.8}$ & mIoU$^{.2}_{.8}$ & Time (s) \\
            \hline
            (1) & 0.2 & 0.8 & 0.01 & 37.02 & 46.46 & 24.73 & 115.74 \\
            (2) & 0.2 & 0.8 & 0.05 & 35.93 & 46.67 & 23.94 & \pzo24.84 \\
            \rowcolor{lblu_tab} (3) & 0.2 & 0.8 & 0.1 & 34.58 & 46.91 & 22.95 & \pzo13.59 \\
            \hline
            (4) & 0.1 & 0.9 & 0.1 & 28.54 & 47.65 & 18.76 & \pzo16.96 \\
            \rowcolor{lblu_tab} (5) & 0.2 & 0.8 & 0.1 & 34.58 & 46.91 & 22.95 & \pzo13.59 \\
            (6) & 0.3 & 0.7 & 0.1 & 40.09 & 45.79 & 26.96 & \pzo\pzo9.54 \\
            (7) & 0.4 & 0.6 & 0.1 & 44.06 & 44.71 & 29.78 & \pzo\pzo5.63 \\
            \bottomrule[1.5pt]
            \end{tabular}
        }
    \vspace{-1.0em}
\end{table}

\begin{table}[tb]
    \centering
    \caption{Ablation for Style Features $\deepred{\bm{\phi}^{S}}$ and SSM.}
    \vspace{-0.5em}
    \label{tab:ablation_component}
    \renewcommand{\arraystretch}{0.9}
        \setlength\tabcolsep{1.0pt}
        \resizebox{1.0\linewidth}{!}{
            \begin{tabular}{p{1.0cm}<{\centering} p{1.0cm}<{\centering} p{1.7cm}<{\centering} p{1.7cm}<{\centering} p{1.7cm}<{\centering} p{1.7cm}<{\centering} p{1.7cm}<{\centering}}
            \toprule[1.5pt]
            $\deepred{\bm{\phi}^{S}}$ & SSM & mAD$_I$ & mAD$_P$ & mAD$^{.2}_{.8}$ & mAU-PRO & mIoU-max \\
            \hline
            \xmark & \xmark & 97.7 & 70.3 & 34.1 & 93.4 & 41.8 \\
            \cmark & \xmark & 98.3 & 71.0 & 34.4 & 93.8 & 42.5 \\
            \xmark & \cmark & 98.5 & 71.3 & 34.5 & 93.9 & 43.0 \\
            \cmark & \cmark & 98.9 & 72.0 & 34.8 & 94.1 & 43.7 \\
            \bottomrule[1.5pt]
            \end{tabular}
        }
    \vspace{-1.0em}
\end{table}

\begin{table}[tb]
    \centering
    \caption{Efficiency analysis over SoTAs on MVTec AD.}
    \vspace{-0.5em}
    \label{tab:ablation_efficiency}
    \renewcommand{\arraystretch}{0.9}
        \setlength\tabcolsep{3.0pt}
        \resizebox{1.0\linewidth}{!}{
            \begin{tabular}{p{2.3cm}<{\centering} p{1.2cm}<{\centering} p{1.0cm}<{\centering} p{0.1cm}<{\centering} p{1.2cm}<{\centering} p{1.0cm}<{\centering} p{1.2cm}<{\centering} p{1.2cm}<{\centering} p{1.2cm}<{\centering}}
            \toprule[1.5pt]
            \multirow{2}{*}{\makecell[c]{Method}} & \multicolumn{2}{c}{Train} & & \multicolumn{2}{c}{Test} & \multirow{2}{*}{\makecell[c]{\#Params}} & \multirow{2}{*}{\makecell[c]{FLOPs}} & \multirow{2}{*}{\makecell[c]{FPS}} \\
            \cline{2-3} \cline{5-6}
            & Memory & Time & & Memory & Time & & & \\
            \hline
            DRÆM~\cite{draem} & 32,192M & 17.2H & & \pzo5,990M & \pzo\pzo31.9s & \pzo\pzo97.4M & 198.0G & \pzo54.0 \\
            RD~\cite{rd} & \pzo9,008M & \pzo4.3H & & \pzo3,186M & \pzo\pzo19.0s & \pzo\pzo80.6M & 28.4G & \pzo90.6 \\
            UniAD~\cite{uniad} & \pzo6,164M & 12.8H & & \pzo2,920M & \pzo\pzo30.4s & \pzo\pzo24.5M & \pzo3.4G & \pzo56.8 \\
            DeSTSeg~\cite{destseg} & 12,698M & \pzo5.1H & & \pzo4,852M & \pzo\pzo23.1s & \pzo\pzo35.2M & 30.7G & \pzo74.7 \\
            SimpleNet~\cite{simplenet} & \pzo6,154M & \pzo8.4H & & \pzo3,123M & \pzo\pzo35.0s & \pzo\pzo72.8M & 17.7G & \pzo49.3 \\
            DiAD~\cite{diad} & 21,772M & 45.1H & & 13,502M & 1552.5s & 1331.0M & 451.5G & \pzo\pzo0.5  \\
            \rowcolor{lblu_tab} InvAD-lite & \pzo5,050M & \pzo4.2H & & \pzo2,738M & \pzo\pzo17.0s & \pzo\pzo17.2M & \pzo9.3G & 101.6 \\
            \rowcolor{lblu_tab} InvAD & 16,280M & \pzo5.4H & & \pzo9,286M & \pzo\pzo20.7s & \pzo\pzo95.6M & 45.4G & \pzo\underline{83.5} \\
            \bottomrule[1.5pt]
            \end{tabular}
        }
    \vspace{-1.0em}
\end{table}

%% file: secs/5_conclusion.tex
\vspace{-1.0em}
\section{Conclusion} \label{sec:con}
\vspace{-0.5em}
This work rethinks the multi-class AD task from its three essential elements. 
Firstly, we construct a large-scale and challenging general-purpose COCO-AD dataset, alleviating the current issues of small dataset size and low scenario complexity. 
Secondly, we propose four practical threshold-dependent AD-specific metrics for comprehensive evaluation. 
Finally, we propose a powerful InvAD framework to achieve high-quality feature reconstruction, greatly enhancing the effectiveness of reconstruction-based methods on several AD datasets. We hope this work can inspire future research.

%% file: secs/6_appendix.tex
\renewcommand\thefigure{A\arabic{figure}}
\renewcommand\thetable{A\arabic{table}}  
\renewcommand\theequation{A\arabic{equation}}
\setcounter{equation}{0}
\setcounter{table}{0}
\setcounter{figure}{0}
\appendix

\section*{Overview}
The supplementary material presents the following sections to strengthen the main manuscript:
\begin{itemize}
    \item \textbf{\Sec\ref{sec:app_statistic_cocoad}} shows the defect distribution statistics for the proposed general-purpose COCO-AD dataset.
    \item \textbf{\Sec\ref{sec:app_ablation}} shows extended ablations for hyper-parameters.
    \item \textbf{\Sec\ref{sec:app_cocoad_vis}} shows more qualitative visualization of COCO-AD dataset.
    \item \textbf{\Sec\ref{sec:app_comprehensive}} shows more comprehensive results on MVTec 3D AD~\cite{mvtec3d}, Uni-Medical~\cite{vitad}, and Real-IAD~\cite{realiad} datasets.
    \item \textbf{\Sec\ref{sec:app_cocoad}} shows more quantitative results for each split on COCO-AD dataset.
    \item \textbf{\Sec\ref{sec:app_mvtec}} shows more quantitative results for each category on MVTec AD dataset under the MUAD setting.
    \item \textbf{\Sec\ref{sec:app_visa}} shows more quantitative results for each category on VisA dataset under the MUAD setting.
    \item \textbf{\Sec\ref{sec:app_mvtec3d}} shows more quantitative results for each category on MVTec 3D AD dataset under the MUAD setting.
    \item \textbf{\Sec\ref{sec:app_unimedical}} shows more quantitative results for each category on Uni-Medical dataset under the MUAD setting.
    \item \textbf{\Sec\ref{sec:app_realiad}} shows more quantitative results for each category on Real-IAD dataset under the MUAD setting. All five view images are used. 
    \item \textbf{We have attached the full codes and model, including the generation script for the COCO-AD dataset, as well as the InvAD model and log files.}
\end{itemize}

\begin{figure*}[tp]
    \centering
    \includegraphics[width=0.9\linewidth]{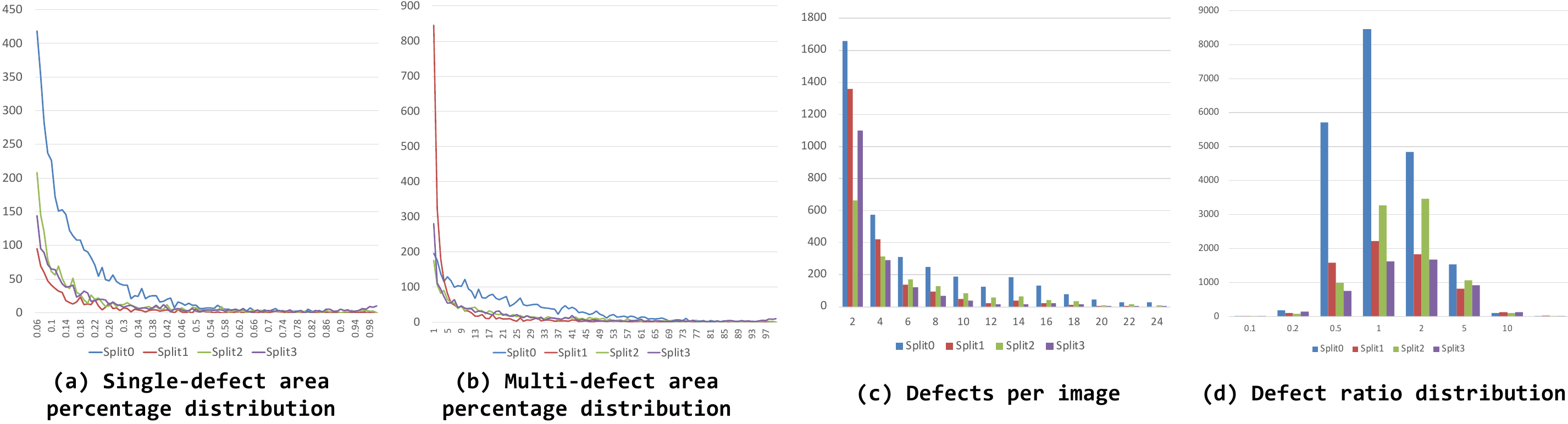}
    \caption{Statistics of the general-purpose COCO-AD dataset.}
    \label{fig:statistic}
    \vspace{-1.5em}
\end{figure*}

\section{Statistic Information of COCO-AD} \label{sec:app_statistic_cocoad}
\Fig\ref{fig:statistic} provides a statistical analysis of the four splits of the COCO-AD dataset across four dimensions to understand its characteristics for the AD task comprehensively. 
(a) The area proportion of a single object region mainly falls within 2\%, with about 90\% anomalous areas falling within a 10\% proportion. And there are individual data points where the proportion can exceed 50\%. 
(b) A single image may contain multiple defect regions. The distribution trend of each split is similar to (a), but there are slight differences between different splits. 
(c) The number of objects (defects) in an image is predominantly within two, but a certain proportion can reach more than 20. 
(d) The ratio of the anomalous region is mostly between 0.5 and 2, but there are certain extreme ratio data, such as those exceeding 10. 
This illustrates the highly complexity of the COCO-AD dataset in the anomalous regions, which is able to serving as a challenging dataset for the AD field.

\begin{figure}[htb]
    \centering
    \includegraphics[width=0.8\linewidth]{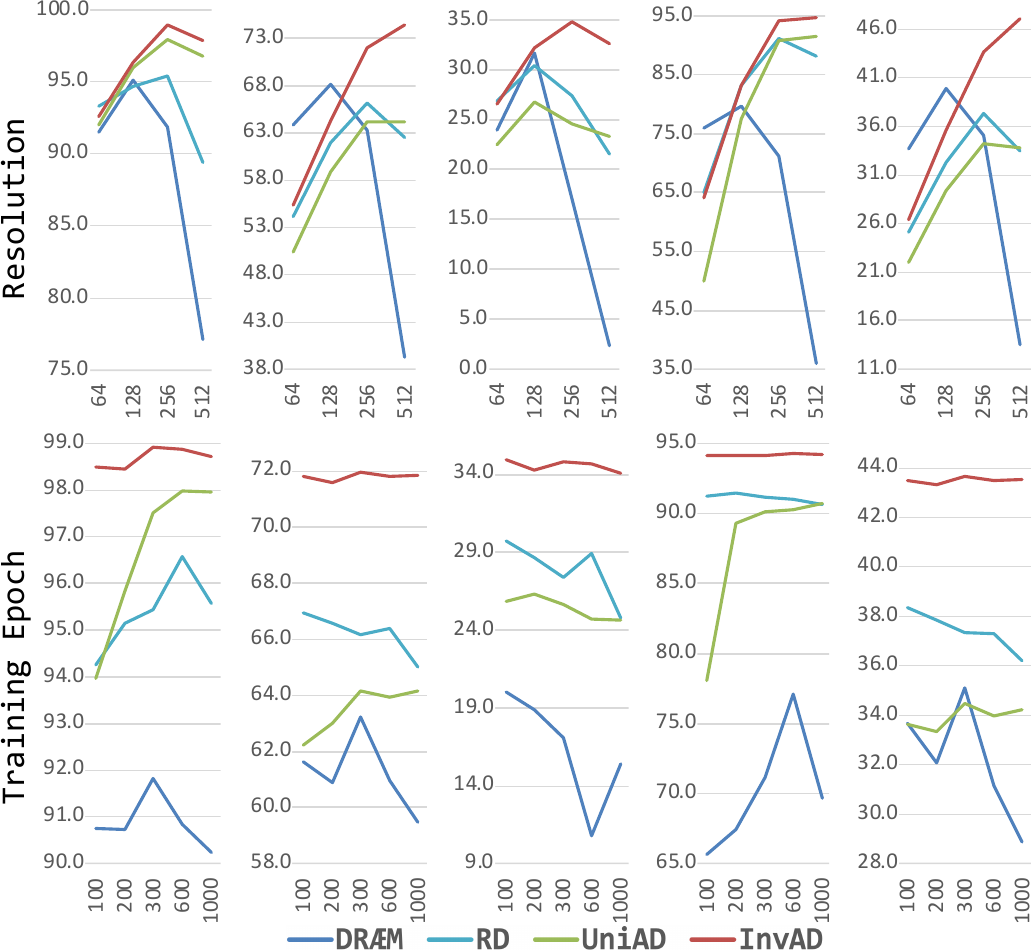}
    \vspace{-1.0em}
    \caption{Ablation of image resolution and training epoch on mAD$_I$, mAD$_P$, mAD$^{.2}_{.8}$, mAU-PRO, and mIoU-max.}
    \label{fig:ablation_resolution_epoch}
    \vspace{-1.0em}
\end{figure}

\begin{table}[htb]
    \centering
    \caption{Ablation for loss function and scheduler.}
    \vspace{-0.5em}
    \label{tab:supp_ablation_loss_scheduler}
    \renewcommand{\arraystretch}{0.9}
    \setlength\tabcolsep{1.0pt}
    \resizebox{1.0\linewidth}{!}{
        \begin{tabular}{p{0.3cm}<{\centering} p{2.0cm}<{\centering} p{1.7cm}<{\centering} p{1.7cm}<{\centering} p{1.7cm}<{\centering} p{1.7cm}<{\centering} p{1.7cm}<{\centering}}
        \toprule[1.5pt]
        & Item & mAD$_I$ & mAD$_P$ & mAD$^{.2}_{.8}$ & mAU-PRO & mIoU-max \\
        \toprule[1.5pt]
        \multirow{4}{*}{\makecell[c]{\rotatebox{90}{Loss}}} & Cos$_f$ & 98.9 & 72.0 & 34.6 & 94.0 & 43.8 \\
        & Cos$_p$ & 98.9 & 71.6 & 34.0 & 94.0 & 43.0 \\
        & L1 & 98.0 & 70.5 & 33.9 & 93.9 & 41.9 \\
        & \cellcolor{lblu_tab}{MSE} & \cellcolor{lblu_tab}{98.9} & \cellcolor{lblu_tab}{72.0} & \cellcolor{lblu_tab}{34.8} & \cellcolor{lblu_tab}{94.1} & \cellcolor{lblu_tab}{43.7} \\
        \hline
        \multirow{2}{*}{\makecell[c]{\rotatebox{90}{Sch.}}} & Cosine & 98.9 & 72.1 & 34.5 & 94.2 & 43.8 \\
        & \cellcolor{lblu_tab}{Step} & \cellcolor{lblu_tab}{98.9} & \cellcolor{lblu_tab}{72.0} & \cellcolor{lblu_tab}{34.8} & \cellcolor{lblu_tab}{94.1} & \cellcolor{lblu_tab}{43.7} \\
        \bottomrule[1.5pt]
        \end{tabular}
    }
    \vspace{-1.0em}
\end{table}

\section{More Ablations for Hyper-parameters} \label{sec:app_ablation}
\textbf{\textit{1)} Resolution.} The upper part of \Fig\ref{fig:ablation_resolution_epoch} shows the result changes of different methods on five comprehensive metrics under resolutions from 64 to 512. Our method significantly improves as the resolution increases, showing a clear advantage over the comparison methods above a resolution of 256$\times$256. Considering the computational load and performance, this paper adopts the default resolution of 256$\times$256 in this field. \\
\textbf{\textit{2)} Training Epoch.} The lower part of \Fig\ref{fig:ablation_resolution_epoch} shows the impact of training epochs on the results. Our method converges faster than the comparison methods, achieving nearly stable results with only 100 epochs. \\
\textbf{\textit{3)} Loss Function.} RD~\cite{rd} and UniAD~\cite{uniad} show significant differences in performance when using different pixel-level constraints. The upper part of \Tab\ref{tab:supp_ablation_loss_scheduler} shows the results of our method under flattened cosine distance, pixel-level cosine distance, L1, and MSE constraints, indicating that our method has strong robustness. \\
\textbf{\textit{4)} Learning Rate Scheduler.} The lower part of \Tab\ref{tab:supp_ablation_loss_scheduler} shows that the cosine scheduler~\cite{cosine}, which is effective in other fields, does not significantly differ from the Step strategy in the AD field.

\section{More Qualitative Visualization of COCO-AD Dataset} \label{sec:app_cocoad_vis}
\begin{figure*}[htp]
    \centering
    \includegraphics[width=1.0\linewidth]{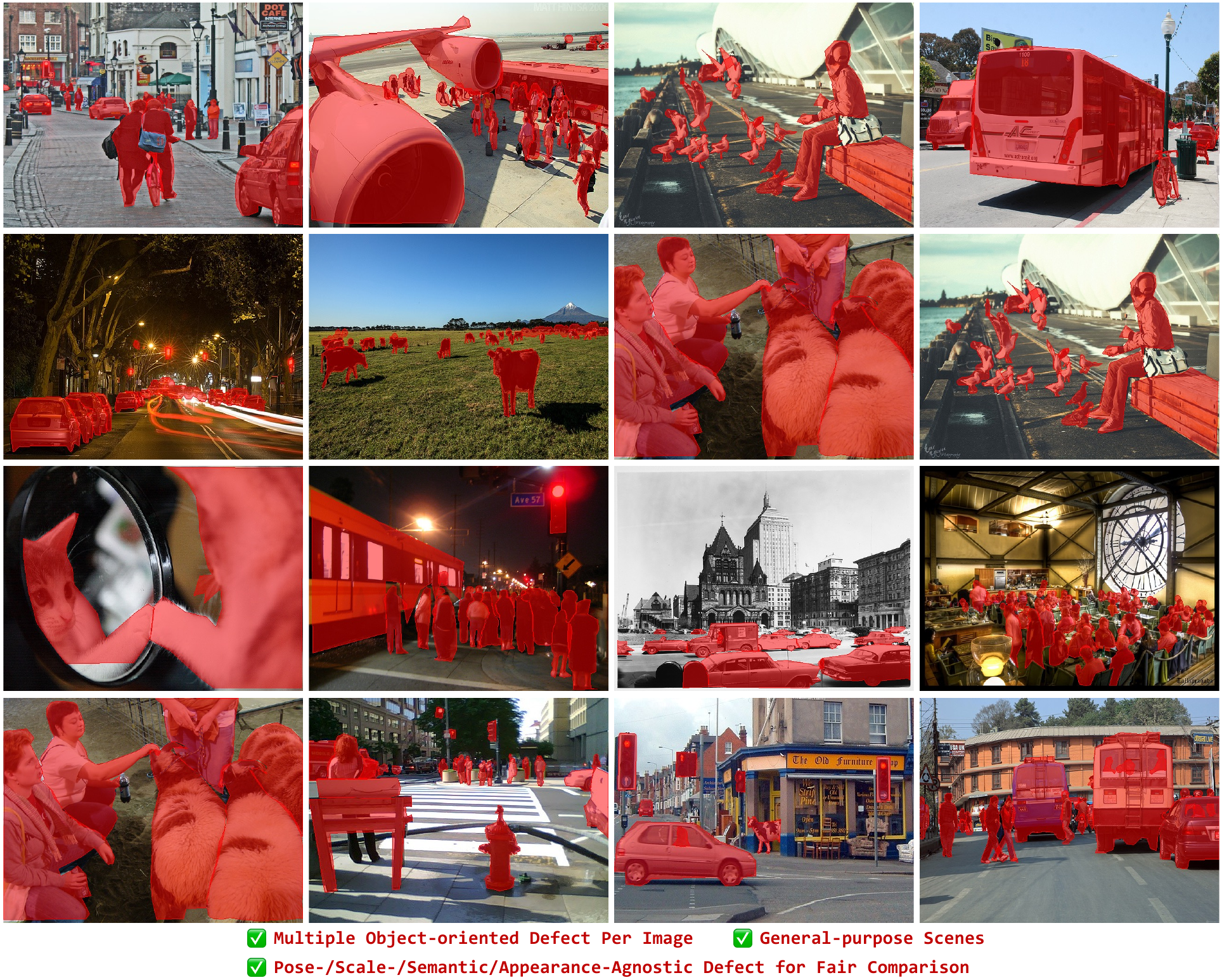}
    \caption{\textbf{More challenging visualizations of COCO-AD dataset for split0.} The red regions represent anomalous objects during testing.}
    \label{fig:supp_coco_vis}
\end{figure*}

\Fig\ref{fig:supp_coco_vis} presents more visualization results from the COCO-AD dataset, where general-purpose scenes and complex anomalous attributes demonstrate its highly challenging nature. 

\section{More Quantitative Comprehensive Results on Popular AD Datasets} \label{sec:app_comprehensive}
We further conduct quantitative experiments on MVTec 3D AD, Uni-Medical, and Real-IAD datasets for evaluating different methods comprehensively beyond Table~3 in the main paper. 
Considering effectiveness and training cost, we only select powerful and representative methods for experiments. 

\begin{table*}[htp]
    \centering
    \caption{\textbf{Different fine-grained multi-class anomaly classification and segmentation results with SoTAs on more AD datasets beyond Table~\blue{3}.} Prefix `m' stands for averaged metric results for all categories. Metrics marked in \redzjn{red} are recommended. \textbf{Bold} and \underline{underline} represent optimal and sub-optimal results, respectively.}
    \vspace{-0.5em}
    \label{tab:supp_sota}
    \renewcommand{\arraystretch}{1.0}
    \setlength\tabcolsep{6.0pt}
    \resizebox{1.0\linewidth}{!}{
        \begin{tabular}{p{0.5cm}<{\centering} p{2.3cm}<{\centering} p{2.0cm}<{\centering} p{2.0cm}<{\centering} p{2.0cm}<{\centering}p{2.0cm}<{\centering} p{2.0cm}<{\centering} p{2.0cm}<{\centering} p{2.0cm}<{\centering}}
            \toprule[0.17em]
            & Fineness-level & Metric & RD$^*$~\cite{rd} & UniAD$^\dagger$~\cite{uniad} & SimpleNet$^*$~\cite{simplenet} & DiAD~\cite{diad} & InvAD-lite & InvAD \\
            \hline
            \multirow{14}{*}{\makecell[c]{\rotatebox{90}{\textbf{MVTec 3D AD}~\cite{mvtec3d}}}} & \multirow{3}{*}{\makecell[c]{Image-level\\ (Classification)}} & mAU-ROC & 77.9 & 78.9 & 72.5 & 84.6 & \underline{85.3} & \textbf{86.1} \\ 
            & & mAP & 92.4 & 93.4 & 91.0 & 94.8 & \underline{95.2} & \textbf{95.8} \\ 
            & & m$F_1$-max & 91.4 & 91.4 & 90.3 & \textbf{95.6} & 93.0 & \underline{93.2} \\  
            \cline{2-9}
            & Region-level & \redzjn{mAU-PRO} & 93.5 & 88.1 & 77.6 & 87.8 & \underline{94.1} & \textbf{94.7} \\  
            \cline{2-9}
            & \multirow{3}{*}{\makecell[c]{Pixel-level\\ (Segmentation)}} & mAU-ROC & 98.4 & 96.5 & 93.6 & 96.4 & \underline{98.6}& \textbf{98.8} \\ 
            & & mAP & 29.8 & 21.2 & 18.3 & 25.3 & \underline{37.2} & \textbf{37.8} \\ 
            & & m$F_1$-max & 36.4 & 28.0 & 25.3 & 32.3 & \underline{41.4} & \textbf{42.5} \\ 
            \cline{3-9}
            & \multirow{4}{*}{\makecell[c]{Proposed}} & m$F_1$$^{.2}_{.8}$ & 16.0 & 12.2 & 11.0 & \pzo5.0 & \underline{21.6} & \textbf{22.0} \\ 
            & & mAcc$^{.2}_{.8}$ & 53.1 & 43.6 & \underline{52.7} & 51.4 & \textbf{55.3} & 50.5 \\   
            & & mIoU$^{.2}_{.8}$ & \pzo9.4 & \pzo7.0 & \pzo6.2 & \pzo2.6 & \underline{12.9} & \textbf{13.2} \\ 
            & & \redzjn{mIoU-max} & 22.8 & 16.8 & 15.0 & \pzo5.4 & \underline{26.5} & \textbf{27.5} \\ 
            \cline{2-9}
            & \multirow{3}{*}{\makecell[c]{Averaged \\ Metrics}} & \redzjn{mAD$_I$} & 87.2 & 87.9 & 84.6 & \textbf{91.7} & \underline{91.2} & \textbf{91.7} \\ 
            & & \redzjn{mAD$_P$} & 54.9 & 48.6 & 45.7 & 51.3 & \underline{59.1} & \textbf{59.7} \\ 
            & & \redzjn{mAD$^{.2}_{.8}$} & 26.2 & 20.9 & 23.3 & 26.3 & \textbf{29.9} & \underline{28.6} \\ 
            
            \hline
            \multirow{14}{*}{\makecell[c]{\rotatebox{90}{\textbf{Uni-Medical}~\cite{vitad}}}} & \multirow{3}{*}{\makecell[c]{Image-level\\ (Classification)}} & mAU-ROC & 75.6 & 78.5 & 75.6 & \textbf{85.1} & 79.5 & \underline{82.2} \\ 
            & & mAP & 75.8 & 75.2 & 76.9 & \textbf{84.5} & 78.3 & \underline{79.6} \\ 
            & & m$F_1$-max & 78.0 & 76.6 & 76.8 & \textbf{81.2} & 79.1 & \underline{80.6} \\ 
            \cline{2-9}
            & Region-level & \redzjn{mAU-PRO} & \underline{86.4} & 85.0 & 80.5 & 85.4 & 85.5 & \textbf{89.6} \\ 
            \cline{2-9}
            & \multirow{3}{*}{\makecell[c]{Pixel-level\\ (Segmentation)}} & mAU-ROC & \underline{96.5} & 96.4 & 95.9 & 95.9 & 96.4 & \textbf{97.4} \\ 
            & & mAP & 38.7 & 37.6 & 38.3 & 38.0 & \underline{40.1} & \textbf{47.5} \\ 
            & & m$F_1$-max & 40.1 & 40.2 & 39.6 & 35.6 & \underline{40.4} & \textbf{47.1} \\ 
            \cline{3-9}
            & \multirow{4}{*}{\makecell[c]{Proposed}} & m$F_1$$^{.2}_{.8}$ & 16.4 & 13.3 & 15.0 & 18.0 & \underline{18.3} & \textbf{21.8} \\ 
            & & mAcc$^{.2}_{.8}$ & 43.2 & 37.8 & \textbf{46.5} & 37.6 & 40.5 & \underline{45.2} \\  
            & & mIoU$^{.2}_{.8}$ & 10.0 & \pzo8.0 & \pzo9.0 & 10.4 & \underline{11.2} & \textbf{13.8} \\ 
            & & \redzjn{mIoU-max} & 27.3 & 26.8 & 25.8 & 25.0 & \underline{27.6} & \textbf{33.3} \\ 
            \cline{2-9}
            & \multirow{3}{*}{\makecell[c]{Averaged \\ Metrics}} & \redzjn{mAD$_I$} & 76.5 & 76.8 & 76.4 & \textbf{83.6} & 79.0 & \underline{80.8} \\ 
            & & \redzjn{mAD$_P$} & 58.4 & 58.1 & 57.9 & 56.5 & \underline{59.0} & \textbf{64.0} \\ 
            & & \redzjn{mAD$^{.2}_{.8}$} & 23.2 & 19.7 & \underline{23.5} & 22.0 & 23.3 & \textbf{26.9} \\ 

            \hline
            \multirow{14}{*}{\makecell[c]{\rotatebox{90}{\textbf{Real-IAD}~\cite{realiad}}}} & \multirow{3}{*}{\makecell[c]{Image-level\\ (Classification)}} & mAU-ROC & 82.4 & 82.9 & 57.2 & 75.6 & \underline{87.2} & \textbf{89.0} \\ 
            & & mAP & 79.0 & 80.8 & 53.4 & 66.4 & \underline{85.1} & \textbf{86.4} \\ 
            & & m$F_1$-max & 73.9 & 74.4 & 61.5 & 69.9 & \underline{77.8} & \textbf{79.6} \\ 
            \cline{2-9}
            & Region-level & \redzjn{mAU-PRO} & 89.6 & 86.4 & 39.0 & 58.1 & \underline{91.6} & \textbf{91.9} \\ 
            \cline{2-9}
            & \multirow{3}{*}{\makecell[c]{Pixel-level\\ (Segmentation)}} & mAU-ROC & 97.3 & 97.4 & 75.7 & 88.0 & \underline{98.1} & \textbf{98.4} \\ 
            & & mAP & 25.0 & 22.9 & \pzo2.8 & \pzo2.9 & \textbf{31.6} & \underline{30.7} \\ 
            & & m$F_1$-max & 32.7 & 30.3 & \pzo6.5 & \pzo7.1 & \textbf{37.9} & \underline{37.6} \\ 
            \cline{3-9}
            & \multirow{4}{*}{\makecell[c]{Proposed}} & m$F_1$$^{.2}_{.8}$ & 12.2 & 10.5 & \pzo2.2 & \pzo2.9 & \underline{17.6} & \textbf{17.7} \\ 
            & & mAcc$^{.2}_{.8}$ & \textbf{46.2} & 35.0 & \underline{41.7} & 34.9 & 36.8 & 36.1 \\ 
            & & mIoU$^{.2}_{.8}$ & \pzo7.0 & \pzo6.0 & \pzo1.2 & \pzo1.5 & \underline{10.3} & \textbf{10.4} \\ 
            & & \redzjn{mIoU-max} & 19.8 & 18.3 & \pzo3.5 & \pzo3.7 & \textbf{23.7} & \underline{23.5} \\ 
            \cline{2-9}
            & \multirow{3}{*}{\makecell[c]{Averaged \\ Metrics}} & \redzjn{mAD$_I$} & 78.4 & 79.4 & 57.4 & 70.6 & \underline{83.4} & \textbf{85.0} \\ 
            & & \redzjn{mAD$_P$} & 51.7 & 50.2 & 28.3 & 32.7 & \textbf{55.9} & \underline{55.6} \\ 
            & & \redzjn{mAD$^{.2}_{.8}$} & \textbf{21.8} & 17.2 & 15.0 & 13.1 & \underline{21.6} & 21.4 \\ 
            \toprule[0.12em] 
        \end{tabular}
    }
    \vspace{-1.5em}
\end{table*}

\section{Quantitative Results for Each Category on COCO-AD} \label{sec:app_cocoad}
\Tab\ref{tab:supp_coco} shows detailed quantitative single-split results on the COCO-AD dataset. 

\begin{table*}[htp]
    \centering
    \caption{\textbf{Quantitative results for each split on COCO-AD dataset.} These are respectively the image-level mAU-ROC/mAP/m$F_1$-max, region-level \redzjn{mAU-PRO}, pixel-level mAU-ROC/mAP/m$F_1$-max, the proposed pixel-level m$F_1$$^{.2}_{.8}$/mAcc$^{.2}_{.8}$/mIoU$^{.2}_{.8}$/\redzjn{mIoU-max}, and averaged \redzjn{mAD$_I$}/\redzjn{mAD$_P$}/\redzjn{mAD$^{.2}_{.8}$}. Subsequent tables will default to this setting.}
    \label{tab:supp_coco}
    \renewcommand{\arraystretch}{1.0}
    \setlength\tabcolsep{6.0pt}
    \resizebox{1.0\linewidth}{!}{
        \begin{tabular}{p{0.3cm}<{\centering} p{2.0cm}<{\centering} p{3.9cm}<{\centering} p{1.6cm}<{\centering} p{3.9cm}<{\centering} p{4.9cm}<{\centering} p{3.3cm}<{\centering}}
            \toprule[0.17em]
            & Method & mAU-ROC/mAP/m$F_1$-max & \redzjn{mAU-PRO} & mAU-ROC/mAP/m$F_1$-max & m$F_1$$^{.2}_{.8}$/mAcc$^{.2}_{.8}$/mIoU$^{.2}_{.8}$/\redzjn{mIoU-max} & \redzjn{mAD$_I$}/\redzjn{mAD$_P$}/\redzjn{mAD$^{.2}_{.8}$} \\
            \hline
            \multirow{8}{*}{\makecell[c]{\rotatebox{90}{\textbf{Split0}}}} & DRÆM$^*$~\cite{draem} & 53.9 / 76.4 / 85.0 & 12.5 & 54.0 / 17.7 / 27.7 & 11.3 / 26.0 / \pzo8.1 / 16.1 & 71.8 / 33.1 / 15.1 \\ 
            & RD$^*$~\cite{rd} & 65.7 / 81.9 / 85.1 & 45.9 & 72.1 / 30.8 / 38.2 & 17.7 / 34.6 / 10.5 / 23.6 & 77.6 / 47.0 / 20.9 \\ 
            & UniAD$^\dagger$~\cite{uniad} & 66.1 / 84.0 / 85.1 & 36.5 & 70.8 / 29.4 / 36.7 & 11.3 / 22.6 / \pzo6.7 / 22.5 & 78.4 / 45.6 / 13.5 \\ 
            & DeSTSeg$^*$~\cite{destseg} & 59.7 / 79.1 / 85.0 & 23.5 & 61.8 / 21.5 / 27.7 & \pzo1.0 / \pzo0.6 / \pzo0.5 / 16.1 & 74.6 / 37.0 / \pzo0.7 \\ 
            & SimpleNet$^*$~\cite{simplenet} & 57.8 / 77.4 / 84.7 & 26.8 & 64.0 / 27.4 / 34.4 & 13.6 / 31.4 / \pzo7.8 / 20.7 & 73.3 / 41.9 / 17.6 \\ 
            & DiAD~\cite{diad} & 57.5 / 77.5 / 85.3 & 28.8 & 67.0 / 33.4 / 26.2 & 19.7 / 30.6 / 12.4 / 20.1 & 73.4 / 42.2 / 20.9 \\ 
            \cline{2-7}
            & InvAD-lite & 77.0 / 90.1 / 85.5 & 46.8 & 76.6 / 40.3 / 42.9 & 21.5 / 38.7 / 13.0 / 27.3 & 84.2 / 53.3 / 24.4 \\ 
            & InvAD & 73.8 / 87.8 / 85.1 & 51.1 & 78.9 / 42.6 / 45.6 & 22.3 / 39.0 / 13.6 / 29.6 & 82.2 / 55.7 / 25.0 \\ 
            
            \hline
            \multirow{8}{*}{\makecell[c]{\rotatebox{90}{\textbf{Split1}}}} & DRÆM$^*$~\cite{draem} & 51.3 / 43.9 / 60.9 & \pzo6.8 & 52.4 / \pzo3.2 / \pzo6.0 & \pzo3.9 / 18.7 / \pzo3.3 / \pzo3.1 & 52.0 / 20.5 / \pzo8.6 \\ 
            & RD$^*$~\cite{rd} & 54.9 / 46.8 / 61.1 & 40.6 & 70.7 / \pzo6.2 / 11.2 & \pzo4.4 / 27.6 / \pzo2.3 / \pzo5.9 & 54.3 / 29.4 / 11.4 \\ 
            & UniAD$^\dagger$~\cite{uniad} & 56.1 / 47.8 / 61.1 & 31.8 & 70.0 / \pzo6.2 / 11.3 & \pzo4.0 / 28.9 / \pzo2.1 / \pzo6.0 & 55.0 / 29.2 / 11.7 \\ 
            & DeSTSeg$^*$~\cite{destseg} & 55.6 / 47.9 / 61.2 & 27.7 & 69.3 / \pzo6.8 / 11.3 & \pzo4.0 / \pzo2.5 / \pzo2.0 / \pzo6.0 & 54.9 / 29.1 / \pzo2.8 \\ 
            & SimpleNet$^*$~\cite{simplenet} & 51.2 / 42.3 / 59.0 & 33.0 & 61.4 / \pzo4.9 / \pzo8.9 & \pzo5.6 / 34.2 / \pzo2.9 / \pzo4.7 & 50.8 / 25.1 / 14.2 \\ 
            & DiAD~\cite{diad} & 54.4 / 49.8 / 62.2 & 28.8 & 71.3 / 11.8 / \pzo7.8 & \pzo1.9 / 21.2 / \pzo1.6 / \pzo6.2 & 55.5 / 30.3 / \pzo8.2 \\ 
            \cline{2-7}
            & InvAD-lite & 53.7 / 46.8 / 61.0 & 33.1 & 72.4 / \pzo7.8 / 12.9 & \pzo5.6 / 31.7 / \pzo2.9 / \pzo6.9 & 53.8 / 31.0 / 13.4 \\ 
            & InvAD & 55.8 / 48.4 / 60.9 & 41.7 & 74.3 / \pzo8.7 / 14.4 & \pzo7.1 / 38.1 / \pzo3.7 / \pzo7.8 & 55.0 / 32.5 / 16.3 \\ 
            
            \hline
            \multirow{8}{*}{\makecell[c]{\rotatebox{90}{\textbf{Split2}}}} & DRÆM$^*$~\cite{draem} & 54.5 / 33.4 / 50.4 & 12.1 & 50.4 / \pzo6.1 / 11.9 & \pzo7.5 / 28.3 / \pzo5.8 / \pzo6.3 & 46.1 / 22.8 / 13.9 \\ 
            & RD$^*$~\cite{rd} & 59.6 / 39.4 / 51.3 & 42.9 & 68.4 / 11.6 / 18.9 & \pzo9.6 / 38.9 / \pzo5.2 / 10.4 & 50.1 / 33.0 / 17.9 \\ 
            & UniAD$^\dagger$~\cite{uniad} & 52.3 / 30.8 / 49.5 & 27.0 & 60.9 / \pzo7.7 / 14.7 & \pzo5.6 / 29.0 / \pzo3.0 / \pzo7.9 & 44.2 / 27.8 / 12.5 \\  
            & DeSTSeg$^*$~\cite{destseg} & 55.8 / 37.6 / 50.1 & 26.3 & 61.1 / \pzo9.8 / 13.9 & \pzo2.8 / \pzo1.9 / \pzo1.5 / \pzo7.4 & 47.8 / 28.3 / \pzo2.1 \\ 
            & SimpleNet$^*$~\cite{simplenet} & 60.1 / 38.5 / 50.7 & 29.2 & 57.4 / \pzo8.2 / 14.4 & \pzo8.5 / 33.9 / \pzo4.5 / \pzo7.8 & 49.8 / 26.7 / 15.6 \\ 
            & DiAD~\cite{diad} & 63.8 / 43.4 / 52.5 & 33.2 & 68.0 / 19.2 / 12.2 & \pzo8.0 / 29.3 / \pzo5.0 / 10.6 & 53.2 / 33.1 / 14.1 \\ 
            \cline{2-7}
            & InvAD-lite & 68.7 / 49.2 / 54.5 & 41.3 & 73.2 / 16.4 / 23.1 & 10.2 / 34.7 / \pzo5.6 / 13.1 & 57.5 / 37.6 / 16.8 \\ 
            & InvAD & 68.2 / 48.0 / 55.0 & 47.7 & 75.7 / 16.9 / 24.9 & 11.3 / 38.2 / \pzo6.2 / 14.2 & 57.1 / 39.2 / 18.6 \\ 
            
            \hline
            \multirow{8}{*}{\makecell[c]{\rotatebox{90}{\textbf{Split3}}}} & DRÆM$^*$~\cite{draem} & 60.5 / 41.7 / 53.5 & 15.6 & 50.4 / \pzo6.5 / 12.4 & \pzo8.9 / 24.5 / \pzo5.2 / \pzo6.6 & 51.9 / 23.1 / 12.9 \\ 
            & RD$^*$~\cite{rd} & 53.5 / 36.4 / 51.5 & 33.4 & 58.3 / \pzo8.4 / 14.2 & \pzo7.8 / 34.1 / \pzo4.1 / \pzo7.6 & 47.1 / 27.0 / 15.3 \\ 
            & UniAD$^\dagger$~\cite{uniad} & 50.1 / 33.5 / 51.2 & 31.4 & 59.8 / \pzo8.3 / 14.8 & \pzo5.4 / 24.9 / \pzo2.9 / \pzo8.0 & 44.9 / 27.6 / 11.1 \\ 
            & DeSTSeg$^*$~\cite{destseg} & 53.5 / 36.5 / 51.2 & 16.8 & 51.2 / \pzo6.9 / 12.4 & \pzo1.6 / \pzo0.9 / \pzo0.8 / \pzo6.6 & 47.1 / 23.5 / \pzo1.1 \\ 
            & SimpleNet$^*$~\cite{simplenet} & 59.2 / 39.2 / 52.2 & 21.0 & 55.3 / \pzo8.2 / 13.9 & \pzo6.4 / 29.6 / \pzo3.4 / \pzo7.5 & 50.2 / 25.8 / 13.1 \\ 
            & DiAD~\cite{diad} & 60.1 / 41.4 / 52.9 & 32.3 & 65.9 / 17.5 / 10.6 & \pzo8.6 / 43.4 / \pzo5.3 / \pzo9.6 & 51.5 / 31.3 / 19.1 \\ 
            \cline{2-7}
            & InvAD-lite & 59.2 / 40.5 / 52.8 & 31.6 & 60.0 / \pzo8.9 / 14.6 & \pzo7.5 / 31.6 / \pzo4.0 / \pzo7.9 & 50.8 / 27.8 / 14.4 \\ 
            & InvAD & 65.8 / 46.8 / 55.3 & 39.1 & 64.1 / 10.5 / 16.8 & \pzo8.9 / 34.8 / \pzo4.8 / \pzo9.2 & 56.0 / 30.5 / 16.2 \\ 
            \toprule[0.12em] 
        \end{tabular}
    }
\end{table*}

\section{Quantitative Results for Each Category on MVTec AD} \label{sec:app_mvtec}

\Tab\ref{tab:supp_mvtec_textures} and \Tab\ref{tab:supp_mvtec_objects} show detailed quantitative single-class results for Textures and Object categories on the MVTec AD dataset.

\begin{table*}[htp]
    \centering
    \caption{\textbf{Quantitative results for each class (Textures) on MVTec AD dataset.} These are respectively the image-level mAU-ROC/mAP/m$F_1$-max, region-level \redzjn{mAU-PRO}, pixel-level mAU-ROC/mAP/m$F_1$-max, the proposed pixel-level m$F_1$$^{.2}_{.8}$/mAcc$^{.2}_{.8}$/mIoU$^{.2}_{.8}$/\redzjn{mIoU-max}, and averaged \redzjn{mAD$_I$}/\redzjn{mAD$_P$}/\redzjn{mAD$^{.2}_{.8}$}.}
    \label{tab:supp_mvtec_textures}
    \renewcommand{\arraystretch}{1.0}
    \setlength\tabcolsep{6.0pt}
    \resizebox{1.0\linewidth}{!}{
        \begin{tabular}{p{0.3cm}<{\centering} p{2.0cm}<{\centering} p{3.9cm}<{\centering} p{1.6cm}<{\centering} p{3.9cm}<{\centering} p{4.9cm}<{\centering} p{3.3cm}<{\centering}}
            \toprule[0.17em]
            & Method & mAU-ROC/mAP/m$F_1$-max & \redzjn{mAU-PRO} & mAU-ROC/mAP/m$F_1$-max & m$F_1$$^{.2}_{.8}$/mAcc$^{.2}_{.8}$/mIoU$^{.2}_{.8}$/\redzjn{mIoU-max} & \redzjn{mAD$_I$}/\redzjn{mAD$_P$}/\redzjn{mAD$^{.2}_{.8}$} \\
            \hline
            \multirow{8}{*}{\makecell[c]{\rotatebox{90}{\textbf{carpet}}}} & DRÆM$^*$~\cite{draem} & 97.2 / 99.1 / 96.7 & 93.1 & 98.1 / 78.7 / 73.1 & 46.5 / 33.7 / 31.9 / 57.6 & 97.7 / 83.3 / 37.4 \\ 
& RD$^*$~\cite{rd} & 98.5 / 99.6 / 97.2 & 95.1 & 99.0 / 58.5 / 60.5 & 36.4 / 50.3 / 24.0 / 43.3 & 98.4 / 72.7 / 36.9 \\ 
& UniAD$^\dagger$~\cite{uniad} & 99.8 / 99.9 / 99.4 & 94.4 & 98.4 / 51.4 / 51.5 & 30.3 / 43.7 / 19.3 / 34.6 & 99.7 / 67.1 / 31.1 \\ 
& DeSTSeg$^*$~\cite{destseg} & 95.9 / 98.8 / 94.9 & 89.3 & 93.6 / 59.9 / 58.9 & 48.3 / 55.4 / 32.3 / 41.7 & 96.5 / 70.8 / 45.3 \\ 
& SimpleNet$^*$~\cite{simplenet} & 95.7 / 98.7 / 93.2 & 90.6 & 97.4 / 38.7 / 43.2 & 24.5 / 41.1 / 14.6 / 27.5 & 95.9 / 59.8 / 26.7 \\ 
& DiAD~\cite{diad} & 99.7 / 96.5 / 91.8 & 86.6 & 98.4 / 52.2 / 54.8 & 12.3 / 34.2 / \pzo6.8 / 15.7 & 96.0 / 68.5 / 17.8 \\ 
& InvAD-lite & 99.8 / 99.9 / 98.9 & 95.4 & 99.1 / 62.2 / 61.3 & 39.8 / 54.7 / 26.3 / 44.2 & 99.5 / 74.2 / 40.3 \\ 
& InvAD & 98.2 / 99.5 / 96.6 & 94.9 & 99.1 / 61.4 / 61.7 & 40.3 / 54.8 / 26.7 / 44.6 & 98.1 / 74.1 / 40.6 \\ 
            
            \hline
            \multirow{8}{*}{\makecell[c]{\rotatebox{90}{\textbf{grid}}}} & DRÆM$^*$~\cite{draem} & 99.2 / 99.7 / 98.2 & 92.1 & 99.0 / 44.5 / 46.2 & 30.7 / 28.2 / 18.8 / 30.0 & 99.0 / 63.2 / 25.9 \\ 
& RD$^*$~\cite{rd} & 98.0 / 99.4 / 96.6 & 97.0 & 99.2 / 46.0 / 47.4 & 31.9 / 57.6 / 19.6 / 31.0 & 98.0 / 64.2 / 36.4 \\ 
& UniAD$^\dagger$~\cite{uniad} & 99.3 / 99.8 / 99.1 & 92.9 & 97.7 / 23.7 / 30.4 & 18.7 / 42.3 / 10.6 / 18.0 & 99.4 / 50.6 / 23.9 \\ 
& DeSTSeg$^*$~\cite{destseg} & 97.9 / 99.2 / 96.6 & 86.8 & 97.0 / 42.1 / 46.9 & 33.0 / 34.5 / 20.5 / 30.6 & 97.9 / 62.0 / 29.3 \\ 
& SimpleNet$^*$~\cite{simplenet} & 97.6 / 99.2 / 96.4 & 88.6 & 96.8 / 20.5 / 27.6 & 14.8 / 52.2 / \pzo8.3 / 16.0 & 97.7 / 48.3 / 25.1 \\ 
& DiAD~\cite{diad} & 94.8 / 98.8 / 95.2 & 80.5 & 96.8 / 50.1 / 57.8 & \pzo5.3 / 23.8 / \pzo2.7 / \pzo5.3 & 96.3 / 68.2 / 10.6 \\ 
& InvAD-lite & 100. / 100. / 100. & 96.9 & 99.3 / 49.7 / 49.8 & 31.4 / 41.2 / 19.6 / 33.2 & 100. / 66.3 / 30.7 \\ 
& InvAD & 100. / 100. / 100. & 96.3 & 99.2 / 44.5 / 47.1 & 28.4 / 35.7 / 17.6 / 30.8 & 100. / 63.6 / 27.2 \\ 
            
            \hline
            \multirow{8}{*}{\makecell[c]{\rotatebox{90}{\textbf{leather}}}} & DRÆM$^*$~\cite{draem} & 97.7 / 99.3 / 95.0 & 88.5 & 98.9 / 60.3 / 57.4 & 30.6 / 19.7 / 18.5 / 40.2 & 97.3 / 72.2 / 22.9 \\ 
& RD$^*$~\cite{rd} & 100. / 100. / 100. & 97.4 & 99.3 / 38.0 / 45.1 & 27.6 / 49.9 / 16.8 / 29.1 & 100. / 60.8 / 31.4 \\ 
& UniAD$^\dagger$~\cite{uniad} & 100. / 100. / 100. & 96.8 & 98.8 / 34.2 / 35.5 & 20.6 / 40.6 / 11.9 / 21.6 & 100. / 56.2 / 24.4 \\ 
& DeSTSeg$^*$~\cite{destseg} & 99.2 / 99.8 / 98.9 & 91.1 & 99.5 / 71.5 / 66.5 & 57.6 / 50.1 / 40.9 / 49.8 & 99.3 / 79.2 / 49.5 \\ 
& SimpleNet$^*$~\cite{simplenet} & 100. / 100. / 100. & 92.7 & 98.7 / 28.5 / 32.9 & 20.3 / 64.5 / 11.6 / 19.7 & 100. / 53.4 / 32.1 \\ 
& DiAD~\cite{diad} & 89.0 / 97.5 / 95.5 & 87.2 & 97.1 / 42.0 / 45.3 & \pzo5.4 / 38.2 / \pzo2.9 / \pzo7.0 & 94.0 / 61.5 / 15.5 \\ 
& InvAD-lite & 100. / 100. / 100. & 98.5 & 99.3 / 44.7 / 48.1 & 32.5 / 70.3 / 19.9 / 31.7 & 100. / 64.0 / 40.9 \\ 
& InvAD & 100. / 100. / 100. & 97.3 & 99.4 / 47.1 / 49.5 & 32.6 / 60.6 / 20.1 / 32.9 & 100. / 65.3 / 37.8 \\ 
            
            \hline
            \multirow{8}{*}{\makecell[c]{\rotatebox{90}{\textbf{tile}}}} & DRÆM$^*$~\cite{draem} & 100. / 100. / 100. & 97.0 & 99.2 / 93.6 / 86.0 & 65.4 / 50.8 / 50.0 / 75.4 & 100. / 92.9 / 55.4 \\ 
& RD$^*$~\cite{rd} & 98.3 / 99.3 / 96.4 & 85.8 & 95.3 / 48.5 / 60.5 & 26.8 / 33.3 / 17.6 / 43.4 & 98.0 / 68.1 / 25.9 \\ 
& UniAD$^\dagger$~\cite{uniad} & 99.9 / 99.9 / 99.4 & 78.4 & 92.3 / 41.5 / 50.3 & 20.7 / 25.9 / 13.1 / 33.6 & 99.7 / 61.4 / 19.9 \\ 
& DeSTSeg$^*$~\cite{destseg} & 97.0 / 98.9 / 95.3 & 87.1 & 93.0 / 71.0 / 66.2 & 41.8 / 28.2 / 27.1 / 49.4 & 97.1 / 76.7 / 32.4 \\ 
& SimpleNet$^*$~\cite{simplenet} & 99.3 / 99.8 / 98.8 & 90.6 & 95.7 / 60.5 / 59.9 & 35.8 / 50.3 / 23.6 / 42.7 & 99.3 / 72.0 / 36.6 \\ 
& DiAD~\cite{diad} & 99.5 / 99.7 / 97.3 & 91.5 & 98.3 / 79.2 / 80.4 & 22.7 / 33.5 / 13.1 / 19.8 & 98.8 / 86.0 / 23.1 \\ 
& InvAD-lite & 98.2 / 99.3 / 97.1 & 80.0 & 94.4 / 46.5 / 55.5 & 31.9 / 43.1 / 20.5 / 38.4 & 98.2 / 65.5 / 31.8 \\ 
& InvAD & 100. / 100. / 100. & 86.0 & 95.7 / 51.0 / 62.2 & 37.4 / 47.4 / 24.9 / 45.1 & 100. / 69.6 / 36.6 \\ 
            
            \hline
            \multirow{8}{*}{\makecell[c]{\rotatebox{90}{\textbf{wood}}}} & DRÆM$^*$~\cite{draem} & 100. / 100. / 100. & 94.2 & 96.9 / 81.4 / 74.6 & 49.1 / 34.0 / 33.7 / 59.5 & 100. / 84.3 / 38.9 \\ 
& RD$^*$~\cite{rd} & 99.2 / 99.8 / 98.3 & 90.0 & 95.3 / 47.8 / 51.0 & 25.9 / 35.0 / 16.2 / 34.3 & 99.1 / 64.7 / 25.7 \\ 
& UniAD$^\dagger$~\cite{uniad} & 98.9 / 99.7 / 97.5 & 86.7 & 93.2 / 37.4 / 42.8 & 22.2 / 29.4 / 13.4 / 27.2 & 98.7 / 57.8 / 21.7 \\ 
& DeSTSeg$^*$~\cite{destseg} & 99.9 / 100. / 99.2 & 83.4 & 95.9 / 77.3 / 71.3 & 36.5 / 24.0 / 23.5 / 55.4 & 99.7 / 81.5 / 28.0 \\ 
& SimpleNet$^*$~\cite{simplenet} & 98.4 / 99.5 / 96.7 & 76.3 & 91.4 / 34.8 / 39.7 & 24.9 / 46.9 / 14.7 / 24.8 & 98.2 / 55.3 / 28.8 \\ 
& DiAD~\cite{diad} & 99.1 / 96.0 / 91.6 & 90.6 & 97.3 / 30.0 / 38.3 & 11.2 / 33.1 / \pzo6.0 / 12.4 & 95.6 / 55.2 / 16.8 \\ 
& InvAD-lite & 97.3 / 99.1 / 96.8 & 88.3 & 94.7 / 45.1 / 47.5 & 29.2 / 51.1 / 18.0 / 31.1 & 97.7 / 62.4 / 32.8 \\ 
& InvAD & 99.5 / 99.8 / 98.3 & 90.3 & 95.5 / 51.2 / 51.8 & 34.4 / 52.1 / 21.8 / 34.9 & 99.2 / 66.2 / 36.1 \\ 
            \toprule[0.12em] 
        \end{tabular}
    }
\end{table*}

\begin{table*}[htp]
    \centering
    \caption{\textbf{Quantitative results for each class (Objects) on MVTec AD dataset.} These are respectively the image-level mAU-ROC/mAP/m$F_1$-max, region-level \redzjn{mAU-PRO}, pixel-level mAU-ROC/mAP/m$F_1$-max, the proposed pixel-level m$F_1$$^{.2}_{.8}$/mAcc$^{.2}_{.8}$/mIoU$^{.2}_{.8}$/\redzjn{mIoU-max}, and averaged \redzjn{mAD$_I$}/\redzjn{mAD$_P$}/\redzjn{mAD$^{.2}_{.8}$}.}
    \label{tab:supp_mvtec_objects}
    \renewcommand{\arraystretch}{1.0}
    \setlength\tabcolsep{6.0pt}
    \resizebox{1.0\linewidth}{!}{
        \begin{tabular}{p{0.3cm}<{\centering} p{2.0cm}<{\centering} p{3.9cm}<{\centering} p{1.6cm}<{\centering} p{3.9cm}<{\centering} p{4.9cm}<{\centering} p{3.3cm}<{\centering}}
            \toprule[0.17em]
            & Method & mAU-ROC/mAP/m$F_1$-max & \redzjn{mAU-PRO} & mAU-ROC/mAP/m$F_1$-max & m$F_1$$^{.2}_{.8}$/mAcc$^{.2}_{.8}$/mIoU$^{.2}_{.8}$/\redzjn{mIoU-max} & \redzjn{mAD$_I$}/\redzjn{mAD$_P$}/\redzjn{mAD$^{.2}_{.8}$} \\
            \hline
            \multirow{8}{*}{\makecell[c]{\rotatebox{90}{\textbf{bottle}}}} & DRÆM$^*$~\cite{draem} & 97.3 / 99.2 / 96.1 & 80.7 & 91.3 / 62.5 / 56.9 & 20.6 / 11.9 / 11.7 / 39.8 & 97.5 / 70.2 / 14.7 \\ 
& RD$^*$~\cite{rd} & 99.6 / 99.9 / 98.4 & 94.0 & 97.8 / 68.2 / 67.6 & 32.9 / 37.7 / 22.7 / 51.1 & 99.3 / 77.9 / 31.1 \\ 
& UniAD$^\dagger$~\cite{uniad} & 100. / 100. / 100. & 93.1 & 98.0 / 67.0 / 67.9 & 22.7 / 26.1 / 15.7 / 51.4 & 100. / 77.6 / 21.5 \\ 
& DeSTSeg$^*$~\cite{destseg} & 98.7 / 99.6 / 96.8 & 67.5 & 93.3 / 61.7 / 56.0 & 31.8 / 19.8 / 19.3 / 38.9 & 98.4 / 70.3 / 23.6 \\ 
& SimpleNet$^*$~\cite{simplenet} & 100. / 100. / 100. & 89.0 & 97.2 / 53.8 / 62.4 & 22.8 / 27.8 / 15.1 / 45.4 & 100. / 71.1 / 21.9 \\ 
& DiAD~\cite{diad} & 95.7 / 98.5 / 94.5 & 89.0 & 95.7 / 46.0 / 51.4 & 33.6 / 49.4 / 20.7 / 33.0 & 96.2 / 64.4 / 34.6 \\ 
& InvAD-lite & 100. / 100. / 100. & 95.5 & 98.5 / 75.9 / 73.6 & 43.9 / 48.5 / 31.7 / 58.3 & 100. / 82.7 / 41.4 \\ 
& InvAD & 100. / 100. / 100. & 95.3 & 98.6 / 76.6 / 73.2 & 44.1 / 48.6 / 31.9 / 57.7 & 100. / 82.8 / 41.5 \\  
            
            \hline
            \multirow{8}{*}{\makecell[c]{\rotatebox{90}{\textbf{cable}}}} & DRÆM$^*$~\cite{draem} & 61.1 / 74.0 / 76.3 & 40.1 & 75.9 / 14.7 / 17.8 & \pzo2.0 / \pzo1.1 / \pzo1.0 / \pzo9.8 & 70.5 / 36.1 / \pzo1.4 \\ 
& RD$^*$~\cite{rd} & 84.1 / 89.5 / 82.5 & 75.1 & 85.1 / 26.3 / 33.6 & 18.0 / 27.6 / 10.3 / 20.2 & 85.4 / 48.3 / 18.6 \\ 
& UniAD$^\dagger$~\cite{uniad} & 94.8 / 97.0 / 90.7 & 86.1 & 96.9 / 45.4 / 50.4 & 21.5 / 37.5 / 13.1 / 33.7 & 94.2 / 64.2 / 24.0 \\ 
& DeSTSeg$^*$~\cite{destseg} & 89.5 / 94.6 / 85.9 & 49.4 & 89.3 / 37.5 / 40.5 & 19.9 / 12.5 / 11.4 / 25.4 & 90.0 / 55.8 / 14.6 \\ 
& SimpleNet$^*$~\cite{simplenet} & 97.5 / 98.5 / 94.7 & 85.4 & 96.7 / 42.4 / 51.2 & 20.6 / 36.9 / 12.6 / 34.4 & 96.9 / 63.4 / 23.4 \\ 
& DiAD~\cite{diad} & 90.7 / 99.7 / 97.9 & 95.0 & 97.9 / 60.6 / 59.6 & 20.2 / 45.3 / 11.6 / 20.1 & 96.1 / 72.7 / 25.7 \\ 
& InvAD-lite & 97.8 / 98.6 / 95.3 & 86.5 & 90.8 / 38.5 / 47.0 & 21.9 / 29.3 / 13.3 / 30.7 & 97.2 / 58.8 / 21.5 \\ 
& InvAD & 98.8 / 99.3 / 95.7 & 92.3 & 97.5 / 48.8 / 53.3 & 27.7 / 39.3 / 17.2 / 36.4 & 97.9 / 66.5 / 28.1 \\  
            
            \hline
            \multirow{8}{*}{\makecell[c]{\rotatebox{90}{\textbf{capsule}}}} & DRÆM$^*$~\cite{draem} & 70.9 / 92.5 / 90.5 & 27.3 & 50.5 / \pzo6.0 / 10.0 & \pzo0.6 / \pzo0.3 / \pzo0.3 / \pzo5.3 & 84.6 / 22.2 / \pzo0.4 \\ 
& RD$^*$~\cite{rd} & 94.1 / 96.9 / 96.9 & 94.8 & 98.8 / 43.4 / 50.1 & 23.7 / 32.0 / 14.6 / 33.4 & 96.0 / 64.1 / 23.4 \\ 
& UniAD$^\dagger$~\cite{uniad} & 93.7 / 98.4 / 96.3 & 92.1 & 98.8 / 45.6 / 47.7 & 28.4 / 44.9 / 17.4 / 31.3 & 96.1 / 64.0 / 30.2 \\ 
& DeSTSeg$^*$~\cite{destseg} & 82.8 / 95.9 / 92.6 & 62.1 & 95.8 / 47.9 / 48.9 & 27.9 / 18.1 / 16.7 / 32.4 & 90.4 / 64.2 / 20.9 \\ 
& SimpleNet$^*$~\cite{simplenet} & 90.7 / 97.9 / 93.5 & 84.5 & 98.5 / 35.4 / 44.3 & 21.6 / 51.9 / 12.9 / 28.4 & 94.0 / 59.4 / 28.8 \\ 
& DiAD~\cite{diad} & 99.7 / 99.9 / 99.2 & 95.0 & 99.0 / 78.7 / 72.8 & 13.1 / 29.2 / \pzo7.2 / 13.7 & 99.6 / 83.5 / 16.5 \\ 
& InvAD-lite & 96.1 / 99.2 / 95.3 & 92.9 & 98.6 / 43.3 / 47.5 & 25.7 / 29.2 / 15.7 / 31.2 & 96.9 / 63.1 / 23.5 \\ 
& InvAD & 97.8 / 99.5 / 96.3 & 95.3 & 98.9 / 46.5 / 50.8 & 24.8 / 26.2 / 15.4 / 34.0 & 97.9 / 65.4 / 22.1 \\ 
            
            \hline
            \multirow{8}{*}{\makecell[c]{\rotatebox{90}{\textbf{hazelnut}}}} & DRÆM$^*$~\cite{draem} & 94.7 / 97.5 / 92.3 & 78.7 & 96.5 / 70.0 / 60.5 & 11.7 / \pzo6.4 / \pzo6.4 / 43.4 & 94.8 / 75.7 / \pzo8.2 \\ 
& RD$^*$~\cite{rd} & 60.8 / 69.8 / 86.4 & 92.7 & 97.9 / 36.2 / 51.6 & \pzo5.5 / \pzo4.8 / \pzo3.3 / 34.8 & 72.3 / 61.9 / \pzo4.5 \\ 
& UniAD$^\dagger$~\cite{uniad} & 100. / 100. / 100. & 94.1 & 98.0 / 53.8 / 56.3 & 25.9 / 50.8 / 16.2 / 39.2 & 100. / 69.4 / 31.0 \\ 
& DeSTSeg$^*$~\cite{destseg} & 98.8 / 99.2 / 98.6 & 84.5 & 98.2 / 65.8 / 61.6 & 39.2 / 27.3 / 25.3 / 44.5 & 98.9 / 75.2 / 30.6 \\ 
& SimpleNet$^*$~\cite{simplenet} & 99.9 / 100. / 99.3 & 87.4 & 98.4 / 44.6 / 51.4 & 21.9 / 37.8 / 13.5 / 34.6 & 99.7 / 64.8 / 24.4 \\ 
& DiAD~\cite{diad} & 99.8 / 99.6 / 97.4 & 90.0 & 95.1 / 15.6 / 31.7 & 21.3 / 51.4 / 12.3 / 23.4 & 98.9 / 47.5 / 28.3 \\ 
& InvAD-lite & 100. / 100. / 100. & 95.4 & 98.8 / 56.5 / 61.8 & 35.1 / 63.2 / 22.9 / 44.7 & 100. / 72.4 / 40.4 \\ 
& InvAD & 100. / 100. / 100. & 96.5 & 98.9 / 59.8 / 62.3 & 38.9 / 64.9 / 25.7 / 45.2 & 100. / 73.7 / 43.2 \\ 

            \hline
            \multirow{8}{*}{\makecell[c]{\rotatebox{90}{\textbf{metal\_nut}}}} & DRÆM$^*$~\cite{draem} & 81.8 / 95.0 / 92.0 & 66.4 & 74.4 / 31.1 / 21.0 & \pzo4.1 / \pzo2.1 / \pzo2.1 / 11.7 & 89.6 / 42.2 / \pzo2.8 \\ 
& RD$^*$~\cite{rd} & 100. / 100. / 99.5 & 91.9 & 93.8 / 62.3 / 65.4 & 28.5 / 39.5 / 18.8 / 48.6 & 99.8 / 73.8 / 28.9 \\ 
& UniAD$^\dagger$~\cite{uniad} & 98.3 / 99.5 / 98.4 & 81.8 & 93.3 / 50.9 / 63.6 & 24.5 / 33.0 / 15.9 / 46.7 & 98.7 / 69.3 / 24.5 \\ 
& DeSTSeg$^*$~\cite{destseg} & 92.9 / 98.4 / 92.2 & 53.0 & 84.2 / 42.0 / 22.8 & \pzo8.8 / \pzo4.7 / \pzo4.6 / 12.9 & 94.5 / 49.7 / \pzo6.0 \\ 
& SimpleNet$^*$~\cite{simplenet} & 96.9 / 99.3 / 96.1 & 85.2 & 98.0 / 83.1 / 79.4 & 40.4 / 54.4 / 29.4 / 65.9 & 97.4 / 86.8 / 41.4 \\ 
& DiAD~\cite{diad} & 95.1 / 99.1 / 94.4 & 91.6 & 96.2 / 60.7 / 60.0 & 53.7 / 50.6 / 38.3 / 57.1 & 96.2 / 72.3 / 47.5 \\ 
& InvAD-lite & 99.7 / 99.9 / 99.5 & 93.4 & 97.3 / 78.6 / 80.4 & 43.9 / 53.2 / 32.0 / 67.2 & 99.7 / 85.4 / 43.0 \\ 
& InvAD & 100. / 100. / 100. & 93.4 & 97.8 / 81.5 / 82.0 & 44.5 / 50.8 / 33.6 / 69.4 & 100. / 87.1 / 43.0 \\ 
            
            \hline
            \multirow{8}{*}{\makecell[c]{\rotatebox{90}{\textbf{pill}}}} & DRÆM$^*$~\cite{draem} & 76.2 / 94.9 / 92.5 & 53.9 & 93.9 / 59.2 / 44.1 & \pzo0.5 / \pzo0.3 / \pzo0.3 / 28.3 & 87.9 / 65.7 / \pzo0.4 \\ 
& RD$^*$~\cite{rd} & 97.5 / 99.6 / 96.8 & 95.8 & 97.5 / 63.4 / 65.2 & 25.9 / 37.8 / 17.1 / 48.4 & 98.0 / 75.4 / 26.9 \\ 
& UniAD$^\dagger$~\cite{uniad} & 94.4 / 99.0 / 95.4 & 95.3 & 96.1 / 44.5 / 52.4 & 18.1 / 39.4 / 11.0 / 35.5 & 96.3 / 64.3 / 22.8 \\ 
& DeSTSeg$^*$~\cite{destseg} & 77.1 / 94.4 / 91.7 & 27.9 & 96.2 / 61.7 / 41.8 & 10.6 / \pzo5.8 / \pzo5.7 / 26.4 & 87.7 / 66.6 / \pzo7.4 \\ 
& SimpleNet$^*$~\cite{simplenet} & 88.2 / 97.7 / 92.5 & 81.9 & 96.5 / 72.4 / 67.7 & 33.5 / 32.4 / 22.2 / 51.2 & 92.8 / 78.9 / 29.4 \\ 
& DiAD~\cite{diad} & 99.4 / 99.9 / 98.3 & 90.6 & 98.6 / 42.2 / 46.4 & 24.8 / 53.4 / 14.9 / 33.0 & 99.2 / 62.4 / 31.0 \\ 
& InvAD-lite & 97.0 / 99.5 / 96.8 & 95.9 & 97.9 / 69.5 / 69.3 & 29.1 / 35.6 / 19.9 / 53.1 & 97.8 / 78.9 / 28.2 \\ 
& InvAD & 98.5 / 99.7 / 97.8 & 96.2 & 98.1 / 69.9 / 69.8 & 34.9 / 43.2 / 24.0 / 53.6 & 98.7 / 79.3 / 34.0 \\ 
            
            \hline
            \multirow{8}{*}{\makecell[c]{\rotatebox{90}{\textbf{screw}}}} & DRÆM$^*$~\cite{draem} & 87.7 / 95.7 / 89.9 & 55.2 & 90.0 / 33.8 / 40.7 & 13.8 / \pzo8.3 / \pzo7.8 / 25.5 & 91.1 / 54.8 / 10.0 \\ 
& RD$^*$~\cite{rd} & 97.7 / 99.3 / 95.8 & 96.8 & 99.4 / 40.2 / 44.7 & 22.7 / 63.9 / 13.7 / 28.7 & 97.6 / 61.4 / 33.4 \\ 
& UniAD$^\dagger$~\cite{uniad} & 95.3 / 98.5 / 92.9 & 95.2 & 99.2 / 37.4 / 42.3 & 16.5 / 54.1 / \pzo9.7 / 26.8 & 95.6 / 59.6 / 26.8 \\ 
& DeSTSeg$^*$~\cite{destseg} & 69.9 / 88.4 / 85.4 & 47.3 & 93.8 / 19.9 / 25.3 & 14.6 / \pzo9.4 / \pzo8.1 / 14.5 & 81.2 / 46.3 / 10.7 \\ 
& SimpleNet$^*$~\cite{simplenet} & 76.7 / 90.5 / 87.7 & 84.0 & 96.5 / 15.9 / 23.2 & \pzo8.8 / 65.2 / \pzo4.8 / 13.1 & 85.0 / 45.2 / 26.3 \\ 
& DiAD~\cite{diad} & 98.5 / 99.8 / 97.7 & 94.0 & 96.6 / 66.0 / 64.1 & \pzo3.3 / 43.1 / \pzo1.7 / \pzo4.3 & 98.7 / 75.6 / 16.0 \\ 
& InvAD-lite & 92.4 / 96.6 / 93.8 & 97.6 & 99.5 / 49.3 / 51.5 & 27.2 / 58.8 / 16.9 / 34.6 & 94.3 / 66.8 / 34.3 \\ 
& InvAD & 97.2 / 99.0 / 95.5 & 97.5 & 99.6 / 48.6 / 50.6 & 28.1 / 50.7 / 17.4 / 33.9 & 97.2 / 66.3 / 32.1 \\ 
            
            \hline
            \multirow{8}{*}{\makecell[c]{\rotatebox{90}{\textbf{toothbrush}}}} & DRÆM$^*$~\cite{draem} & 90.8 / 96.8 / 90.0 & 68.9 & 97.3 / 55.2 / 55.8 & 21.3 / 13.3 / 12.5 / 38.7 & 92.5 / 69.4 / 15.7 \\ 
& RD$^*$~\cite{rd} & 97.2 / 99.0 / 94.7 & 92.0 & 99.0 / 53.6 / 58.8 & 32.5 / 65.9 / 20.8 / 41.6 & 97.0 / 70.5 / 39.7 \\ 
& UniAD$^\dagger$~\cite{uniad} & 89.7 / 95.3 / 95.2 & 87.9 & 98.4 / 37.8 / 49.1 & 25.0 / 43.9 / 15.2 / 32.6 & 93.4 / 61.8 / 28.0 \\ 
& DeSTSeg$^*$~\cite{destseg} & 71.7 / 89.3 / 84.5 & 30.9 & 96.2 / 52.9 / 58.8 & 39.7 / 29.9 / 25.6 / 41.6 & 81.8 / 69.3 / 31.7 \\ 
& SimpleNet$^*$~\cite{simplenet} & 89.7 / 95.7 / 92.3 & 87.4 & 98.4 / 46.9 / 52.5 & 31.5 / 64.0 / 19.7 / 35.6 & 92.6 / 65.9 / 38.4 \\ 
& DiAD~\cite{diad} & 99.8 / 99.7 / 97.6 & 91.3 & 98.8 / 56.1 / 62.3 & 11.2 / 46.8 / \pzo6.0 / \pzo9.7 & 99.0 / 72.4 / 21.3 \\ 
& InvAD-lite & 96.9 / 98.8 / 95.2 & 91.5 & 98.9 / 46.1 / 58.4 & 34.7 / 58.2 / 22.5 / 41.2 & 97.0 / 67.8 / 38.5 \\ 
& InvAD & 94.7 / 97.8 / 95.2 & 91.2 & 99.1 / 55.4 / 60.8 & 42.8 / 70.3 / 28.4 / 43.7 & 95.9 / 71.8 / 47.2 \\ 
            
            \hline
            \multirow{8}{*}{\makecell[c]{\rotatebox{90}{\textbf{transistor}}}} & DRÆM$^*$~\cite{draem} & 77.2 / 77.4 / 71.1 & 39.0 & 68.0 / 23.6 / 15.1 & \pzo2.4 / \pzo1.2 / \pzo1.2 / \pzo8.2 & 75.2 / 35.6 / \pzo1.6 \\ 
& RD$^*$~\cite{rd} & 94.2 / 95.2 / 90.0 & 74.7 & 85.9 / 42.3 / 45.2 & 23.5 / 29.1 / 14.2 / 29.2 & 93.1 / 57.8 / 22.3 \\ 
& UniAD$^\dagger$~\cite{uniad} & 99.8 / 99.8 / 97.5 & 93.5 & 97.4 / 61.2 / 63.0 & 23.0 / 28.0 / 15.2 / 46.0 & 99.0 / 73.9 / 22.1 \\ 
& DeSTSeg$^*$~\cite{destseg} & 78.2 / 79.5 / 68.8 & 43.9 & 73.6 / 38.4 / 39.2 & 19.2 / 10.9 / 10.7 / 24.4 & 75.5 / 50.4 / 13.6 \\ 
& SimpleNet$^*$~\cite{simplenet} & 99.2 / 98.7 / 97.6 & 83.2 & 95.8 / 58.2 / 56.0 & 29.8 / 39.5 / 18.8 / 38.9 & 98.5 / 70.0 / 29.4 \\ 
& DiAD~\cite{diad} & 96.8 / 99.9 / 98.4 & 90.7 & 92.4 / 65.7 / 64.1 & 44.7 / 47.9 / 30.4 / 51.9 & 98.4 / 74.1 / 41.0 \\ 
& InvAD-lite & 98.3 / 97.9 / 92.3 & 86.8 & 94.6 / 60.0 / 57.3 & 30.1 / 29.1 / 19.5 / 40.2 & 96.2 / 70.6 / 26.2 \\ 
& InvAD & 99.8 / 99.8 / 97.6 & 94.3 & 97.3 / 70.7 / 68.5 & 36.5 / 34.7 / 24.9 / 52.1 & 99.1 / 78.8 / 32.0 \\ 
            
            \hline
            \multirow{8}{*}{\makecell[c]{\rotatebox{90}{\textbf{zipper}}}} & DRÆM$^*$~\cite{draem} & 99.9 / 100. / 99.2 & 91.9 & 98.6 / 74.3 / 69.3 & 27.5 / 18.1 / 17.2 / 53.0 & 99.7 / 80.7 / 20.9 \\ 
& RD$^*$~\cite{rd} & 99.5 / 99.9 / 99.2 & 94.1 & 98.5 / 53.9 / 60.3 & 25.8 / 32.6 / 16.7 / 43.1 & 99.5 / 70.9 / 25.0 \\ 
& UniAD$^\dagger$~\cite{uniad} & 98.6 / 99.6 / 97.1 & 92.6 & 98.0 / 45.0 / 51.9 & 17.9 / 23.3 / 11.1 / 35.0 & 98.4 / 65.0 / 17.4 \\ 
& DeSTSeg$^*$~\cite{destseg} & 88.4 / 96.3 / 93.1 & 66.9 & 97.3 / 64.7 / 59.2 & 17.4 / 10.1 / 10.0 / 42.1 & 92.6 / 73.7 / 12.5 \\ 
& SimpleNet$^*$~\cite{simplenet} & 99.0 / 99.7 / 98.3 & 90.7 & 97.9 / 53.4 / 54.6 & 28.4 / 50.0 / 17.9 / 37.5 & 99.0 / 68.6 / 32.1 \\ 
& DiAD~\cite{diad} & 99.7 / 100. / 100. & 97.5 & 93.3 / 43.3 / 43.5 & 10.1 / 31.1 / \pzo5.5 / 13.8 & 99.9 / 60.0 / 15.6 \\ 
& InvAD-lite & 98.9 / 99.7 / 97.5 & 95.3 & 98.6 / 59.7 / 62.7 & 32.5 / 41.5 / 21.3 / 45.6 & 98.7 / 73.7 / 31.8 \\ 
& InvAD & 99.7 / 99.9 / 99.2 & 94.6 & 98.4 / 51.2 / 58.0 & 23.3 / 24.4 / 14.9 / 40.9 & 99.6 / 69.2 / 20.9 \\ 
            \toprule[0.12em] 
        \end{tabular}
    }
\end{table*}

\section{Quantitative Results for Each Category on VisA} \label{sec:app_visa}
\Tab\ref{tab:supp_visa_complex_structure}, \Tab\ref{tab:supp_visa_multiple_instance}, and \Tab\ref{tab:supp_visa_single_instance} show detailed quantitative single-class results for Complex Structure, Multiple Instance Object, and Single Instance categories on VisA dataset.

\begin{table*}[htp]
    \centering
    \caption{\textbf{Quantitative results for each class (Complex Structure) on VisA dataset.} These are respectively the image-level mAU-ROC/mAP/m$F_1$-max, region-level \redzjn{mAU-PRO}, pixel-level mAU-ROC/mAP/m$F_1$-max, the proposed pixel-level m$F_1$$^{.2}_{.8}$/mAcc$^{.2}_{.8}$/mIoU$^{.2}_{.8}$/\redzjn{mIoU-max}, and averaged \redzjn{mAD$_I$}/\redzjn{mAD$_P$}/\redzjn{mAD$^{.2}_{.8}$}.}
    \label{tab:supp_visa_complex_structure}
    \renewcommand{\arraystretch}{1.0}
    \setlength\tabcolsep{6.0pt}
    \resizebox{1.0\linewidth}{!}{
        \begin{tabular}{p{0.3cm}<{\centering} p{2.0cm}<{\centering} p{3.9cm}<{\centering} p{1.6cm}<{\centering} p{3.9cm}<{\centering} p{4.9cm}<{\centering} p{3.3cm}<{\centering}}
            \toprule[0.17em]
            & Method & mAU-ROC/mAP/m$F_1$-max & \redzjn{mAU-PRO} & mAU-ROC/mAP/m$F_1$-max & m$F_1$$^{.2}_{.8}$/mAcc$^{.2}_{.8}$/mIoU$^{.2}_{.8}$/\redzjn{mIoU-max} & \redzjn{mAD$_I$}/\redzjn{mAD$_P$}/\redzjn{mAD$^{.2}_{.8}$} \\
            \hline
            \multirow{8}{*}{\makecell[c]{\rotatebox{90}{\textbf{pcb1}}}} & DRÆM$^*$~\cite{draem} & 71.9 / 72.3 / 70.0 & 52.9 & 94.7 / 31.9 / 37.3 & 15.3 / \pzo9.6 / \pzo8.7 / 22.9 & 71.4 / 54.6 / 11.2 \\ 
& RD$^*$~\cite{rd} & 96.2 / 95.5 / 91.9 & 95.8 & 99.4 / 66.2 / 62.4 & 35.6 / 61.2 / 23.4 / 45.3 & 94.5 / 76.0 / 40.1 \\ 
& UniAD$^\dagger$~\cite{uniad} & 94.2 / 92.9 / 90.8 & 88.8 & 99.2 / 59.6 / 59.6 & 29.5 / 51.3 / 19.1 / 42.5 & 92.6 / 72.8 / 33.3 \\ 
& DeSTSeg$^*$~\cite{destseg} & 87.6 / 83.1 / 83.7 & 83.2 & 95.8 / 46.4 / 49.0 & 32.4 / 52.1 / 17.0 / 32.4 & 84.8 / 63.7 / 33.8 \\ 
& SimpleNet$^*$~\cite{simplenet} & 91.6 / 91.9 / 86.0 & 83.6 & 99.2 / 86.1 / 78.8 & 38.8 / 51.4 / 27.5 / 65.1 & 89.8 / 88.0 / 39.2 \\ 
& DiAD~\cite{diad} & 88.1 / 88.7 / 80.7 & 80.2 & 98.7 / 49.6 / 52.8 & \pzo1.4 / 39.1 / \pzo0.7 / \pzo1.3 & 85.8 / 67.0 / 13.7 \\ 
& InvAD-lite & 95.7 / 93.4 / 92.3 & 96.4 & 99.8 / 78.9 / 72.7 & 47.3 / 57.6 / 32.9 / 57.1 & 93.8 / 83.8 / 45.9 \\ 
& InvAD & 96.4 / 96.1 / 93.1 & 95.6 & 99.8 / 80.7 / 76.2 & 47.6 / 57.2 / 33.6 / 61.6 & 95.2 / 85.6 / 46.1 \\ 
            
            \hline
            \multirow{8}{*}{\makecell[c]{\rotatebox{90}{\textbf{pcb2}}}} & DRÆM$^*$~\cite{draem} & 78.5 / 78.3 / 76.3 & 66.2 & 92.3 / 10.0 / 18.6 & \pzo7.1 / \pzo5.1 / \pzo3.8 / 10.3 & 77.7 / 40.3 / \pzo5.3 \\ 
& RD$^*$~\cite{rd} & 97.8 / 97.8 / 94.2 & 90.8 & 98.0 / 22.3 / 30.0 & 16.5 / 51.8 / \pzo9.3 / 17.7 & 96.6 / 50.1 / 25.9 \\ 
& UniAD$^\dagger$~\cite{uniad} & 91.1 / 91.6 / 85.1 & 82.2 & 98.0 / \pzo9.2 / 16.9 & \pzo9.9 / 45.8 / \pzo5.3 / \pzo9.2 & 89.3 / 41.4 / 20.3 \\ 
& DeSTSeg$^*$~\cite{destseg} & 86.5 / 85.8 / 82.6 & 79.9 & 97.3 / 14.6 / 28.2 & 25.0 / 55.4 / 11.3 / 16.4 & 85.0 / 46.7 / 30.6 \\ 
& SimpleNet$^*$~\cite{simplenet} & 92.4 / 93.3 / 84.5 & 85.7 & 96.6 / \pzo8.9 / 18.6 & \pzo8.4 / 58.4 / \pzo4.5 / 10.3 & 90.1 / 41.4 / 23.8 \\ 
& DiAD~\cite{diad} & 91.4 / 91.4 / 84.7 & 67.0 & 95.2 / \pzo7.5 / 16.7 & \pzo7.3 / 59.1 / \pzo3.9 / \pzo8.9 & 89.2 / 39.8 / 23.4 \\ 
& InvAD-lite & 94.5 / 94.6 / 88.8 & 90.9 & 98.9 / 11.7 / 22.8 & 15.1 / 43.5 / \pzo8.3 / 12.9 & 92.6 / 44.5 / 22.3 \\ 
& InvAD & 97.1 / 97.1 / 92.9 & 91.6 & 99.0 / 15.2 / 24.0 & 16.1 / 37.7 / \pzo8.9 / 13.6 & 95.7 / 46.1 / 20.9 \\ 
            
            \hline
            \multirow{8}{*}{\makecell[c]{\rotatebox{90}{\textbf{pcb3}}}} & DRÆM$^*$~\cite{draem} & 76.6 / 77.5 / 74.8 & 43.0 & 90.8 / 14.1 / 24.4 & 12.1 / \pzo9.5 / \pzo6.5 / 13.9 & 76.3 / 43.1 / \pzo9.4 \\ 
& RD$^*$~\cite{rd} & 96.4 / 96.2 / 91.0 & 93.9 & 97.9 / 26.2 / 35.2 & 21.2 / 56.6 / 12.3 / 21.4 & 94.5 / 53.1 / 30.0 \\ 
& UniAD$^\dagger$~\cite{uniad} & 82.2 / 83.2 / 77.5 & 79.3 & 98.2 / 13.3 / 24.0 & 13.8 / 50.4 / \pzo7.6 / 13.6 & 81.0 / 45.2 / 23.9 \\ 
& DeSTSeg$^*$~\cite{destseg} & 93.7 / 95.1 / 87.0 & 62.4 & 97.7 / 28.1 / 33.4 & 31.9 / 57.5 / 15.3 / 20.0 & 91.9 / 53.1 / 34.9 \\ 
& SimpleNet$^*$~\cite{simplenet} & 89.1 / 91.1 / 82.6 & 85.1 & 97.2 / 31.0 / 36.1 & 17.7 / 57.1 / 10.2 / 22.0 & 87.6 / 54.8 / 28.3 \\ 
& DiAD~\cite{diad} & 86.2 / 87.6 / 77.6 & 68.9 & 96.7 / \pzo8.0 / 18.8 & 17.9 / 56.1 / 10.7 / 33.4 & 83.8 / 41.2 / 28.2 \\ 
& InvAD-lite & 94.7 / 95.0 / 87.2 & 92.2 & 99.2 / 17.0 / 29.0 & 20.3 / 48.0 / 11.5 / 17.0 & 92.3 / 48.4 / 26.6 \\ 
& InvAD & 97.0 / 97.4 / 92.4 & 92.0 & 99.2 / 16.0 / 28.4 & 20.6 / 49.6 / 11.6 / 16.6 & 95.6 / 47.9 / 27.3 \\ 
            
            \hline
            \multirow{8}{*}{\makecell[c]{\rotatebox{90}{\textbf{pcb4}}}} & DRÆM$^*$~\cite{draem} & 97.4 / 97.6 / 93.5 & 75.7 & 94.4 / 31.0 / 37.6 & \pzo9.0 / \pzo5.5 / \pzo5.0 / 23.2 & 96.2 / 54.3 / \pzo6.5 \\ 
& RD$^*$~\cite{rd} & 99.9 / 99.9 / 99.0 & 88.7 & 97.8 / 31.4 / 37.0 & 17.9 / 39.9 / 10.4 / 22.7 & 99.6 / 55.4 / 22.7 \\ 
& UniAD$^\dagger$~\cite{uniad} & 99.0 / 99.1 / 95.5 & 82.9 & 97.2 / 29.4 / 33.5 & 17.2 / 41.5 / \pzo9.9 / 20.1 & 97.9 / 53.4 / 22.9 \\ 
& DeSTSeg$^*$~\cite{destseg} & 97.8 / 97.8 / 92.7 & 76.9 & 95.8 / 53.0 / 53.2 & 25.8 / 54.1 / 17.4 / 36.2 & 96.1 / 67.3 / 32.4 \\ 
& SimpleNet$^*$~\cite{simplenet} & 97.0 / 97.0 / 93.5 & 61.1 & 93.9 / 23.9 / 32.9 & 10.1 / 30.5 / \pzo5.7 / 19.7 & 95.8 / 50.2 / 15.4 \\ 
& DiAD~\cite{diad} & 99.6 / 99.5 / 97.0 & 85.0 & 97.0 / 17.6 / 27.2 & 12.2 / 52.3 / \pzo7.0 / 18.0 & 98.7 / 47.3 / 23.8 \\ 
& InvAD-lite & 99.8 / 99.8 / 98.0 & 89.6 & 98.8 / 51.1 / 49.8 & 27.5 / 33.3 / 17.2 / 33.2 & 99.2 / 66.6 / 26.0 \\ 
& InvAD & 99.8 / 99.8 / 98.0 & 89.9 & 98.6 / 46.6 / 45.2 & 29.8 / 37.6 / 18.3 / 29.2 & 99.2 / 63.5 / 28.6 \\ 
            \toprule[0.12em] 
        \end{tabular}
    }
\end{table*}

\begin{table*}[htp]
    \centering
    \caption{\textbf{Quantitative results for each class (Multiple Instances) on VisA dataset.} These are respectively the image-level mAU-ROC/mAP/m$F_1$-max, region-level \redzjn{mAU-PRO}, pixel-level mAU-ROC/mAP/m$F_1$-max, the proposed pixel-level m$F_1$$^{.2}_{.8}$/mAcc$^{.2}_{.8}$/mIoU$^{.2}_{.8}$/\redzjn{mIoU-max}, and averaged \redzjn{mAD$_I$}/\redzjn{mAD$_P$}/\redzjn{mAD$^{.2}_{.8}$}.}
    \label{tab:supp_visa_multiple_instance}
    \renewcommand{\arraystretch}{1.0}
    \setlength\tabcolsep{6.0pt}
    \resizebox{1.0\linewidth}{!}{
        \begin{tabular}{p{0.3cm}<{\centering} p{2.0cm}<{\centering} p{3.9cm}<{\centering} p{1.6cm}<{\centering} p{3.9cm}<{\centering} p{4.9cm}<{\centering} p{3.3cm}<{\centering}}
            \toprule[0.17em]
            & Method & mAU-ROC/mAP/m$F_1$-max & \redzjn{mAU-PRO} & mAU-ROC/mAP/m$F_1$-max & m$F_1$$^{.2}_{.8}$/mAcc$^{.2}_{.8}$/mIoU$^{.2}_{.8}$/\redzjn{mIoU-max} & \redzjn{mAD$_I$}/\redzjn{mAD$_P$}/\redzjn{mAD$^{.2}_{.8}$} \\
            \hline
            \multirow{8}{*}{\makecell[c]{\rotatebox{90}{\textbf{macaroni1}}}} & DRÆM$^*$~\cite{draem} & 69.8 / 68.6 / 70.9 & 67.0 & 95.0 / 19.1 / 24.1 & 15.9 / \pzo9.9 / \pzo8.7 / 13.7 & 69.8 / 46.1 / 11.5 \\ 
& RD$^*$~\cite{rd} & 75.9 / 61.5 / 76.8 & 95.3 & 99.4 / \pzo2.9 / \pzo6.9 & \pzo2.3 / \pzo7.7 / \pzo1.1 / \pzo3.6 & 71.4 / 36.4 / \pzo3.7 \\ 
& UniAD$^\dagger$~\cite{uniad} & 82.8 / 79.3 / 75.7 & 92.6 & 99.0 / \pzo7.6 / 16.1 & \pzo7.1 / 37.0 / \pzo3.8 / \pzo8.7 & 79.3 / 40.9 / 16.0 \\ 
& DeSTSeg$^*$~\cite{destseg} & 76.6 / 69.0 / 71.0 & 62.4 & 99.1 / \pzo5.8 / 13.4 & 11.0 / 18.8 / \pzo5.8 / \pzo7.2 & 72.2 / 39.4 / 11.9 \\ 
& SimpleNet$^*$~\cite{simplenet} & 85.9 / 82.5 / 73.1 & 92.0 & 98.9 / \pzo3.5 / \pzo8.4 & \pzo4.1 / 52.3 / \pzo2.1 / \pzo4.4 & 80.5 / 36.9 / 19.5 \\ 
& DiAD~\cite{diad} & 85.7 / 85.2 / 78.8 & 68.5 & 94.1 / 10.2 / 16.7 & 41.9 / 59.6 / 27.8 / 44.7 & 83.2 / 40.3 / 43.1 \\ 
& InvAD-lite & 93.8 / 91.4 / 87.8 & 96.9 & 99.7 / 19.8 / 27.3 & 17.5 / 42.5 / \pzo9.8 / 15.8 & 91.0 / 48.9 / 23.3 \\ 
& InvAD & 94.1 / 92.5 / 88.0 & 96.6 & 99.7 / 20.8 / 29.4 & 16.9 / 42.4 / \pzo9.5 / 17.2 & 91.5 / 50.0 / 22.9 \\ 
            
            \hline
            \multirow{8}{*}{\makecell[c]{\rotatebox{90}{\textbf{macaroni2}}}} & DRÆM$^*$~\cite{draem} & 59.4 / 60.7 / 68.1 & 65.3 & 94.6 / \pzo3.9 / 12.5 & \pzo2.1 / \pzo1.2 / \pzo1.1 / \pzo6.6 & 62.7 / 37.0 / \pzo1.5 \\ 
& RD$^*$~\cite{rd} & 88.3 / 84.5 / 83.8 & 97.4 & 99.7 / 13.2 / 21.8 & \pzo9.8 / 61.9 / \pzo5.3 / 12.3 & 85.5 / 44.9 / 25.7 \\ 
& UniAD$^\dagger$~\cite{uniad} & 76.0 / 75.8 / 70.2 & 87.0 & 97.3 / \pzo5.1 / 12.2 & \pzo4.7 / 46.1 / \pzo2.5 / \pzo6.5 & 74.0 / 38.2 / 17.8 \\ 
& DeSTSeg$^*$~\cite{destseg} & 68.9 / 62.1 / 67.7 & 70.0 & 98.5 / \pzo6.3 / 14.4 & 10.8 / 13.4 / \pzo5.7 / \pzo7.7 & 66.2 / 39.7 / 10.0 \\ 
& SimpleNet$^*$~\cite{simplenet} & 68.3 / 54.3 / 59.7 & 77.8 & 93.2 / \pzo0.6 / \pzo3.9 & \pzo1.0 / 59.6 / \pzo0.5 / \pzo2.0 & 60.8 / 32.6 / 20.4 \\ 
& DiAD~\cite{diad} & 62.5 / 57.4 / 69.6 & 73.1 & 93.6 / \pzo0.9 / \pzo2.8 & \pzo0.0 / \pzo8.4 / \pzo0.0 / \pzo0.0 & 63.2 / 32.4 / \pzo2.8 \\ 
& InvAD-lite & 88.0 / 85.1 / 81.9 & 97.8 & 99.7 / 12.2 / 18.8 & 11.8 / 46.0 / \pzo6.4 / 10.4 & 85.0 / 43.6 / 21.4 \\ 
& InvAD & 85.8 / 83.2 / 80.6 & 97.2 & 99.6 / 11.4 / 18.6 & 10.8 / 45.0 / \pzo5.8 / 10.2 & 83.2 / 43.2 / 20.5 \\ 
            
            \hline
            \multirow{8}{*}{\makecell[c]{\rotatebox{90}{\textbf{capsules}}}} & DRÆM$^*$~\cite{draem} & 83.4 / 91.1 / 82.1 & 62.9 & 97.1 / 27.8 / 33.8 & 22.5 / 17.9 / 12.9 / 20.3 & 85.5 / 52.9 / 17.8 \\ 
& RD$^*$~\cite{rd} & 82.2 / 90.4 / 81.3 & 93.1 & 99.4 / 60.4 / 60.8 & 27.6 / 66.5 / 18.0 / 43.7 & 84.6 / 73.5 / 37.4 \\ 
& UniAD$^\dagger$~\cite{uniad} & 70.3 / 83.2 / 77.8 & 72.2 & 97.4 / 40.4 / 44.7 & 13.8 / 62.4 / \pzo8.2 / 28.8 & 77.1 / 60.8 / 28.1 \\ 
& DeSTSeg$^*$~\cite{destseg} & 87.1 / 93.0 / 84.2 & 76.7 & 96.9 / 33.2 / 39.1 & 24.4 / 37.9 / 15.9 / 18.3 & 88.1 / 56.4 / 26.1 \\ 
& SimpleNet$^*$~\cite{simplenet} & 74.1 / 82.8 / 74.6 & 73.7 & 97.1 / 52.9 / 53.3 & 19.6 / 55.4 / 12.3 / 36.3 & 77.2 / 67.8 / 29.1 \\ 
& DiAD~\cite{diad} & 58.2 / 69.0 / 78.5 & 77.9 & 97.3 / 10.0 / 21.0 & \pzo0.0 / 27.1 / \pzo0.0 / \pzo0.0 & 68.6 / 42.8 / \pzo9.0 \\ 
& InvAD-lite & 90.7 / 94.0 / 88.8 & 94.7 & 99.3 / 59.7 / 58.1 & 33.8 / 61.9 / 21.9 / 40.9 & 91.2 / 72.4 / 39.2 \\ 
& InvAD & 90.5 / 94.3 / 88.3 & 94.7 & 99.7 / 68.1 / 64.7 & 39.2 / 63.0 / 26.5 / 47.8 & 91.0 / 77.5 / 42.9 \\ 
            
            \hline
            \multirow{8}{*}{\makecell[c]{\rotatebox{90}{\textbf{candle}}}} & DRÆM$^*$~\cite{draem} & 69.3 / 73.9 / 68.1 & 65.6 & 82.2 / 10.1 / 19.0 & \pzo4.1 / \pzo2.2 / \pzo2.1 / 10.5 & 70.4 / 37.1 / \pzo2.8 \\ 
& RD$^*$~\cite{rd} & 92.3 / 92.9 / 86.0 & 94.9 & 99.1 / 25.3 / 35.8 & 16.9 / 38.4 / \pzo9.8 / 21.8 & 90.4 / 53.4 / 21.7 \\ 
& UniAD$^\dagger$~\cite{uniad} & 95.8 / 96.2 / 90.0 & 93.0 & 99.0 / 23.6 / 32.6 & 16.0 / 46.7 / \pzo9.1 / 19.5 & 94.0 / 51.7 / 23.9 \\ 
& DeSTSeg$^*$~\cite{destseg} & 94.9 / 94.8 / 89.2 & 69.0 & 98.7 / 39.9 / 45.8 & 30.9 / 37.4 / 21.9 / 29.7 & 93.0 / 61.5 / 30.1 \\ 
& SimpleNet$^*$~\cite{simplenet} & 84.1 / 73.3 / 76.6 & 87.6 & 97.6 / \pzo8.4 / 16.5 & \pzo7.6 / 46.5 / \pzo4.1 / \pzo9.0 & 78.0 / 40.8 / 19.4 \\ 
& DiAD~\cite{diad} & 92.8 / 92.0 / 87.6 & 89.4 & 97.3 / 12.8 / 22.8 & 20.3 / 54.0 / 12.0 / 24.6 & 90.8 / 44.3 / 28.8 \\ 
& InvAD-lite & 96.6 / 96.8 / 91.3 & 95.9 & 99.1 / 21.4 / 30.8 & 18.5 / 34.0 / 10.5 / 18.2 & 94.9 / 50.4 / 21.0 \\ 
& InvAD & 94.8 / 94.1 / 90.6 & 95.4 & 99.3 / 23.6 / 34.5 & 17.8 / 33.7 / 10.2 / 20.9 & 93.2 / 52.5 / 20.6 \\ 
            
            \toprule[0.12em] 
        \end{tabular}
    }
\end{table*}

\begin{table*}[htp]
    \centering
    \caption{\textbf{Quantitative results for each class (Single Instance) on VisA dataset.} These are respectively the image-level mAU-ROC/mAP/m$F_1$-max, region-level \redzjn{mAU-PRO}, pixel-level mAU-ROC/mAP/m$F_1$-max, the proposed pixel-level m$F_1$$^{.2}_{.8}$/mAcc$^{.2}_{.8}$/mIoU$^{.2}_{.8}$/\redzjn{mIoU-max}, and averaged \redzjn{mAD$_I$}/\redzjn{mAD$_P$}/\redzjn{mAD$^{.2}_{.8}$}.}
    \label{tab:supp_visa_single_instance}
    \renewcommand{\arraystretch}{1.0}
    \setlength\tabcolsep{6.0pt}
    \resizebox{1.0\linewidth}{!}{
        \begin{tabular}{p{0.3cm}<{\centering} p{2.0cm}<{\centering} p{3.9cm}<{\centering} p{1.6cm}<{\centering} p{3.9cm}<{\centering} p{4.9cm}<{\centering} p{3.3cm}<{\centering}}
            \toprule[0.17em]
            & Method & mAU-ROC/mAP/m$F_1$-max & \redzjn{mAU-PRO} & mAU-ROC/mAP/m$F_1$-max & m$F_1$$^{.2}_{.8}$/mAcc$^{.2}_{.8}$/mIoU$^{.2}_{.8}$/\redzjn{mIoU-max} & \redzjn{mAD$_I$}/\redzjn{mAD$_P$}/\redzjn{mAD$^{.2}_{.8}$} \\
            \hline
            \multirow{8}{*}{\makecell[c]{\rotatebox{90}{\textbf{cashew}}}} & DRÆM$^*$~\cite{draem} & 81.7 / 89.7 / 87.3 & 38.5 & 80.7 / \pzo9.9 / 15.8 & \pzo1.6 / \pzo0.8 / \pzo0.8 / \pzo8.6 & 86.2 / 35.5 / \pzo1.1 \\ 
& RD$^*$~\cite{rd} & 92.0 / 95.8 / 90.7 & 86.2 & 91.7 / 44.2 / 49.7 & 20.2 / 29.4 / 12.5 / 33.0 & 92.8 / 61.9 / 20.7 \\ 
& UniAD$^\dagger$~\cite{uniad} & 94.3 / 97.2 / 91.1 & 88.5 & 99.0 / 56.2 / 58.9 & 25.2 / 33.1 / 16.4 / 41.8 & 94.2 / 71.4 / 24.9 \\ 
& DeSTSeg$^*$~\cite{destseg} & 92.0 / 96.1 / 88.1 & 66.3 & 87.9 / 47.6 / 52.1 & 32.4 / 30.5 / 25.1 / 35.2 & 92.1 / 62.5 / 29.3 \\ 
& SimpleNet$^*$~\cite{simplenet} & 88.0 / 91.3 / 84.7 & 84.1 & 98.9 / 68.9 / 66.0 & 31.3 / 52.5 / 20.9 / 49.2 & 88.0 / 77.9 / 34.9 \\ 
& DiAD~\cite{diad} & 91.5 / 95.7 / 89.7 & 61.8 & 90.9 / 53.1 / 60.9 & \pzo6.3 / 52.7 / \pzo3.3 / \pzo6.5 & 92.3 / 68.3 / 20.8 \\ 
& InvAD-lite & 92.3 / 96.2 / 89.3 & 89.7 & 94.7 / 52.9 / 56.0 & 30.2 / 37.9 / 19.3 / 38.9 & 92.6 / 67.9 / 29.1 \\ 
& InvAD & 96.6 / 98.5 / 94.9 & 90.3 & 96.4 / 59.1 / 61.5 & 36.8 / 44.8 / 24.3 / 44.4 & 96.7 / 72.3 / 35.3 \\ 
            
            \hline
            \multirow{8}{*}{\makecell[c]{\rotatebox{90}{\textbf{chewinggum}}}} & DRÆM$^*$~\cite{draem} & 93.7 / 97.2 / 91.0 & 41.0 & 91.1 / 62.4 / 63.3 & 44.5 / 33.8 / 30.0 / 46.4 & 94.0 / 72.3 / 36.1 \\ 
& RD$^*$~\cite{rd} & 94.9 / 97.5 / 92.1 & 76.9 & 98.7 / 59.9 / 61.7 & 29.5 / 63.9 / 19.4 / 44.6 & 94.8 / 73.4 / 37.6 \\ 
& UniAD$^\dagger$~\cite{uniad} & 97.5 / 98.9 / 96.4 & 85.0 & 99.1 / 59.5 / 58.0 & 27.3 / 58.8 / 17.6 / 40.8 & 97.6 / 72.2 / 34.6 \\ 
& DeSTSeg$^*$~\cite{destseg} & 95.8 / 98.3 / 94.7 & 68.3 & 98.8 / 86.9 / 81.0 & 38.7 / 68.8 / 31.5 / 48.0 & 96.3 / 88.9 / 46.3 \\ 
& SimpleNet$^*$~\cite{simplenet} & 96.4 / 98.2 / 93.8 & 78.3 & 97.9 / 26.8 / 29.8 & 17.8 / 59.7 / 10.1 / 17.5 & 96.1 / 51.5 / 29.2 \\ 
& DiAD~\cite{diad} & 99.1 / 99.5 / 95.9 & 59.5 & 94.7 / 11.9 / 25.8 & \pzo5.7 / 40.3 / \pzo2.9 / \pzo5.7 & 98.2 / 44.1 / 16.3 \\ 
& InvAD-lite & 97.9 / 99.0 / 95.3 & 79.6 & 98.3 / 55.2 / 57.2 & 28.9 / 61.1 / 18.6 / 40.1 & 97.4 / 70.2 / 36.2 \\ 
& InvAD & 98.3 / 99.1 / 95.9 & 80.7 & 98.7 / 60.0 / 62.5 & 33.6 / 64.6 / 22.1 / 45.5 & 97.8 / 73.7 / 40.1 \\ 
            
            \hline
            \multirow{8}{*}{\makecell[c]{\rotatebox{90}{\textbf{fryum}}}} & DRÆM$^*$~\cite{draem} & 89.2 / 95.0 / 86.6 & 69.5 & 92.4 / 38.8 / 38.6 & \pzo3.5 / \pzo1.8 / \pzo1.8 / 23.9 & 90.3 / 56.6 / \pzo2.4 \\ 
& RD$^*$~\cite{rd} & 95.3 / 97.9 / 91.5 & 93.4 & 97.0 / 47.6 / 51.5 & 24.7 / 35.7 / 15.2 / 34.7 & 94.9 / 65.4 / 25.2 \\ 
& UniAD$^\dagger$~\cite{uniad} & 86.9 / 93.9 / 86.0 & 82.0 & 97.3 / 46.6 / 52.4 & 22.2 / 43.0 / 13.6 / 35.5 & 88.9 / 65.4 / 26.3 \\ 
& DeSTSeg$^*$~\cite{destseg} & 92.1 / 96.1 / 89.5 & 47.7 & 88.1 / 35.2 / 38.5 & 21.6 / 13.0 / 12.3 / 23.8 & 92.6 / 53.9 / 15.6 \\ 
& SimpleNet$^*$~\cite{simplenet} & 88.4 / 93.0 / 83.3 & 85.1 & 93.0 / 39.1 / 45.4 & 19.8 / 30.8 / 11.9 / 29.4 & 88.2 / 59.2 / 20.8 \\ 
& DiAD~\cite{diad} & 89.8 / 95.0 / 87.2 & 81.3 & 97.6 / 58.6 / 60.1 & 11.3 / 54.4 / \pzo6.1 / 10.1 & 90.7 / 72.1 / 23.9 \\ 
& InvAD-lite & 96.2 / 98.3 / 92.7 & 91.7 & 96.9 / 47.5 / 52.0 & 24.8 / 23.3 / 15.5 / 35.1 & 95.7 / 65.5 / 21.2 \\ 
& InvAD & 96.0 / 98.2 / 92.3 & 91.8 & 97.1 / 48.4 / 52.5 & 26.4 / 26.9 / 16.6 / 35.6 & 95.5 / 66.0 / 23.3 \\ 
            
            \hline
            \multirow{8}{*}{\makecell[c]{\rotatebox{90}{\textbf{pipe\_fryum}}}} & DRÆM$^*$~\cite{draem} & 82.8 / 91.2 / 84.0 & 61.9 & 91.1 / 38.2 / 39.7 & 13.2 / \pzo7.4 / \pzo7.2 / 24.8 & 86.0 / 56.3 / \pzo9.3 \\ 
& RD$^*$~\cite{rd} & 97.9 / 98.9 / 96.5 & 95.4 & 99.1 / 56.8 / 58.8 & 32.5 / 44.4 / 21.0 / 41.7 & 97.8 / 71.6 / 32.6 \\ 
& UniAD$^\dagger$~\cite{uniad} & 95.3 / 97.6 / 92.9 & 93.0 & 99.1 / 53.4 / 58.6 & 28.6 / 49.3 / 18.3 / 41.5 & 95.3 / 70.4 / 32.1 \\ 
& DeSTSeg$^*$~\cite{destseg} & 94.1 / 97.1 / 91.9 & 45.9 & 98.9 / 78.8 / 72.7 & 43.4 / 53.1 / 28.9 / 47.1 & 94.4 / 83.5 / 41.8 \\ 
& SimpleNet$^*$~\cite{simplenet} & 90.8 / 95.5 / 88.6 & 83.0 & 98.5 / 65.6 / 63.4 & 33.7 / 52.4 / 22.1 / 46.4 & 91.6 / 75.8 / 36.1 \\ 
& DiAD~\cite{diad} & 96.2 / 98.1 / 93.7 & 89.9 & 99.4 / 72.7 / 69.9 & 33.8 / 50.7 / 21.9 / 41.0 & 96.0 / 80.7 / 35.5 \\ 
& InvAD-lite & 98.7 / 99.3 / 96.0 & 94.4 & 99.0 / 55.6 / 57.3 & 33.9 / 41.4 / 22.0 / 40.1 & 98.0 / 70.6 / 32.4 \\ 
& InvAD & 99.4 / 99.7 / 98.0 & 94.8 & 99.4 / 67.4 / 66.3 & 39.8 / 44.5 / 27.1 / 49.6 & 99.0 / 77.7 / 37.1 \\ 
            \toprule[0.12em] 
        \end{tabular}
    }
\end{table*}

\section{Quantitative Results for Each Category on MVTec 3D AD} \label{sec:app_mvtec3d}
\Tab\ref{tab:supp_mvtec3d} shows detailed quantitative single-class results on MVTec 3D AD dataset.

\begin{table*}[htp]
    \centering
    \caption{\textbf{Quantitative results for each class on MVTec 3D AD dataset.} These are respectively the image-level mAU-ROC/mAP/m$F_1$-max, region-level \redzjn{mAU-PRO}, pixel-level mAU-ROC/mAP/m$F_1$-max, the proposed pixel-level m$F_1$$^{.2}_{.8}$/mAcc$^{.2}_{.8}$/mIoU$^{.2}_{.8}$/\redzjn{mIoU-max}, and averaged \redzjn{mAD$_I$}/\redzjn{mAD$_P$}/\redzjn{mAD$^{.2}_{.8}$}.}
    \label{tab:supp_mvtec3d}
    \renewcommand{\arraystretch}{1.0}
    \setlength\tabcolsep{6.0pt}
    \resizebox{1.0\linewidth}{!}{
        \begin{tabular}{p{0.3cm}<{\centering} p{2.0cm}<{\centering} p{3.9cm}<{\centering} p{1.6cm}<{\centering} p{3.9cm}<{\centering} p{4.9cm}<{\centering} p{3.3cm}<{\centering}}
            \toprule[0.17em]
            & Method & mAU-ROC/mAP/m$F_1$-max & \redzjn{mAU-PRO} & mAU-ROC/mAP/m$F_1$-max & m$F_1$$^{.2}_{.8}$/mAcc$^{.2}_{.8}$/mIoU$^{.2}_{.8}$/\redzjn{mIoU-max} & \redzjn{mAD$_I$}/\redzjn{mAD$_P$}/\redzjn{mAD$^{.2}_{.8}$} \\
            \hline
            \multirow{6}{*}{\makecell[c]{\rotatebox{90}{\textbf{bagel}}}} & RD$^*$~\cite{rd} & 82.5 / 95.4 / 89.6 & 91.3 & 98.6 / 39.0 / 45.1 & 18.9 / 65.4 / 11.3 / 29.1 & 89.2 / 60.9 / 31.9 \\ 
& UniAD$^\dagger$~\cite{uniad} & 82.7 / 95.8 / 89.3 & 84.4 & 97.6 / 30.4 / 35.8 & 13.0 / 51.4 / \pzo7.4 / 21.8 & 89.3 / 54.6 / 23.9 \\ 
& SimpleNet$^*$~\cite{simplenet} & 76.2 / 93.3 / 89.3 & 70.4 & 93.2 / 23.6 / 30.9 & 12.6 / 55.3 / \pzo7.1 / 18.3 & 86.3 / 49.2 / 25.0 \\ 
& DiAD~\cite{diad} & 100. / 100. / 100. & 93.8 & 98.5 / 49.6 / 54.2 & \pzo4.6 / 53.2 / \pzo2.4 / \pzo5.3 & 100. / 67.4 / 20.1 \\ 
& InvAD-lite & 90.9 / 97.6 / 93.0 & 93.6 & 98.8 / 43.0 / 46.6 & 21.1 / 68.4 / 12.8 / 30.4 & 93.8 / 62.8 / 34.1 \\ 
& InvAD & 93.5 / 98.3 / 93.8 & 94.7 & 99.1 / 44.9 / 48.7 & 24.0 / 62.2 / 14.6 / 32.2 & 95.2 / 64.2 / 33.6 \\ 
            
            \hline
            \multirow{6}{*}{\makecell[c]{\rotatebox{90}{\textbf{cable\_gland}}}} & RD$^*$~\cite{rd} & 90.1 / 97.5 / 92.6 & 98.2 & 99.4 / 37.9 / 43.2 & 18.6 / 63.9 / 11.0 / 27.5 & 93.4 / 60.2 / 31.2 \\ 
& UniAD$^\dagger$~\cite{uniad} & 89.8 / 97.2 / 93.9 & 96.3 & 98.9 / 26.4 / 34.8 & 11.7 / 47.5 / \pzo6.6 / 21.1 & 93.6 / 53.4 / 21.9 \\ 
& SimpleNet$^*$~\cite{simplenet} & 70.3 / 91.0 / 90.2 & 86.8 & 95.2 / 14.4 / 23.1 & \pzo9.1 / 68.2 / \pzo4.9 / 13.0 & 83.8 / 44.2 / 27.4 \\ 
& DiAD~\cite{diad} & 68.1 / 91.0 / 92.3 & 94.5 & 98.4 / 25.2 / 32.0 & \pzo4.8 / 44.3 / \pzo2.5 / \pzo5.2 & 83.8 / 51.9 / 17.2 \\ 
& InvAD-lite & 94.4 / 98.7 / 95.4 & 98.5 & 99.4 / 41.1 / 46.8 & 23.6 / 54.0 / 14.3 / 30.5 & 96.2 / 62.4 / 30.6 \\ 
& InvAD & 97.3 / 99.3 / 96.7 & 98.8 & 99.6 / 41.5 / 47.9 & 25.6 / 49.9 / 15.5 / 31.5 & 97.8 / 63.0 / 30.3 \\ 

            \hline
            \multirow{6}{*}{\makecell[c]{\rotatebox{90}{\textbf{carrot}}}} & RD$^*$~\cite{rd} & 87.3 / 96.7 / 93.2 & 97.2 & 99.4 / 27.5 / 33.7 & 15.5 / 47.8 / \pzo8.8 / 20.2 & 92.4 / 53.5 / 24.0 \\ 
& UniAD$^\dagger$~\cite{uniad} & 76.8 / 93.8 / 92.5 & 93.4 & 98.0 / 12.2 / 19.3 & \pzo9.4 / 36.5 / \pzo5.1 / 10.7 & 87.7 / 43.2 / 17.0 \\ 
& SimpleNet$^*$~\cite{simplenet} & 71.4 / 92.6 / 91.2 & 84.4 & 96.4 / 13.8 / 21.2 & 10.1 / 55.0 / \pzo5.5 / 11.9 & 85.1 / 43.8 / 23.5 \\ 
& DiAD~\cite{diad} & 94.4 / 99.3 / 98.0 & 94.6 & 98.6 / 20.0 / 26.9 & \pzo6.0 / 41.7 / \pzo3.1 / \pzo5.8 & 97.2 / 48.5 / 16.9 \\ 
& InvAD-lite & 90.0 / 97.3 / 94.4 & 98.0 & 99.4 / 27.0 / 33.2 & 18.0 / 43.5 / 10.3 / 19.9 & 93.9 / 53.2 / 23.9 \\ 
& InvAD & 89.7 / 97.4 / 94.5 & 96.9 & 99.3 / 26.3 / 30.6 & 18.7 / 39.2 / 10.6 / 18.1 & 93.9 / 52.1 / 22.8 \\ 

            \hline
            \multirow{6}{*}{\makecell[c]{\rotatebox{90}{\textbf{cookie}}}} & RD$^*$~\cite{rd} & 46.0 / 77.4 / 88.0 & 86.6 & 96.6 / 27.5 / 32.9 & 15.5 / 66.8 / \pzo8.7 / 19.7 & 70.5 / 52.3 / 30.3 \\ 
& UniAD$^\dagger$~\cite{uniad} & 77.3 / 93.5 / 88.0 & 88.7 & 97.5 / 40.4 / 45.6 & 18.3 / 50.0 / 10.8 / 29.5 & 86.3 / 61.2 / 26.4 \\ 
& SimpleNet$^*$~\cite{simplenet} & 66.7 / 89.6 / 88.4 & 66.6 & 90.5 / 26.7 / 31.4 & 12.0 / 42.3 / \pzo6.7 / 18.6 & 81.6 / 49.5 / 20.3 \\ 
& DiAD~\cite{diad} & 69.4 / 78.8 / 90.9 & 83.5 & 94.3 / 14.0 / 23.8 & \pzo6.8 / 53.9 / \pzo3.6 / \pzo5.6 & 79.7 / 44.0 / 21.4 \\ 
& InvAD-lite & 64.8 / 87.3 / 88.4 & 87.7 & 97.3 / 37.1 / 40.5 & 18.3 / 60.7 / 10.6 / 25.4 & 80.2 / 58.3 / 29.9 \\ 
& InvAD & 64.5 / 88.1 / 88.4 & 93.4 & 98.5 / 49.6 / 50.4 & 20.7 / 51.9 / 12.5 / 33.7 & 80.3 / 66.2 / 28.4 \\ 

            \hline
            \multirow{6}{*}{\makecell[c]{\rotatebox{90}{\textbf{dowel}}}} & RD$^*$~\cite{rd} & 96.7 / 98.9 / 97.6 & 98.8 & 99.7 / 47.7 / 50.8 & 23.5 / 47.5 / 14.5 / 34.0 & 97.7 / 66.1 / 28.5 \\ 
& UniAD$^\dagger$~\cite{uniad} & 96.7 / 99.3 / 96.2 & 96.1 & 99.1 / 32.1 / 37.7 & 20.0 / 56.2 / 11.6 / 23.2 & 97.4 / 56.3 / 29.3 \\ 
& SimpleNet$^*$~\cite{simplenet} & 83.7 / 95.1 / 91.7 & 83.0 & 95.3 / 17.4 / 25.6 & 11.3 / 62.0 / \pzo6.2 / 14.7 & 90.2 / 46.1 / 26.5 \\ 
& DiAD~\cite{diad} & 98.0 / 99.3 / 97.3 & 89.6 & 97.2 / 31.4 / 40.1 & \pzo3.9 / 29.8 / \pzo2.0 / \pzo4.2 & 98.2 / 56.2 / 11.9 \\ 
& InvAD-lite & 97.2 / 99.4 / 96.2 & 97.6 & 99.6 / 52.6 / 51.8 & 31.9 / 56.7 / 20.0 / 34.9 & 97.6 / 68.0 / 36.2 \\ 
& InvAD & 99.0 / 99.8 / 99.0 & 98.2 & 99.7 / 50.8 / 49.5 & 31.8 / 50.5 / 19.7 / 32.9 & 99.3 / 66.7 / 34.0 \\ 

            \hline
            \multirow{6}{*}{\makecell[c]{\rotatebox{90}{\textbf{foam}}}} & RD$^*$~\cite{rd} & 74.3 / 92.9 / 90.6 & 79.9 & 94.2 / 15.0 / 26.4 & 12.3 / 31.5 / \pzo6.8 / 15.2 & 85.9 / 45.2 / 16.9 \\ 
& UniAD$^\dagger$~\cite{uniad} & 70.5 / 92.4 / 88.9 & 55.8 & 82.2 / \pzo6.8 / 18.9 & \pzo9.3 / 20.5 / \pzo5.0 / 10.4 & 83.9 / 36.0 / 11.6 \\ 
& SimpleNet$^*$~\cite{simplenet} & 77.4 / 94.2 / 89.7 & 66.7 & 87.8 / 15.7 / 26.7 & 12.1 / 22.5 / \pzo6.7 / 15.4 & 87.1 / 43.4 / 13.8 \\ 
& DiAD~\cite{diad} & 100. / 100. / 100. & 69.1 & 89.8 / \pzo9.6 / 23.5 & \pzo1.8 / 54.9 / \pzo0.9 / \pzo2.7 & 100. / 41.0 / 19.2 \\ 
& InvAD-lite & 83.6 / 95.7 / 90.4 & 81.6 & 94.8 / 24.3 / 32.3 & 20.4 / 39.1 / 11.7 / 19.3 & 89.9 / 50.5 / 23.7 \\ 
& InvAD & 77.3 / 93.5 / 90.8 & 78.4 & 93.8 / 18.3 / 30.8 & 16.2 / 35.5 / \pzo9.1 / 18.2 & 87.2 / 47.6 / 20.3 \\ 

            \hline
            \multirow{6}{*}{\makecell[c]{\rotatebox{90}{\textbf{peach}}}} & RD$^*$~\cite{rd} & 64.3 / 84.8 / 90.6 & 93.2 & 98.5 / 15.5 / 22.7 & \pzo9.7 / 55.3 / \pzo5.2 / 12.8 & 79.9 / 45.6 / 23.4 \\ 
& UniAD$^\dagger$~\cite{uniad} & 70.0 / 91.0 / 90.5 & 90.4 & 97.4 / 11.7 / 17.9 & \pzo7.8 / 61.8 / \pzo4.1 / \pzo9.9 & 83.8 / 42.3 / 24.6 \\ 
& SimpleNet$^*$~\cite{simplenet} & 62.0 / 86.9 / 89.7 & 74.8 & 92.9 / \pzo8.1 / 15.0 & \pzo6.6 / 47.6 / \pzo3.5 / \pzo8.1 & 79.5 / 38.7 / 19.2 \\ 
& DiAD~\cite{diad} & 58.0 / 91.3 / 94.3 & 94.2 & 98.4 / 27.6 / 31.3 & \pzo4.6 / 60.7 / \pzo2.4 / \pzo4.5 & 81.2 / 52.5 / 22.6 \\ 
& InvAD-lite & 86.8 / 96.1 / 94.1 & 96.5 & 99.2 / 40.6 / 43.3 & 18.9 / 62.7 / 11.1 / 27.6 & 92.3 / 61.0 / 30.9 \\ 
& InvAD & 88.8 / 96.9 / 92.6 & 96.2 & 99.2 / 39.1 / 43.6 & 19.6 / 63.4 / 11.6 / 27.9 & 92.8 / 60.6 / 31.5 \\ 

            \hline
            \multirow{6}{*}{\makecell[c]{\rotatebox{90}{\textbf{potato}}}} & RD$^*$~\cite{rd} & 62.5 / 88.5 / 90.5 & 95.9 & 99.1 / 14.9 / 22.5 & 10.4 / 61.6 / \pzo5.7 / 12.7 & 80.5 / 45.5 / 25.9 \\ 
& UniAD$^\dagger$~\cite{uniad} & 51.6 / 81.8 / 89.3 & 91.1 & 97.6 / \pzo5.1 / \pzo8.9 & \pzo3.3 / 32.7 / \pzo1.7 / \pzo4.7 & 74.2 / 37.2 / 12.6 \\ 
& SimpleNet$^*$~\cite{simplenet} & 56.7 / 82.2 / 89.8 & 72.8 & 91.0 / \pzo4.3 / 10.9 & \pzo4.8 / 56.8 / \pzo2.5 / \pzo5.8 & 76.2 / 35.4 / 21.4 \\ 
& DiAD~\cite{diad} & 76.3 / 94.3 / 95.0 & 93.9 & 98.0 / \pzo8.6 / 17.8 & \pzo6.2 / 66.1 / \pzo3.3 / \pzo7.8 & 88.5 / 41.5 / 25.2 \\ 
& InvAD-lite & 63.4 / 86.0 / 92.0 & 95.5 & 99.1 / 15.6 / 20.9 & 11.5 / 61.5 / \pzo6.2 / 11.7 & 80.5 / 45.2 / 26.4 \\ 
& InvAD & 67.7 / 90.5 / 90.2 & 96.0 & 99.1 / 18.9 / 25.2 & 13.8 / 63.8 / \pzo7.6 / 14.4 & 82.8 / 47.7 / 28.4 \\ 

            \hline
            \multirow{6}{*}{\makecell[c]{\rotatebox{90}{\textbf{rope}}}} & RD$^*$~\cite{rd} & 96.3 / 98.5 / 93.2 & 97.9 & 99.6 / 50.3 / 55.9 & 22.9 / 37.5 / 14.4 / 38.8 & 96.0 / 68.6 / 24.9 \\ 
& UniAD$^\dagger$~\cite{uniad} & 97.4 / 99.0 / 95.5 & 94.3 & 99.0 / 34.5 / 40.7 & 23.7 / 50.7 / 14.1 / 25.5 & 97.3 / 58.1 / 29.5 \\ 
& SimpleNet$^*$~\cite{simplenet} & 95.6 / 98.4 / 94.7 & 92.8 & 99.3 / 51.1 / 52.9 & 25.8 / 67.4 / 16.2 / 36.0 & 96.2 / 67.8 / 36.5 \\ 
& DiAD~\cite{diad} & 89.2 / 95.4 / 91.9 & 96.5 & 99.3 / 61.0 / 59.9 & \pzo7.5 / 56.3 / \pzo3.9 / \pzo7.5 & 92.2 / 73.4 / 22.6 \\ 
& InvAD-lite & 96.1 / 98.5 / 95.5 & 95.7 & 99.4 / 50.0 / 51.8 & 29.4 / 52.4 / 18.4 / 35.0 & 96.7 / 67.1 / 33.4 \\ 
& InvAD & 95.6 / 98.3 / 93.3 & 96.7 & 99.6 / 56.1 / 57.3 & 29.4 / 42.0 / 18.9 / 40.2 & 95.7 / 71.0 / 30.1 \\ 

            \hline
            \multirow{6}{*}{\makecell[c]{\rotatebox{90}{\textbf{tire}}}} & RD$^*$~\cite{rd} & 79.2 / 93.0 / 88.4 & 96.4 & 99.2 / 23.2 / 31.1 & 13.2 / 53.4 / \pzo7.4 / 18.4 & 86.9 / 51.2 / 24.7 \\ 
& UniAD$^\dagger$~\cite{uniad} & 75.7 / 90.7 / 89.7 & 90.6 & 98.0 / 11.9 / 20.3 & \pzo5.7 / 28.9 / \pzo3.0 / 11.3 & 85.4 / 43.4 / 12.5 \\ 
& SimpleNet$^*$~\cite{simplenet} & 65.3 / 86.8 / 87.9 & 77.9 & 93.8 / \pzo8.1 / 15.3 & \pzo5.8 / 50.4 / \pzo3.1 / \pzo8.3 & 80.0 / 39.1 / 19.8 \\ 
& DiAD~\cite{diad} & 92.7 / 98.9 / 95.8 & 68.8 & 91.8 / \pzo5.9 / 13.7 & \pzo3.9 / 52.5 / \pzo2.0 / \pzo5.1 & 95.8 / 37.1 / 19.5 \\ 
& InvAD-lite & 85.7 / 95.6 / 90.2 & 96.4 & 99.3 / 40.9 / 46.7 & 22.9 / 54.3 / 14.0 / 30.5 & 90.5 / 62.3 / 30.4 \\ 
& InvAD & 87.8 / 96.2 / 92.4 & 97.6 & 99.6 / 32.3 / 40.9 & 20.4 / 46.3 / 11.9 / 25.7 & 92.1 / 57.6 / 26.2 \\ 
            \toprule[0.12em] 
        \end{tabular}
    }
\end{table*}

\section{Quantitative Results for Each Category on Uni-Medical} \label{sec:app_unimedical}
\Tab\ref{tab:supp_uni_medical} shows detailed quantitative single-class results on Uni-Medical dataset.

\begin{table*}[htp]
    \centering
    \caption{\textbf{Quantitative results for each class on Uni-Medical dataset.} These are respectively the image-level mAU-ROC/mAP/m$F_1$-max, region-level \redzjn{mAU-PRO}, pixel-level mAU-ROC/mAP/m$F_1$-max, the proposed pixel-level m$F_1$$^{.2}_{.8}$/mAcc$^{.2}_{.8}$/mIoU$^{.2}_{.8}$/\redzjn{mIoU-max}, and averaged \redzjn{mAD$_I$}/\redzjn{mAD$_P$}/\redzjn{mAD$^{.2}_{.8}$}.}
    \label{tab:supp_uni_medical}
    \renewcommand{\arraystretch}{1.0}
    \setlength\tabcolsep{6.0pt}
    \resizebox{1.0\linewidth}{!}{
        \begin{tabular}{p{0.3cm}<{\centering} p{2.0cm}<{\centering} p{3.9cm}<{\centering} p{1.6cm}<{\centering} p{3.9cm}<{\centering} p{4.9cm}<{\centering} p{3.3cm}<{\centering}}
            \toprule[0.17em]
            & Method & mAU-ROC/mAP/m$F_1$-max & \redzjn{mAU-PRO} & mAU-ROC/mAP/m$F_1$-max & m$F_1$$^{.2}_{.8}$/mAcc$^{.2}_{.8}$/mIoU$^{.2}_{.8}$/\redzjn{mIoU-max} & \redzjn{mAD$_I$}/\redzjn{mAD$_P$}/\redzjn{mAD$^{.2}_{.8}$} \\
            \hline
            \multirow{6}{*}{\makecell[c]{\rotatebox{90}{\textbf{brain}}}} & RD$^*$~\cite{rd} & 82.4 / 94.4 / 91.5 & 82.6 & 96.5 / 45.9 / 49.2 & 18.6 / 44.2 / 11.2 / 32.6 & 89.4 / 63.9 / 24.7 \\ 
& UniAD$^\dagger$~\cite{uniad} & 89.9 / 97.5 / 92.6 & 82.4 & 97.4 / 55.7 / 55.7 & 18.2 / 45.5 / 11.2 / 38.6 & 93.3 / 69.6 / 25.0 \\ 
& SimpleNet$^*$~\cite{simplenet} & 82.3 / 95.6 / 90.9 & 73.0 & 94.8 / 42.1 / 42.4 & 13.7 / 33.9 / \pzo8.1 / 26.9 & 89.6 / 59.8 / 18.6 \\ 
& DiAD~\cite{diad} & 93.2 / 98.3 / 95.0 & 80.3 & 95.4 / 42.9 / 36.7 & 22.5 / 55.1 / 12.7 / 27.3 & 95.5 / 58.3 / 30.1 \\ 
& InvAD-lite & 86.8 / 96.3 / 92.0 & 82.6 & 96.6 / 47.4 / 50.1 & 24.2 / 49.6 / 14.8 / 33.5 & 91.7 / 64.7 / 29.5 \\ 
& InvAD & 90.5 / 97.6 / 93.1 & 87.2 & 98.1 / 64.0 / 61.7 & 28.3 / 54.4 / 18.3 / 44.6 & 93.7 / 74.6 / 33.7 \\ 
            
            \hline
            \multirow{6}{*}{\makecell[c]{\rotatebox{90}{\textbf{liver}}}} & RD$^*$~\cite{rd} & 55.1 / 46.3 / 64.1 & 89.9 & 96.6 / \pzo5.7 / 10.3 & \pzo6.4 / 43.6 / \pzo3.4 / \pzo5.4 & 55.2 / 37.5 / 17.8 \\ 
& UniAD$^\dagger$~\cite{uniad} & 61.0 / 48.8 / 63.2 & 92.7 & 97.1 / \pzo7.8 / 13.7 & \pzo5.6 / 40.4 / \pzo2.9 / \pzo7.4 & 57.7 / 39.5 / 16.3 \\ 
& SimpleNet$^*$~\cite{simplenet} & 55.8 / 47.6 / 60.9 & 86.3 & 97.4 / 13.2 / 20.1 & \pzo8.5 / 55.7 / \pzo4.6 / 11.2 & 54.8 / 43.6 / 22.9 \\ 
& DiAD~\cite{diad} & 70.6 / 64.7 / 65.6 & 91.4 & 97.1 / 13.7 / \pzo7.3 & \pzo7.1 / 26.7 / \pzo3.3 / \pzo7.3 & 67.0 / 39.4 / 12.3 \\ 
& InvAD-lite & 60.0 / 48.6 / 63.7 & 89.2 & 96.4 / \pzo5.7 / 10.1 & \pzo6.6 / 39.2 / \pzo3.5 / \pzo5.3 & 57.4 / 37.4 / 16.4 \\ 
& InvAD & 64.8 / 52.5 / 66.6 & 93.3 & 97.2 / \pzo8.7 / 15.8 & \pzo9.1 / 45.9 / \pzo4.8 / \pzo8.6 & 61.3 / 40.6 / 19.9 \\ 

            \hline
            \multirow{6}{*}{\makecell[c]{\rotatebox{90}{\textbf{retinal}}}} & RD$^*$~\cite{rd} & 89.2 / 86.7 / 78.5 & 86.5 & 96.4 / 64.7 / 60.9 & 24.0 / 41.9 / 15.4 / 43.8 & 84.8 / 74.0 / 27.1 \\ 
& UniAD$^\dagger$~\cite{uniad} & 84.6 / 79.4 / 73.9 & 79.9 & 94.8 / 49.3 / 51.3 & 16.2 / 27.4 / 10.0 / 34.5 & 79.3 / 65.1 / 17.9 \\ 
& SimpleNet$^*$~\cite{simplenet} & 88.8 / 87.6 / 78.6 & 82.1 & 95.5 / 59.5 / 56.3 & 22.6 / 49.8 / 14.2 / 39.2 & 85.0 / 70.4 / 28.9 \\ 
& DiAD~\cite{diad} & 91.6 / 90.7 / 83.0 & 84.4 & 95.3 / 57.5 / 62.8 & 24.5 / 30.9 / 15.2 / 40.4 & 88.4 / 71.9 / 23.5 \\ 
& InvAD-lite & 91.7 / 89.9 / 81.7 & 84.7 & 96.2 / 67.2 / 61.1 & 23.9 / 32.8 / 15.4 / 44.0 & 87.8 / 74.8 / 24.0 \\ 
& InvAD & 91.4 / 88.7 / 82.0 & 88.4 & 97.0 / 69.8 / 63.7 & 27.9 / 35.2 / 18.4 / 46.8 & 87.4 / 76.8 / 27.2 \\ 
            \toprule[0.12em] 
        \end{tabular}
    }
\end{table*}

\section{Quantitative Results for Each Category on Real-IAD} \label{sec:app_realiad}
\Tab\ref{tab:supp_realiad_1}, \Tab\ref{tab:supp_realiad_2}, and \Tab\ref{tab:supp_realiad_3} show detailed quantitative single-class results on Real-IAD dataset.

\begin{table*}[htp]
    \centering
    \caption{\textbf{Quantitative results for each class on Real-IAD dataset.} These are respectively the image-level mAU-ROC/mAP/m$F_1$-max, region-level \redzjn{mAU-PRO}, pixel-level mAU-ROC/mAP/m$F_1$-max, the proposed pixel-level m$F_1$$^{.2}_{.8}$/mAcc$^{.2}_{.8}$/mIoU$^{.2}_{.8}$/\redzjn{mIoU-max}, and averaged \redzjn{mAD$_I$}/\redzjn{mAD$_P$}/\redzjn{mAD$^{.2}_{.8}$}. Part~\redzjn{I}.}
    \label{tab:supp_realiad_1}
    \renewcommand{\arraystretch}{1.0}
    \setlength\tabcolsep{6.0pt}
    \resizebox{1.0\linewidth}{!}{
        \begin{tabular}{p{0.3cm}<{\centering} p{2.0cm}<{\centering} p{3.9cm}<{\centering} p{1.6cm}<{\centering} p{3.9cm}<{\centering} p{4.9cm}<{\centering} p{3.3cm}<{\centering}}
            \toprule[0.17em]
            & Method & mAU-ROC/mAP/m$F_1$-max & \redzjn{mAU-PRO} & mAU-ROC/mAP/m$F_1$-max & m$F_1$$^{.2}_{.8}$/mAcc$^{.2}_{.8}$/mIoU$^{.2}_{.8}$/\redzjn{mIoU-max} & \redzjn{mAD$_I$}/\redzjn{mAD$_P$}/\redzjn{mAD$^{.2}_{.8}$} \\
            \hline
            \multirow{6}{*}{\makecell[c]{\rotatebox{90}{\textbf{audiojack}}}} & RD$^*$~\cite{rd} & 76.2 / 63.2 / 60.8 & 79.6 & 96.6 / 12.8 / 22.1 & \pzo7.0 / 62.9 / \pzo3.8 / 12.2 & 66.7 / 43.8 / 24.6 \\ 
& UniAD$^\dagger$~\cite{uniad} & 82.0 / 76.7 / 65.5 & 84.0 & 97.6 / 20.5 / 31.2 & 11.7 / 44.6 / \pzo6.6 / 18.5 & 74.7 / 49.8 / 21.0 \\ 
& SimpleNet$^*$~\cite{simplenet} & 58.4 / 44.2 / 50.9 & 38.0 & 74.4 / \pzo0.9 / \pzo4.8 & \pzo1.3 / 45.0 / \pzo0.7 / \pzo2.5 & 51.2 / 26.7 / 15.7 \\ 
& DiAD~\cite{diad} & 76.5 / 54.3 / 65.7 & 63.3 & 91.6 / \pzo1.0 / \pzo3.9 & \pzo1.7 / 41.0 / \pzo0.9 / \pzo2.0 & 65.5 / 32.2 / 14.5 \\ 
& InvAD-lite & 84.2 / 78.8 / 67.9 & 86.7 & 98.1 / 31.6 / 41.0 & 17.2 / 52.0 / 10.1 / 25.8 & 77.0 / 56.9 / 26.4 \\ 
& InvAD & 84.4 / 79.5 / 67.7 & 88.8 & 98.4 / 32.1 / 42.4 & 19.8 / 50.7 / 11.8 / 26.9 & 77.2 / 57.6 / 27.4 \\ 
            
            \hline
            \multirow{6}{*}{\makecell[c]{\rotatebox{90}{\textbf{\makecell[c]{bottle\\cap}}}}} & RD$^*$~\cite{rd} & 89.5 / 86.3 / 81.0 & 95.7 & 99.5 / 18.9 / 29.9 & 11.8 / 38.5 / \pzo6.6 / 17.6 & 85.6 / 49.4 / 19.0 \\ 
& UniAD$^\dagger$~\cite{uniad} & 92.0 / 91.3 / 80.7 & 95.5 & 99.5 / 21.0 / 29.9 & 13.2 / 35.9 / \pzo7.4 / 17.6 & 88.0 / 50.1 / 18.8 \\ 
& SimpleNet$^*$~\cite{simplenet} & 54.1 / 47.6 / 60.3 & 45.1 & 85.3 / \pzo2.3 / \pzo5.7 & \pzo1.6 / 47.7 / \pzo0.8 / \pzo2.9 & 54.0 / 31.1 / 16.7 \\ 
& DiAD~\cite{diad} & 91.6 / 94.0 / 87.9 & 73.0 & 94.6 / \pzo4.9 / 11.4 & \pzo4.6 / 41.4 / \pzo2.4 / \pzo6.0 & 91.1 / 37.0 / 16.1 \\ 
& InvAD-lite & 92.1 / 90.2 / 81.6 & 97.1 & 99.6 / 20.2 / 26.6 & 13.1 / 23.9 / \pzo7.3 / 15.3 & 88.0 / 48.8 / 14.8 \\ 
& InvAD & 94.5 / 93.6 / 84.2 & 96.2 & 99.6 / 22.3 / 32.3 & 16.6 / 30.5 / \pzo9.4 / 19.2 & 90.8 / 51.4 / 18.8 \\ 

            \hline
            \multirow{6}{*}{\makecell[c]{\rotatebox{90}{\textbf{\makecell[c]{button\\battery}}}}} & RD$^*$~\cite{rd} & 73.3 / 78.9 / 76.1 & 86.5 & 97.6 / 33.8 / 37.8 & 15.4 / 47.9 / \pzo8.9 / 23.6 & 76.1 / 56.4 / 24.1 \\ 
& UniAD$^\dagger$~\cite{uniad} & 73.4 / 79.8 / 75.1 & 78.0 & 96.4 / 28.5 / 35.4 & 10.0 / 39.2 / \pzo5.6 / 21.5 & 76.1 / 53.4 / 18.3 \\ 
& SimpleNet$^*$~\cite{simplenet} & 52.5 / 60.5 / 72.4 & 40.5 & 75.9 / \pzo3.2 / \pzo6.6 & \pzo2.7 / 40.7 / \pzo1.4 / \pzo3.4 & 61.8 / 28.6 / 14.9 \\ 
& DiAD~\cite{diad} & 80.5 / 71.3 / 70.6 & 66.9 & 84.1 / \pzo1.4 / \pzo5.3 & \pzo2.0 / 21.2 / \pzo1.0 / \pzo2.7 & 74.2 / 30.2 / \pzo8.0 \\ 
& InvAD-lite & 82.3 / 87.2 / 78.8 & 88.4 & 98.5 / 50.7 / 52.5 & 23.7 / 38.4 / 14.7 / 35.6 & 82.8 / 67.2 / 25.6 \\ 
& InvAD & 89.3 / 91.7 / 83.5 & 92.0 & 99.1 / 55.4 / 54.9 & 26.9 / 37.1 / 17.1 / 37.8 & 88.2 / 69.8 / 27.0 \\ 
            
            \hline
            \multirow{6}{*}{\makecell[c]{\rotatebox{90}{\textbf{\makecell[c]{end\\cap}}}}} & RD$^*$~\cite{rd} & 79.8 / 84.0 / 77.8 & 89.2 & 96.7 / 12.5 / 22.5 & \pzo7.2 / 38.0 / \pzo3.9 / 12.0 & 80.5 / 43.9 / 16.4 \\ 
& UniAD$^\dagger$~\cite{uniad} & 80.4 / 85.9 / 77.6 & 85.0 & 95.7 / \pzo9.2 / 17.4 & \pzo6.1 / 29.8 / \pzo3.3 / \pzo9.5 & 81.3 / 40.8 / 13.1 \\ 
& SimpleNet$^*$~\cite{simplenet} & 51.6 / 60.8 / 72.9 & 25.7 & 63.1 / \pzo0.5 / \pzo2.8 & \pzo0.9 / 37.5 / \pzo0.4 / \pzo1.4 & 61.8 / 22.1 / 12.9 \\ 
& DiAD~\cite{diad} & 85.1 / 83.4 / 84.8 & 38.2 & 81.3 / \pzo2.0 / \pzo6.9 & \pzo2.2 / 37.2 / \pzo1.1 / \pzo3.6 & 84.4 / 30.1 / 13.5 \\ 
& InvAD-lite & 79.9 / 84.4 / 78.2 & 90.3 & 97.4 / 13.4 / 20.8 & \pzo9.4 / 20.8 / \pzo5.1 / 11.6 & 80.8 / 43.9 / 11.8 \\ 
& InvAD & 84.8 / 88.3 / 80.8 & 93.4 & 98.0 / 14.4 / 22.6 & 10.3 / 26.2 / \pzo5.6 / 12.7 & 84.6 / 45.0 / 14.0 \\ 
            
            \hline
            \multirow{6}{*}{\makecell[c]{\rotatebox{90}{\textbf{eraser}}}} & RD$^*$~\cite{rd} & 90.0 / 88.7 / 79.7 & 96.0 & 99.5 / 30.8 / 36.7 & 13.6 / 47.9 / \pzo7.8 / 22.3 & 86.1 / 55.7 / 23.1 \\ 
& UniAD$^\dagger$~\cite{uniad} & 89.6 / 88.0 / 79.6 & 94.1 & 99.3 / 24.7 / 31.3 & 11.4 / 35.5 / \pzo6.5 / 18.5 & 85.7 / 51.8 / 17.8 \\ 
& SimpleNet$^*$~\cite{simplenet} & 46.4 / 39.1 / 55.8 & 42.8 & 80.6 / \pzo2.7 / \pzo7.1 & \pzo2.4 / 38.7 / \pzo1.2 / \pzo3.7 & 47.1 / 30.1 / 14.1 \\ 
& DiAD~\cite{diad} & 80.0 / 80.0 / 77.3 & 67.5 & 91.1 / \pzo7.7 / 15.4 & \pzo7.9 / 32.1 / \pzo4.2 / \pzo8.4 & 79.1 / 38.1 / 14.7 \\ 
& InvAD-lite & 88.9 / 87.4 / 77.8 & 94.7 & 99.3 / 28.9 / 33.8 & 16.3 / 38.8 / \pzo9.3 / 20.3 & 84.7 / 54.0 / 21.5 \\ 
& InvAD & 91.7 / 90.0 / 81.0 & 94.9 & 99.4 / 28.2 / 34.2 & 18.0 / 42.8 / 10.3 / 20.6 & 87.6 / 53.9 / 23.7 \\ 
            
            \hline
            \multirow{6}{*}{\makecell[c]{\rotatebox{90}{\textbf{\makecell[c]{fire\\hood}}}}} & RD$^*$~\cite{rd} & 78.3 / 70.1 / 64.5 & 87.9 & 98.9 / 27.7 / 35.2 & 13.0 / 56.7 / \pzo7.4 / 20.9 & 71.0 / 53.9 / 25.7 \\ 
& UniAD$^\dagger$~\cite{uniad} & 80.6 / 74.5 / 67.3 & 84.7 & 98.6 / 24.5 / 32.5 & 13.0 / 52.9 / \pzo7.4 / 19.4 & 74.1 / 51.9 / 24.4 \\ 
& SimpleNet$^*$~\cite{simplenet} & 58.1 / 41.9 / 54.4 & 25.3 & 70.5 / \pzo0.3 / \pzo2.2 & \pzo0.6 / 40.7 / \pzo0.3 / \pzo1.1 & 51.5 / 24.3 / 13.9 \\ 
& DiAD~\cite{diad} & 83.3 / 81.7 / 80.5 & 66.7 & 91.8 / \pzo3.2 / \pzo9.2 & \pzo3.3 / 33.8 / \pzo1.7 / \pzo4.8 & 81.8 / 34.7 / 12.9 \\ 
& InvAD-lite & 79.9 / 73.9 / 65.6 & 87.4 & 98.8 / 24.1 / 31.2 & 12.6 / 40.2 / \pzo7.1 / 18.5 & 73.1 / 51.4 / 20.0 \\ 
& InvAD & 81.4 / 74.1 / 67.3 & 88.6 & 99.0 / 25.6 / 33.2 & 12.9 / 40.4 / \pzo7.3 / 19.9 & 74.3 / 52.6 / 20.2 \\ 
            
            \hline
            \multirow{6}{*}{\makecell[c]{\rotatebox{90}{\textbf{mint}}}} & RD$^*$~\cite{rd} & 65.8 / 63.1 / 64.8 & 72.3 & 95.0 / 11.7 / 23.0 & \pzo5.6 / 42.9 / \pzo3.1 / 12.1 & 64.6 / 43.2 / 17.2 \\ 
& UniAD$^\dagger$~\cite{uniad} & 66.8 / 67.6 / 64.2 & 62.4 & 94.5 / \pzo8.5 / 20.9 & \pzo4.6 / 26.7 / \pzo2.5 / 11.7 & 66.2 / 41.3 / 11.3 \\ 
& SimpleNet$^*$~\cite{simplenet} & 52.4 / 50.3 / 63.7 & 43.3 & 79.9 / \pzo0.9 / \pzo3.6 & \pzo1.0 / 43.2 / \pzo0.5 / \pzo1.8 & 55.5 / 28.1 / 14.9 \\ 
& DiAD~\cite{diad} & 76.7 / 76.7 / 76.0 & 64.2 & 91.1 / \pzo5.7 / 11.6 & \pzo5.3 / 44.5 / \pzo2.8 / \pzo6.2 & 76.5 / 36.1 / 17.5 \\ 
& InvAD-lite & 71.7 / 73.0 / 65.8 & 79.9 & 97.3 / 18.2 / 28.5 & 12.2 / 33.6 / \pzo6.9 / 16.6 & 70.2 / 48.0 / 17.6 \\ 
& InvAD & 76.5 / 76.6 / 68.9 & 82.6 & 98.2 / 16.9 / 28.8 & 11.2 / 24.5 / \pzo6.3 / 16.8 & 74.0 / 48.0 / 14.0 \\ 
            
            \hline
            \multirow{6}{*}{\makecell[c]{\rotatebox{90}{\textbf{mounts}}}} & RD$^*$~\cite{rd} & 88.6 / 79.9 / 74.8 & 94.9 & 99.3 / 30.6 / 37.1 & 15.5 / 49.2 / \pzo8.9 / 22.8 & 81.1 / 55.7 / 24.5 \\ 
& UniAD$^\dagger$~\cite{uniad} & 86.8 / 73.6 / 76.3 & 95.1 & 99.4 / 27.4 / 32.9 & 13.7 / 36.0 / \pzo7.9 / 19.7 & 78.9 / 53.2 / 19.2 \\ 
& SimpleNet$^*$~\cite{simplenet} & 58.7 / 48.1 / 52.4 & 46.1 & 80.5 / \pzo2.2 / \pzo6.8 & \pzo2.1 / 34.3 / \pzo1.1 / \pzo3.5 & 53.1 / 29.8 / 12.5 \\ 
& DiAD~\cite{diad} & 75.3 / 74.5 / 82.5 & 48.8 & 84.3 / \pzo0.4 / \pzo1.1 & \pzo0.6 / 29.1 / \pzo0.3 / \pzo0.5 & 77.4 / 28.6 / 10.0 \\ 
& InvAD-lite & 87.8 / 77.1 / 76.1 & 94.5 & 99.4 / 27.4 / 33.0 & 17.4 / 31.4 / \pzo9.9 / 19.8 & 80.3 / 53.3 / 19.6 \\ 
& InvAD & 87.9 / 76.6 / 77.4 & 93.7 & 99.4 / 28.1 / 32.9 & 18.3 / 32.4 / 10.4 / 19.7 & 80.6 / 53.5 / 20.4 \\ 
            
            \hline
            \multirow{6}{*}{\makecell[c]{\rotatebox{90}{\textbf{pcb}}}} & RD$^*$~\cite{rd} & 79.5 / 85.8 / 79.7 & 88.3 & 97.5 / 15.8 / 24.3 & \pzo7.6 / 53.9 / \pzo4.1 / 14.7 & 81.7 / 45.9 / 21.9 \\ 
& UniAD$^\dagger$~\cite{uniad} & 80.6 / 87.8 / 79.1 & 81.6 & 97.2 / 20.6 / 31.0 & \pzo7.6 / 39.8 / \pzo4.2 / 18.4 & 82.5 / 49.6 / 17.2 \\ 
& SimpleNet$^*$~\cite{simplenet} & 54.5 / 66.0 / 75.5 & 41.3 & 78.0 / \pzo1.4 / \pzo4.3 & \pzo1.4 / 49.5 / \pzo0.7 / \pzo2.2 & 65.3 / 27.9 / 17.2 \\ 
& DiAD~\cite{diad} & 86.0 / 85.1 / 85.4 & 66.5 & 92.0 / \pzo3.7 / \pzo7.4 & \pzo3.4 / 18.6 / \pzo1.7 / \pzo3.8 & 85.5 / 34.4 / \pzo7.9 \\ 
& InvAD-lite & 90.7 / 94.5 / 85.5 & 94.4 & 99.4 / 49.1 / 52.1 & 22.0 / 41.0 / 13.8 / 35.2 & 90.2 / 66.9 / 25.6 \\ 
& InvAD & 91.9 / 95.2 / 86.9 & 94.6 & 99.4 / 48.2 / 51.6 & 24.4 / 47.2 / 15.2 / 34.8 & 91.3 / 66.4 / 28.9 \\ 
            
            \hline
            \multirow{6}{*}{\makecell[c]{\rotatebox{90}{\textbf{\makecell[c]{phone\\battery}}}}} & RD$^*$~\cite{rd} & 87.5 / 83.3 / 77.1 & 94.5 & 77.3 / 22.6 / 31.7 & 13.7 / 37.8 / \pzo7.8 / 18.9 & 82.6 / 43.9 / 19.8 \\ 
& UniAD$^\dagger$~\cite{uniad} & 83.3 / 77.2 / 73.5 & 86.7 & 88.8 / \pzo8.8 / 16.4 & \pzo6.4 / 25.4 / \pzo3.4 / \pzo8.9 & 78.0 / 38.0 / 11.7 \\ 
& SimpleNet$^*$~\cite{simplenet} & 51.6 / 43.8 / 58.0 & 11.8 & 43.4 / \pzo0.1 / \pzo0.9 & \pzo0.3 / 34.6 / \pzo0.2 / \pzo0.4 & 51.1 / 14.8 / 11.7 \\ 
& DiAD~\cite{diad} & 82.3 / 77.7 / 75.9 & 85.4 & 96.8 / \pzo5.3 / 11.4 & \pzo4.6 / 29.5 / \pzo2.4 / \pzo6.0 & 78.6 / 37.8 / 12.2 \\ 
& InvAD-lite & 91.2 / 89.6 / 81.8 & 95.5 & 80.2 / 25.3 / 33.6 & 17.3 / 30.2 / \pzo9.9 / 20.2 & 87.5 / 46.4 / 19.1 \\ 
& InvAD & 91.0 / 87.0 / 80.4 & 93.9 & 86.2 / 21.1 / 29.5 & 15.7 / 29.3 / \pzo8.8 / 17.3 & 86.1 / 45.6 / 17.9 \\ 
            \toprule[0.12em] 
        \end{tabular}
    }
\end{table*}

\begin{table*}[htp]
    \centering
    \caption{\textbf{Quantitative results for each class on Real-IAD dataset.} These are respectively the image-level mAU-ROC/mAP/m$F_1$-max, region-level \redzjn{mAU-PRO}, pixel-level mAU-ROC/mAP/m$F_1$-max, the proposed pixel-level m$F_1$$^{.2}_{.8}$/mAcc$^{.2}_{.8}$/mIoU$^{.2}_{.8}$/\redzjn{mIoU-max}, and averaged \redzjn{mAD$_I$}/\redzjn{mAD$_P$}/\redzjn{mAD$^{.2}_{.8}$}. Part~\redzjn{II}.}
    \label{tab:supp_realiad_2}
    \renewcommand{\arraystretch}{1.0}
    \setlength\tabcolsep{6.0pt}
    \resizebox{1.0\linewidth}{!}{
        \begin{tabular}{p{0.3cm}<{\centering} p{2.0cm}<{\centering} p{3.9cm}<{\centering} p{1.6cm}<{\centering} p{3.9cm}<{\centering} p{4.9cm}<{\centering} p{3.3cm}<{\centering}}
            \toprule[0.17em]
            & Method & mAU-ROC/mAP/m$F_1$-max & \redzjn{mAU-PRO} & mAU-ROC/mAP/m$F_1$-max & m$F_1$$^{.2}_{.8}$/mAcc$^{.2}_{.8}$/mIoU$^{.2}_{.8}$/\redzjn{mIoU-max} & \redzjn{mAD$_I$}/\redzjn{mAD$_P$}/\redzjn{mAD$^{.2}_{.8}$} \\
            \hline
            \multirow{6}{*}{\makecell[c]{\rotatebox{90}{\textbf{\makecell[c]{plastic\\nut}}}}} & RD$^*$~\cite{rd} & 80.3 / 68.0 / 64.4 & 91.0 & 98.8 / 21.1 / 29.6 & 11.1 / 37.4 / \pzo6.2 / 17.3 & 70.9 / 49.8 / 18.2 \\ 
& UniAD$^\dagger$~\cite{uniad} & 79.3 / 69.7 / 62.5 & 89.3 & 98.4 / 22.1 / 28.1 & 10.9 / 30.8 / \pzo6.1 / 16.4 & 70.5 / 49.5 / 15.9 \\ 
& SimpleNet$^*$~\cite{simplenet} & 59.2 / 40.3 / 51.8 & 41.5 & 77.4 / \pzo0.6 / \pzo3.6 & \pzo0.8 / 58.1 / \pzo0.4 / \pzo1.8 & 50.4 / 27.2 / 19.8 \\ 
& DiAD~\cite{diad} & 71.9 / 58.2 / 65.6 & 38.6 & 81.1 / \pzo0.4 / \pzo3.4 & \pzo1.3 / 27.9 / \pzo0.7 / \pzo1.7 & 65.2 / 28.3 / 10.0 \\ 
& InvAD-lite & 88.1 / 80.6 / 72.5 & 96.4 & 99.5 / 30.2 / 34.3 & 18.9 / 35.5 / 10.9 / 20.7 & 80.4 / 54.7 / 21.8 \\ 
& InvAD & 89.1 / 81.5 / 73.5 & 95.8 & 99.5 / 27.2 / 31.9 & 19.3 / 35.1 / 11.0 / 19.0 & 81.4 / 52.9 / 21.8 \\ 
            
            \hline
            \multirow{6}{*}{\makecell[c]{\rotatebox{90}{\textbf{\makecell[c]{plastic\\plug}}}}} & RD$^*$~\cite{rd} & 81.9 / 74.3 / 68.8 & 94.9 & 99.1 / 20.5 / 28.4 & \pzo8.5 / 40.5 / \pzo4.7 / 16.0 & 75.0 / 49.3 / 17.9 \\ 
& UniAD$^\dagger$~\cite{uniad} & 81.1 / 76.7 / 67.1 & 88.7 & 98.6 / 17.7 / 26.7 & \pzo9.2 / 43.2 / \pzo5.1 / 15.4 & 75.0 / 47.7 / 19.2 \\ 
& SimpleNet$^*$~\cite{simplenet} & 48.2 / 38.4 / 54.6 & 38.8 & 78.6 / \pzo0.7 / \pzo1.9 & \pzo0.6 / 46.7 / \pzo0.3 / \pzo0.9 & 47.1 / 27.1 / 15.9 \\ 
& DiAD~\cite{diad} & 88.7 / 89.2 / 90.9 & 66.1 & 92.9 / \pzo8.7 / 15.0 & \pzo7.4 / 36.0 / \pzo3.9 / \pzo8.1 & 89.6 / 38.9 / 15.8 \\ 
& InvAD-lite & 88.2 / 84.6 / 74.7 & 93.4 & 99.2 / 26.7 / 34.1 & 14.6 / 38.0 / \pzo8.3 / 20.5 & 82.5 / 53.3 / 20.3 \\ 
& InvAD & 89.0 / 82.5 / 76.4 & 93.0 & 99.0 / 22.1 / 29.9 & 12.6 / 38.3 / \pzo7.0 / 17.6 & 82.6 / 50.3 / 19.3 \\ 
            
            \hline
            \multirow{6}{*}{\makecell[c]{\rotatebox{90}{\textbf{\makecell[c]{porcelain\\doll}}}}} & RD$^*$~\cite{rd} & 86.3 / 76.3 / 71.5 & 95.7 & 99.2 / 24.8 / 34.6 & 12.7 / 40.0 / \pzo7.2 / 20.9 & 78.0 / 52.9 / 20.0 \\ 
& UniAD$^\dagger$~\cite{uniad} & 84.7 / 75.2 / 69.5 & 92.8 & 98.7 / 13.8 / 23.5 & \pzo7.0 / 28.2 / \pzo3.8 / 13.3 & 76.5 / 45.3 / 13.0 \\ 
& SimpleNet$^*$~\cite{simplenet} & 66.3 / 54.5 / 52.1 & 47.0 & 81.8 / \pzo2.0 / \pzo6.4 & \pzo1.8 / 53.7 / \pzo0.9 / \pzo3.3 & 57.6 / 30.1 / 18.8 \\ 
& DiAD~\cite{diad} & 72.6 / 66.8 / 65.2 & 70.4 & 93.1 / \pzo1.4 / \pzo4.8 & \pzo2.2 / 19.6 / \pzo1.1 / \pzo2.5 & 68.2 / 33.1 / \pzo7.6 \\ 
& InvAD-lite & 88.7 / 82.8 / 74.6 & 95.6 & 99.3 / 32.0 / 36.7 & 18.9 / 31.7 / 11.0 / 22.5 & 82.0 / 56.0 / 20.5 \\ 
& InvAD & 86.6 / 76.5 / 69.6 & 91.9 & 98.6 / 21.3 / 32.0 & 14.6 / 31.8 / \pzo8.3 / 19.1 & 77.6 / 50.6 / 18.2 \\ 
            
            \hline
            \multirow{6}{*}{\makecell[c]{\rotatebox{90}{\textbf{regulator}}}} & RD$^*$~\cite{rd} & 66.9 / 48.8 / 47.7 & 88.6 & 98.0 / \pzo7.8 / 16.1 & \pzo4.8 / 49.4 / \pzo2.5 / \pzo8.2 & 54.5 / 40.6 / 18.9 \\ 
& UniAD$^\dagger$~\cite{uniad} & 52.6 / 34.9 / 44.3 & 70.3 & 94.2 / \pzo7.6 / 15.0 & \pzo4.8 / 41.3 / \pzo2.5 / \pzo8.1 & 43.9 / 38.9 / 16.2 \\ 
& SimpleNet$^*$~\cite{simplenet} & 50.5 / 29.0 / 43.9 & 38.1 & 76.6 / \pzo0.1 / \pzo0.6 & \pzo0.2 / 61.8 / \pzo0.1 / \pzo0.3 & 41.1 / 25.8 / 20.7 \\ 
& DiAD~\cite{diad} & 72.1 / 71.4 / 78.2 & 44.4 & 84.2 / \pzo0.4 / \pzo1.5 & \pzo0.6 / 35.9 / \pzo0.3 / \pzo0.7 & 73.9 / 28.7 / 12.3 \\ 
& InvAD-lite & 75.7 / 66.7 / 57.7 & 92.3 & 98.7 / 18.7 / 29.9 & 10.5 / 22.2 / \pzo5.9 / 17.6 & 66.7 / 49.1 / 12.9 \\ 
& InvAD & 84.6 / 78.1 / 67.8 & 95.8 & 99.5 / 24.2 / 33.2 & 14.8 / 30.0 / \pzo8.5 / 19.9 & 76.8 / 52.3 / 17.8 \\ 
            
            \hline
            \multirow{6}{*}{\makecell[c]{\rotatebox{90}{\textbf{\makecell[c]{rolled\\strip base}}}}} & RD$^*$~\cite{rd} & 97.5 / 98.7 / 94.7 & 98.4 & 99.7 / 31.4 / 39.9 & 14.2 / 40.8 / \pzo8.3 / 24.2 & 97.0 / 57.0 / 21.1 \\ 
& UniAD$^\dagger$~\cite{uniad} & 98.6 / 99.3 / 96.0 & 97.8 & 99.6 / 21.5 / 32.4 & \pzo9.6 / 24.5 / \pzo5.4 / 19.4 & 98.0 / 51.2 / 13.2 \\ 
& SimpleNet$^*$~\cite{simplenet} & 59.0 / 75.7 / 79.8 & 52.1 & 80.5 / \pzo1.7 / \pzo5.1 & \pzo2.0 / 27.4 / \pzo1.0 / \pzo2.6 & 71.5 / 29.1 / 10.1 \\ 
& DiAD~\cite{diad} & 68.4 / 55.9 / 56.8 & 63.4 & 87.7 / \pzo0.6 / \pzo3.2 & \pzo1.3 / 36.7 / \pzo0.6 / \pzo1.6 & 60.4 / 30.5 / 12.9 \\ 
& InvAD-lite & 98.0 / 99.0 / 95.2 & 98.9 & 99.8 / 31.8 / 41.1 & 18.3 / 33.5 / 10.8 / 25.9 & 97.4 / 57.6 / 20.9 \\ 
& InvAD & 99.3 / 99.6 / 97.5 & 98.8 & 99.8 / 32.2 / 42.0 & 16.4 / 29.5 / \pzo9.7 / 26.6 & 98.8 / 58.0 / 18.5 \\ 
                        
            \hline
            \multirow{6}{*}{\makecell[c]{\rotatebox{90}{\textbf{\makecell[c]{sim card \\set}}}}} & RD$^*$~\cite{rd} & 91.6 / 91.8 / 84.8 & 89.5 & 98.5 / 40.2 / 44.2 & 18.4 / 51.6 / 11.0 / 28.4 & 89.4 / 61.0 / 27.0 \\ 
& UniAD$^\dagger$~\cite{uniad} & 90.7 / 91.2 / 83.5 & 84.3 & 97.8 / 33.6 / 39.9 & 14.3 / 31.3 / \pzo8.4 / 24.9 & 88.5 / 57.1 / 18.0 \\ 
& SimpleNet$^*$~\cite{simplenet} & 63.1 / 69.7 / 70.8 & 30.8 & 71.0 / \pzo6.8 / 14.3 & \pzo4.2 / 31.5 / \pzo2.2 / \pzo7.7 & 67.9 / 30.7 / 12.6 \\ 
& DiAD~\cite{diad} & 72.6 / 53.7 / 61.5 & 60.4 & 89.9 / \pzo1.7 / \pzo5.8 & \pzo2.3 / 40.6 / \pzo1.2 / \pzo3.0 & 62.6 / 32.5 / 14.7 \\ 
& InvAD-lite & 94.2 / 94.7 / 87.9 & 88.7 & 98.6 / 46.3 / 47.0 & 24.3 / 46.1 / 14.8 / 30.7 & 92.3 / 64.0 / 28.4 \\ 
& InvAD & 93.9 / 94.1 / 87.7 & 83.1 & 97.9 / 35.4 / 41.5 & 19.9 / 43.0 / 11.8 / 26.2 & 91.9 / 58.3 / 24.9 \\ 
            
            \hline
            \multirow{6}{*}{\makecell[c]{\rotatebox{90}{\textbf{switch}}}} & RD$^*$~\cite{rd} & 84.3 / 87.2 / 77.9 & 90.9 & 94.4 / 18.9 / 26.6 & 11.7 / 34.2 / \pzo6.5 / 15.8 & 83.1 / 46.6 / 17.5 \\ 
& UniAD$^\dagger$~\cite{uniad} & 85.5 / 88.9 / 78.5 & 91.2 & 99.0 / 55.6 / 57.9 & 19.9 / 46.6 / 12.8 / 40.8 & 84.3 / 70.8 / 26.4 \\ 
& SimpleNet$^*$~\cite{simplenet} & 62.2 / 66.8 / 68.6 & 44.2 & 71.7 / \pzo3.7 / \pzo9.3 & \pzo3.0 / 37.5 / \pzo1.6 / \pzo4.9 & 65.9 / 28.2 / 14.0 \\ 
& DiAD~\cite{diad} & 73.4 / 49.4 / 61.2 & 64.2 & 90.5 / \pzo1.4 / \pzo5.3 & \pzo1.9 / 49.4 / \pzo1.0 / \pzo2.7 & 61.3 / 32.4 / 17.4 \\ 
& InvAD-lite & 92.9 / 94.7 / 87.2 & 94.9 & 99.0 / 41.0 / 46.4 & 21.7 / 30.0 / 13.1 / 30.2 & 91.6 / 62.1 / 21.6 \\ 
& InvAD & 95.7 / 96.7 / 91.1 & 95.7 & 98.9 / 42.4 / 49.7 & 24.3 / 32.6 / 14.9 / 33.1 & 94.5 / 63.7 / 23.9 \\ 
            
            \hline
            \multirow{6}{*}{\makecell[c]{\rotatebox{90}{\textbf{tape}}}} & RD$^*$~\cite{rd} & 96.0 / 95.1 / 87.6 & 98.4 & 99.7 / 42.4 / 47.8 & 21.6 / 58.1 / 13.1 / 30.4 & 92.9 / 63.3 / 30.9 \\ 
& UniAD$^\dagger$~\cite{uniad} & 97.1 / 96.1 / 89.0 & 97.3 & 99.7 / 31.4 / 37.8 & 15.7 / 42.2 / \pzo9.2 / 23.3 & 94.1 / 56.3 / 22.4 \\ 
& SimpleNet$^*$~\cite{simplenet} & 49.9 / 41.1 / 54.5 & 41.4 & 77.5 / \pzo1.2 / \pzo3.9 & \pzo1.3 / 37.6 / \pzo0.7 / \pzo2.0 & 48.5 / 27.5 / 13.2 \\ 
& DiAD~\cite{diad} & 73.9 / 57.8 / 66.1 & 47.3 & 81.7 / \pzo0.4 / \pzo2.7 & \pzo0.8 / 33.4 / \pzo0.4 / \pzo1.3 & 65.9 / 28.2 / 11.5 \\ 
& InvAD-lite & 97.5 / 96.6 / 89.9 & 98.3 & 99.7 / 42.0 / 46.3 & 22.4 / 39.7 / 13.5 / 30.1 & 94.7 / 62.7 / 25.2 \\ 
& InvAD & 97.5 / 96.3 / 90.8 & 97.8 & 99.7 / 36.9 / 42.9 & 21.0 / 40.5 / 12.5 / 27.3 & 94.9 / 59.8 / 24.7 \\ 
            
            \hline
            \multirow{6}{*}{\makecell[c]{\rotatebox{90}{\textbf{terminalblock}}}} & RD$^*$~\cite{rd} & 89.4 / 89.7 / 83.1 & 97.6 & 99.5 / 27.4 / 35.8 & 12.8 / 45.8 / \pzo7.3 / 21.6 & 87.4 / 54.2 / 22.0 \\ 
& UniAD$^\dagger$~\cite{uniad} & 87.4 / 89.4 / 80.9 & 94.1 & 99.2 / 23.1 / 31.1 & 11.1 / 42.2 / \pzo6.3 / 18.4 & 85.9 / 51.1 / 19.9 \\ 
& SimpleNet$^*$~\cite{simplenet} & 59.8 / 64.7 / 68.8 & 54.8 & 87.0 / \pzo0.8 / \pzo3.6 & \pzo1.1 / 51.7 / \pzo0.6 / \pzo1.8 & 64.4 / 30.5 / 17.8 \\ 
& DiAD~\cite{diad} & 62.1 / 36.4 / 47.8 & 38.5 & 75.5 / \pzo0.1 / \pzo1.1 & \pzo0.4 / 13.3 / \pzo0.2 / \pzo0.5 & 48.8 / 25.6 / \pzo4.6 \\ 
& InvAD-lite & 96.6 / 97.3 / 90.7 & 98.7 & 99.8 / 34.5 / 37.9 & 20.9 / 40.7 / 12.2 / 23.4 & 94.9 / 57.4 / 24.6 \\ 
& InvAD & 97.1 / 97.8 / 91.9 & 98.4 & 99.8 / 34.6 / 37.6 & 22.1 / 40.3 / 12.9 / 23.2 & 95.6 / 57.3 / 25.1 \\ 
            
            \hline
            \multirow{6}{*}{\makecell[c]{\rotatebox{90}{\textbf{toothbrush}}}} & RD$^*$~\cite{rd} & 82.0 / 83.8 / 77.2 & 88.7 & 96.9 / 26.1 / 34.2 & 14.2 / 46.6 / \pzo8.1 / 20.8 & 81.0 / 52.4 / 23.0 \\ 
& UniAD$^\dagger$~\cite{uniad} & 81.4 / 82.6 / 78.0 & 85.8 & 96.1 / 20.0 / 29.7 & 11.5 / 40.7 / \pzo6.4 / 17.4 & 80.7 / 48.6 / 19.5 \\ 
& SimpleNet$^*$~\cite{simplenet} & 65.9 / 70.0 / 70.1 & 52.6 & 84.7 / \pzo7.2 / 14.8 & \pzo5.0 / 40.8 / \pzo2.6 / \pzo8.0 & 68.7 / 35.6 / 16.1 \\ 
& DiAD~\cite{diad} & 91.2 / 93.7 / 90.9 & 54.5 & 82.0 / \pzo1.9 / \pzo6.6 & \pzo2.4 / 14.3 / \pzo1.2 / \pzo3.4 & 91.9 / 30.1 / \pzo6.0 \\ 
& InvAD-lite & 86.8 / 88.5 / 80.4 & 90.8 & 97.4 / 29.2 / 37.0 & 18.6 / 41.7 / 10.8 / 22.7 & 85.2 / 54.5 / 23.7 \\ 
& InvAD & 88.3 / 89.8 / 81.7 & 90.6 & 97.5 / 28.7 / 36.9 & 17.9 / 36.1 / 10.4 / 22.6 & 86.6 / 54.4 / 21.5 \\ 
            \toprule[0.12em] 
        \end{tabular}
    }
\end{table*}

\begin{table*}[htp]
    \centering
    \caption{\textbf{Quantitative results for each class on Real-IAD dataset.} These are respectively the image-level mAU-ROC/mAP/m$F_1$-max, region-level \redzjn{mAU-PRO}, pixel-level mAU-ROC/mAP/m$F_1$-max, the proposed pixel-level m$F_1$$^{.2}_{.8}$/mAcc$^{.2}_{.8}$/mIoU$^{.2}_{.8}$/\redzjn{mIoU-max}, and averaged \redzjn{mAD$_I$}/\redzjn{mAD$_P$}/\redzjn{mAD$^{.2}_{.8}$}. Part~\redzjn{III}.}
    \label{tab:supp_realiad_3}
    \renewcommand{\arraystretch}{1.0}
    \setlength\tabcolsep{6.0pt}
    \resizebox{1.0\linewidth}{!}{
        \begin{tabular}{p{0.3cm}<{\centering} p{2.0cm}<{\centering} p{3.9cm}<{\centering} p{1.6cm}<{\centering} p{3.9cm}<{\centering} p{4.9cm}<{\centering} p{3.3cm}<{\centering}}
            \toprule[0.17em]
            & Method & mAU-ROC/mAP/m$F_1$-max & \redzjn{mAU-PRO} & mAU-ROC/mAP/m$F_1$-max & m$F_1$$^{.2}_{.8}$/mAcc$^{.2}_{.8}$/mIoU$^{.2}_{.8}$/\redzjn{mIoU-max} & \redzjn{mAD$_I$}/\redzjn{mAD$_P$}/\redzjn{mAD$^{.2}_{.8}$} \\
            \hline
            \multirow{6}{*}{\makecell[c]{\rotatebox{90}{\textbf{toy}}}} & RD$^*$~\cite{rd} & 69.4 / 74.2 / 75.9 & 82.3 & 95.2 / \pzo5.1 / 12.8 & \pzo3.2 / 33.2 / \pzo1.6 / \pzo6.2 & 73.2 / 37.7 / 12.7 \\ 
& UniAD$^\dagger$~\cite{uniad} & 67.9 / 74.6 / 74.4 & 70.2 & 93.6 / \pzo4.4 / 10.7 & \pzo2.8 / 27.1 / \pzo1.5 / \pzo5.6 & 72.3 / 36.2 / 10.5 \\ 
& SimpleNet$^*$~\cite{simplenet} & 57.8 / 64.4 / 73.4 & 25.0 & 67.7 / \pzo0.1 / \pzo0.4 & \pzo0.2 / 23.4 / \pzo0.1 / \pzo0.2 & 65.2 / 22.7 / \pzo7.9 \\ 
& DiAD~\cite{diad} & 66.2 / 57.3 / 59.8 & 50.3 & 82.1 / \pzo1.1 / \pzo4.2 & \pzo1.3 / 58.5 / \pzo0.7 / \pzo2.1 & 61.1 / 29.1 / 20.2 \\ 
& InvAD-lite & 84.2 / 88.8 / 80.1 & 89.1 & 96.7 / 18.7 / 29.3 & 12.2 / 30.0 / \pzo6.8 / 17.2 & 84.4 / 48.2 / 16.3 \\ 
& InvAD & 85.7 / 89.0 / 82.4 & 91.0 & 98.1 / 15.6 / 24.8 & \pzo8.3 / 15.4 / \pzo4.6 / 14.2 & 85.7 / 46.2 / \pzo9.4 \\ 
            
            \hline
            \multirow{6}{*}{\makecell[c]{\rotatebox{90}{\textbf{toy brick}}}} & RD$^*$~\cite{rd} & 63.6 / 56.1 / 59.0 & 75.3 & 96.4 / 16.0 / 24.6 & \pzo7.6 / 48.7 / \pzo4.1 / 13.1 & 59.6 / 45.7 / 20.1 \\ 
& UniAD$^\dagger$~\cite{uniad} & 78.6 / 74.4 / 67.4 & 81.5 & 97.7 / 21.3 / 29.4 & \pzo8.7 / 34.4 / \pzo4.9 / 17.2 & 73.5 / 49.5 / 16.0 \\ 
& SimpleNet$^*$~\cite{simplenet} & 58.3 / 49.7 / 58.2 & 56.3 & 86.5 / \pzo5.2 / 11.1 & \pzo3.9 / 44.3 / \pzo2.1 / \pzo5.9 & 55.4 / 34.3 / 16.8 \\ 
& DiAD~\cite{diad} & 68.4 / 45.3 / 55.9 & 66.4 & 93.5 / \pzo3.1 / \pzo8.1 & \pzo2.8 / 52.4 / \pzo1.4 / \pzo4.2 & 56.5 / 34.9 / 18.9 \\ 
& InvAD-lite & 69.3 / 61.9 / 60.8 & 78.1 & 96.5 / 15.9 / 23.9 & \pzo8.6 / 46.6 / \pzo4.7 / 13.6 & 64.0 / 45.4 / 20.0 \\ 
& InvAD & 72.4 / 65.5 / 63.3 & 80.2 & 97.3 / 23.4 / 30.6 & 10.4 / 35.0 / \pzo5.8 / 18.1 & 67.1 / 50.4 / 17.1 \\ 
            
            \hline
            \multirow{6}{*}{\makecell[c]{\rotatebox{90}{\textbf{transistor1}}}} & RD$^*$~\cite{rd} & 91.0 / 94.0 / 85.1 & 95.1 & 99.1 / 29.6 / 35.5 & 14.6 / 53.4 / \pzo8.4 / 21.9 & 90.0 / 54.7 / 25.5 \\ 
& UniAD$^\dagger$~\cite{uniad} & 94.2 / 96.2 / 89.4 & 94.2 & 99.0 / 26.6 / 33.6 & 12.4 / 43.0 / \pzo7.1 / 20.2 & 93.3 / 53.1 / 20.8 \\ 
& SimpleNet$^*$~\cite{simplenet} & 62.2 / 69.2 / 72.1 & 35.3 & 71.7 / \pzo5.1 / 11.3 & \pzo3.7 / 45.0 / \pzo1.9 / \pzo6.0 & 67.8 / 29.4 / 16.9 \\ 
& DiAD~\cite{diad} & 73.1 / 63.1 / 62.7 & 58.1 & 88.6 / \pzo7.2 / 15.3 & \pzo5.6 / 33.9 / \pzo3.0 / \pzo8.3 & 66.3 / 37.0 / 14.2 \\ 
& InvAD-lite & 96.2 / 97.3 / 92.2 & 96.9 & 99.5 / 40.6 / 42.1 & 19.1 / 37.9 / 11.3 / 26.7 & 95.2 / 60.7 / 22.8 \\ 
& InvAD & 97.3 / 98.1 / 93.6 & 97.0 & 99.5 / 35.1 / 37.9 & 20.9 / 43.2 / 12.2 / 23.4 & 96.3 / 57.5 / 25.4 \\ 
            
            \hline
            \multirow{6}{*}{\makecell[c]{\rotatebox{90}{\textbf{u block}}}} & RD$^*$~\cite{rd} & 89.5 / 85.0 / 74.2 & 96.9 & 99.6 / 40.5 / 45.2 & 17.7 / 53.2 / 10.5 / 29.1 & 82.9 / 61.8 / 27.1 \\ 
& UniAD$^\dagger$~\cite{uniad} & 88.1 / 83.1 / 73.5 & 94.7 & 99.3 / 22.8 / 30.1 & \pzo9.9 / 25.6 / \pzo5.6 / 17.7 & 81.6 / 50.7 / 13.7 \\ 
& SimpleNet$^*$~\cite{simplenet} & 62.4 / 48.4 / 51.8 & 34.0 & 76.2 / \pzo4.8 / 12.2 & \pzo3.8 / 30.7 / \pzo2.0 / \pzo6.5 & 54.2 / 31.1 / 12.2 \\ 
& DiAD~\cite{diad} & 75.2 / 68.4 / 67.9 & 54.2 & 88.8 / \pzo1.6 / \pzo5.4 & \pzo2.0 / 56.0 / \pzo1.0 / \pzo2.8 & 70.5 / 31.9 / 19.6 \\ 
& InvAD-lite & 89.9 / 85.4 / 75.5 & 96.4 & 99.6 / 33.8 / 42.0 & 20.0 / 36.3 / 11.7 / 26.6 & 83.6 / 58.5 / 22.7 \\ 
& InvAD & 92.3 / 88.5 / 79.3 & 96.5 & 99.6 / 30.9 / 39.6 & 17.6 / 32.7 / 10.2 / 24.7 & 86.7 / 56.7 / 20.2 \\ 
            
            \hline
            \multirow{6}{*}{\makecell[c]{\rotatebox{90}{\textbf{usb}}}} & RD$^*$~\cite{rd} & 84.9 / 84.3 / 75.1 & 91.0 & 98.1 / 26.4 / 35.2 & 11.8 / 41.9 / \pzo6.8 / 21.1 & 81.4 / 53.2 / 20.2 \\ 
& UniAD$^\dagger$~\cite{uniad} & 80.5 / 81.4 / 71.0 & 85.9 & 98.1 / 19.9 / 29.7 & \pzo8.0 / 30.5 / \pzo4.5 / 17.4 & 77.6 / 49.2 / 14.3 \\ 
& SimpleNet$^*$~\cite{simplenet} & 57.0 / 55.3 / 62.9 & 52.4 & 81.1 / \pzo1.5 / \pzo4.9 & \pzo1.2 / 30.8 / \pzo0.6 / \pzo2.5 & 58.4 / 29.2 / 10.9 \\ 
& DiAD~\cite{diad} & 58.9 / 37.4 / 45.7 & 28.0 & 78.0 / \pzo1.0 / \pzo3.1 & \pzo1.3 / 39.7 / \pzo0.7 / \pzo1.6 & 47.3 / 27.4 / 13.9 \\ 
& InvAD-lite & 93.1 / 92.8 / 85.5 & 96.6 & 99.5 / 37.7 / 43.6 & 23.9 / 46.9 / 14.3 / 27.9 & 90.5 / 60.3 / 28.4 \\ 
& InvAD & 93.6 / 92.8 / 86.3 & 97.1 & 99.5 / 36.7 / 42.7 & 19.0 / 32.2 / 11.4 / 27.1 & 90.9 / 59.6 / 20.9 \\ 
                        
            \hline
            \multirow{6}{*}{\makecell[c]{\rotatebox{90}{\textbf{usb adaptor}}}} & RD$^*$~\cite{rd} & 71.1 / 61.4 / 62.2 & 73.1 & 94.5 / \pzo9.8 / 17.9 & \pzo7.6 / 35.3 / \pzo4.1 / 10.3 & 64.9 / 40.7 / 15.7 \\ 
& UniAD$^\dagger$~\cite{uniad} & 76.6 / 71.8 / 64.2 & 78.9 & 96.4 / 10.9 / 19.7 & \pzo7.6 / 27.5 / \pzo4.1 / 10.9 & 70.9 / 42.3 / 13.1 \\ 
& SimpleNet$^*$~\cite{simplenet} & 47.5 / 38.4 / 56.5 & 28.9 & 67.9 / \pzo0.2 / \pzo1.3 & \pzo0.4 / 46.8 / \pzo0.2 / \pzo0.6 & 47.5 / 23.1 / 15.8 \\ 
& DiAD~\cite{diad} & 76.9 / 60.2 / 67.2 & 75.5 & 94.0 / \pzo2.3 / \pzo6.6 & \pzo2.7 / 39.0 / \pzo1.4 / \pzo3.4 & 68.1 / 34.3 / 14.3 \\ 
& InvAD-lite & 79.5 / 73.3 / 67.3 & 80.1 & 97.1 / 16.1 / 24.5 & 12.7 / 30.2 / \pzo7.0 / 13.9 & 73.4 / 45.9 / 16.6 \\ 
& InvAD & 79.7 / 73.1 / 66.3 & 77.0 & 96.2 / 15.0 / 23.6 & 10.8 / 27.4 / \pzo5.9 / 13.4 & 73.0 / 44.9 / 14.7 \\ 
            
            \hline
            \multirow{6}{*}{\makecell[c]{\rotatebox{90}{\textbf{vcpill}}}} & RD$^*$~\cite{rd} & 85.1 / 80.3 / 72.4 & 88.7 & 98.3 / 43.1 / 48.6 & 13.6 / 37.8 / \pzo8.2 / 32.1 & 79.3 / 63.3 / 19.9 \\ 
& UniAD$^\dagger$~\cite{uniad} & 88.0 / 85.7 / 75.4 & 92.0 & 99.2 / 47.3 / 47.4 & 12.0 / 21.6 / \pzo7.3 / 31.1 & 83.0 / 64.6 / 13.6 \\ 
& SimpleNet$^*$~\cite{simplenet} & 59.0 / 48.7 / 56.4 & 22.0 & 68.2 / \pzo1.1 / \pzo3.3 & \pzo1.2 / 39.5 / \pzo0.6 / \pzo1.7 & 54.7 / 24.2 / 13.8 \\ 
& DiAD~\cite{diad} & 64.1 / 40.4 / 56.2 & 60.8 & 90.2 / \pzo1.3 / \pzo5.2 & \pzo1.9 / 32.8 / \pzo1.0 / \pzo2.7 & 53.6 / 32.3 / 11.9 \\ 
& InvAD-lite & 84.1 / 81.4 / 70.9 & 87.0 & 98.4 / 41.6 / 49.5 & 18.3 / 37.6 / 11.0 / 32.8 & 78.8 / 63.2 / 22.3 \\ 
& InvAD & 89.4 / 87.2 / 77.9 & 90.3 & 98.8 / 47.7 / 54.0 & 18.4 / 30.9 / 11.5 / 37.0 & 84.8 / 66.8 / 20.3 \\ 
            
            \hline
            \multirow{6}{*}{\makecell[c]{\rotatebox{90}{\textbf{\makecell[c]{wooden\\beads}}}}} & RD$^*$~\cite{rd} & 81.2 / 78.9 / 70.9 & 85.7 & 98.0 / 27.1 / 34.7 & 13.6 / 57.3 / \pzo7.8 / 21.3 & 77.0 / 53.3 / 26.2 \\ 
& UniAD$^\dagger$~\cite{uniad} & 80.5 / 79.8 / 69.5 & 84.0 & 97.7 / 19.9 / 27.0 & 10.0 / 35.7 / \pzo5.6 / 15.6 & 76.6 / 48.2 / 17.1 \\ 
& SimpleNet$^*$~\cite{simplenet} & 55.1 / 52.0 / 60.2 & 28.3 & 68.1 / \pzo2.4 / \pzo6.0 & \pzo2.0 / 32.7 / \pzo1.0 / \pzo3.1 & 55.8 / 25.5 / 11.9 \\ 
& DiAD~\cite{diad} & 62.1 / 56.4 / 65.9 & 45.6 & 85.0 / \pzo1.1 / \pzo4.7 & \pzo1.8 / 39.6 / \pzo0.9 / \pzo2.4 & 61.5 / 30.3 / 14.1 \\ 
& InvAD-lite & 85.4 / 84.4 / 74.8 & 88.1 & 98.3 / 30.6 / 37.7 & 16.4 / 38.3 / \pzo9.5 / 23.2 & 81.5 / 55.5 / 21.4 \\ 
& InvAD & 86.4 / 85.2 / 75.8 & 87.3 & 98.4 / 33.9 / 39.1 & 19.8 / 44.6 / 11.6 / 24.3 & 82.5 / 57.1 / 25.3 \\ 
            
            \hline
            \multirow{6}{*}{\makecell[c]{\rotatebox{90}{\textbf{woodstick}}}} & RD$^*$~\cite{rd} & 76.9 / 61.2 / 58.1 & 85.0 & 97.8 / 30.7 / 38.4 & 15.4 / 52.8 / \pzo8.9 / 23.2 & 65.4 / 55.6 / 25.7 \\ 
& UniAD$^\dagger$~\cite{uniad} & 81.2 / 73.5 / 64.4 & 76.4 & 94.7 / 40.2 / 45.9 & 19.5 / 39.1 / 11.8 / 29.8 & 73.0 / 60.3 / 23.5 \\ 
& SimpleNet$^*$~\cite{simplenet} & 58.2 / 35.6 / 45.2 & 32.0 & 76.1 / \pzo1.4 / \pzo6.0 & \pzo1.9 / 48.1 / \pzo1.0 / \pzo3.1 & 46.3 / 27.8 / 17.0 \\ 
& DiAD~\cite{diad} & 74.1 / 66.0 / 62.1 & 60.7 & 90.9 / \pzo2.6 / \pzo8.0 & \pzo3.3 / 27.1 / \pzo1.7 / \pzo4.2 & 67.4 / 33.8 / 10.7 \\ 
& InvAD-lite & 78.5 / 65.9 / 60.3 & 83.1 & 97.6 / 36.7 / 41.6 & 19.9 / 49.3 / 11.9 / 26.2 & 68.2 / 58.6 / 27.0 \\ 
& InvAD & 80.8 / 67.4 / 62.2 & 85.5 & 98.0 / 37.4 / 43.3 & 22.4 / 50.6 / 13.4 / 27.6 & 70.1 / 59.6 / 28.8 \\ 
            
            \hline
            \multirow{6}{*}{\makecell[c]{\rotatebox{90}{\textbf{zipper}}}} & RD$^*$~\cite{rd} & 95.3 / 97.2 / 91.2 & 96.3 & 99.1 / 44.7 / 50.2 & 20.0 / 53.6 / 12.2 / 34.2 & 94.6 / 64.7 / 28.6 \\ 
& UniAD$^\dagger$~\cite{uniad} & 98.1 / 98.9 / 95.5 & 94.7 & 98.3 / 32.2 / 35.5 & 12.7 / 28.3 / \pzo7.3 / 21.6 & 97.5 / 55.3 / 16.1 \\ 
& SimpleNet$^*$~\cite{simplenet} & 77.2 / 86.7 / 77.6 & 55.5 & 89.9 / 23.3 / 31.2 & 13.2 / 50.6 / \pzo7.4 / 18.5 & 80.5 / 48.1 / 23.7 \\ 
& DiAD~\cite{diad} & 86.0 / 87.0 / 84.0 & 53.5 & 90.2 / 12.5 / 18.8 & \pzo9.7 / 31.6 / \pzo5.2 / 10.4 & 85.6 / 40.5 / 15.5 \\ 
& InvAD-lite & 98.9 / 99.4 / 95.8 & 96.7 & 99.2 / 55.5 / 57.9 & 25.0 / 42.5 / 15.8 / 40.8 & 98.0 / 70.9 / 27.8 \\ 
& InvAD & 98.7 / 99.1 / 95.9 & 96.1 & 99.1 / 48.3 / 52.4 & 26.7 / 52.0 / 16.6 / 35.5 & 97.9 / 66.6 / 31.8 \\ 
            \toprule[0.12em] 
        \end{tabular}
    }
\end{table*}